# Analysis and Mortality Prediction using Multiclass Classification for Older Adults with Type 2 Diabetes

**Ruchika Desure**[1,2] and **Gutha Jaya Krishna**[2,3,*]

[1]Liverpool John Moores University, Liverpool, L3 3AF, UK

[2]Upgrad Education Pvt. Ltd, Mumbai, India

[3]Administrative Staff College of India, Raj Bhavan, Hyderabad, Telangana, India

ruchikadesure1995@gmail.com ; krishna.gutha@gmail.com[*]

**ABSTRACT**

Designing proper treatment plans to manage diabetes requires health practitioners to pay heed to the individual's remaining life along with the comorbidities affecting them. Older adults with Type 2 Diabetes Mellitus (T2DM) are prone to experience premature death or even hypoglycaemia. Hence, the methodology proposed in this study is to build a prognostic model that performs three-class classification. The structured dataset utilized has 68 potential mortality predictors for 275,190 diabetic U.S. military Veterans aged 65 years or older. A new target variable is invented by combining the two original target variables. Outliers are handled by discretizing the continuous variables. Categorical variables have been dummy encoded. Class balancing is achieved by random under-sampling. A benchmark regression model is built using Multinomial Logistic Regression with LASSO. Chi-Squared and Information Gain are the filter-based feature selection techniques utilized. Classifiers such as Multinomial Logistic Regression, Random Forest, Extreme Gradient Boosting (XGBoost), and One-vs-Rest classifier are employed to build various models. Contrary to expectations, all the models have constantly underperformed. XGBoost has given the highest accuracy of 53.03% with Chi-Squared feature selection. All the models have consistently shown an acceptable performance for Class 3 (remaining life is more than 10 years), significantly low for Class 1 (remaining life is up to 5 years), and the worst for Class 2 (remaining life is more than 5 but up to 10 years). Features analysis has deduced that almost all input variables are associated with multiple target classes. The high dimensionality of the input data after dummy encoding seems to have confused the models, leading to misclassifications. The approach taken in this study is ineffective in producing a high-performing predictive model but lays a foundation as this problem has never been viewed from a multiclass classification perspective.

# 1. INTRODUCTION

## 1.1	Background of the Study

Diabetes is a non-communicable but chronic disease. According to the World Health Organization (WHO), diabetes has been the major cause of other complications such as blindness, heart attacks, kidney failure, lower limb amputation, and stroke (The World Health Organization, 2022). There are more chances of diabetic patients experiencing poor outcomes for COVID-19 as well as other infections. It is of three types – Type 1 diabetes, Type 2 diabetes and Gestational diabetes (The World Health Organization, 2022). Type 1 diabetes is caused due to deficiency of insulin production and hence insulin needs to be administered every day. Type 2 diabetes or Type 2 Diabetes Mellitus (T2DM) is caused largely due to physical inactivity and excess body weight. It happens when the body does not use insulin effectively. Gestational diabetes develops in some pregnant women.

As reported by the WHO, diabetes caused a total of 1.5 million deaths in the year 2019 with 48% being experienced by people before 70 years of age (The World Health Organization, 2022). Additionally, diabetes caused 460,000 kidney disease deaths, and around 20% of cardiovascular deaths were due to the rise in blood glucose. More than 95% of diabetic people have T2DM (The World Health Organization, 2022). According to the Centers for Disease Control and Prevention (CDC), people aged 50 years live 6 years more than same-aged people having T2DM (Centers for Disease Control and Prevention, 2022).

The risk of mortality, as well as various microvascular and macrovascular complications increases due to diabetes. It is uncertain as to when more intensive glucose control should be pursued especially for individuals who have other complications and comorbidities that lessen the life span (Griffith et al., 2020). Patients' comorbidities and remaining life expectancy need to be taken into account when treatment plans are being devised for the patients. Many influencing factors that can increase diabetes complications need to be considered and hence determining the remaining life expectancy of a person having diabetes is a challenge for health practitioners. Thus, the development of an all-cause mortality prediction model can be very useful for health practitioners as it can identify risky diabetic patients and also determine the remaining life expectancy of their patients.

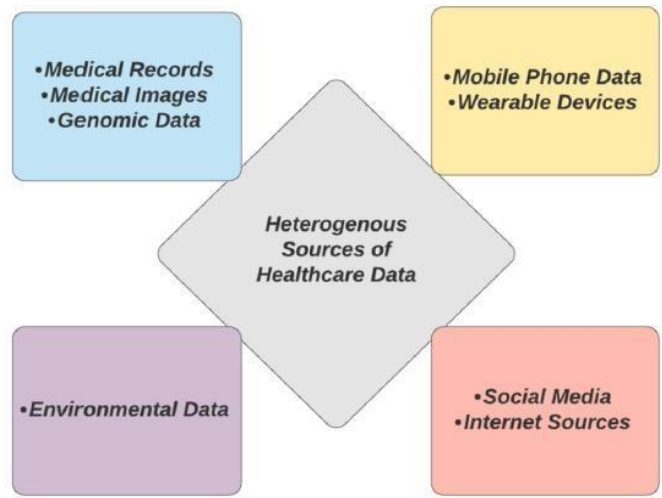

Figure 1.1 Various heterogeneous sources of healthcare data

(Image taken from Swain et al., 2022)

Predictive Analytics uses historical medical data of patients to help predict health related events that may occur in the future. Healthcare data is huge and figure 1.1 depicts the various sources from which the healthcare data is obtained. To process these types of data, ML techniques can be used in various healthcare industry applications. There have been some researches done in the past using predictive analytics majorly for diagnosis and prognosis of various diseases, mortality prediction, predicting in-hospital length of stay and mortality, and prediction of hospital readmissions. Figure 1.2 depicts the various applications of Machine Learning in the healthcare sector.

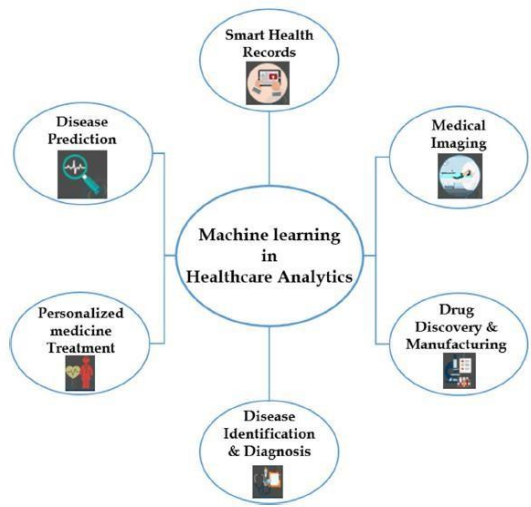

Figure 1.2 Applications of Machine Learning (ML) in healthcare

(Image taken from Malik et al., 2022)

People having T2DM can live longer if they can reduce complication risks by achieving their treatment goals. But health practitioners can form proper diabetes treatment plans only when the life expectancy is considered along with the patient's comorbidities (Vaidya et al., 2022). There have been several machine learning models developed in the past to predict the occurrence of all-cause mortality in patients with T2DM. But there are challenges due to which models are often not directly applicable to every population or the broader age groups.

The ENFORCE (Estimation of Mortality Risk in Type 2 Diabetic Patients) is a 6-year mortality risk prediction model trained on real-life samples (Copetti et al., 2019) and RECODe (Risk Equations for Complications of Type 2 Diabetes) is a 10-year mortality risk prediction model (Copetti et al., 2021) that have proved by validating externally that models perform differently on a clinical trial data and real-life data. These well-established models were even successfully applied in the early stage of T2DM (Copetti et al., 2021).

The UKPDS (United Kingdom Prospective Diabetes Study) model which is originally developed using UK diabetic population data needed a new model for the US population as there are significant ethnicity and treatment differences in both populations. Thus, the model named BRAVO (Building, Relating, Assessing, and Validating Outcomes) (Shao et al., 2018) was developed. A new 5-year mortality risk prediction model was developed using the Chinese population data because of the differences between the Asian and the Western populations in terms of the diabetes management approach, socioeconomic factors, and genetics (Qi et al., 2022).

7-year and 10-year models were developed on a sample of Taiwanese patients by considering some comorbidities and drug treatments along with clinical biomarkers as predictors (Chiu et al., 2021). As adults aged 65 years or older suffering from T2DM are more prone to experience hypoglycaemia or even premature death, a significant elderly-specific model for 5year mortality prediction was developed in another study (Chang et al., 2017).

There have been more research and major findings have been documented through these past studies. The prediction model's performance varies on clinical-trial data and real-world data. There has been a difference in these model performances based on the population samples as well. Some studies have not considered the comorbidities and prescriptions while developing the models even though these play a vital role and are influencing factors of mortality risk. Thus, the impact of including comorbidities and medicine prescriptions, the transportability of models to different genetics and cultural backgrounds, and the impact of good-performing models on different age groups; these points are still needed to be explored more and preferably with large-scale data.

In all the studies mentioned above, the Cox proportional hazards regression was used to develop the prediction models. But there is another study in which two point-based mortality risk prediction models have been developed using Logistic Least Absolute Shrinkage and Selection Operator (LASSO) Regressions (Griffith et al., 2020). The separate models for predicting 5-year and 10-year mortality are developed on large-scale data including adults with T2DM and aged 65 years or older (Griffith et al., 2020). The most significant part of this study is the number of potential predictors that have not been considered in any previous study. Demographics, medications, comorbidities and biomarkers have been included as the 68 risk factors for predicting mortality (Vaidya et al., 2022).

There are some possibilities that can be explored for this 5-year and 10-year mortality data. In some recent studies, various feature selection techniques have been used for medical datasets. The filter-based feature selection techniques have an advantage over embedded and wrapper-based ones by being classifier-independent (Zhang et al., 2022). For low computation cost, the filter-based feature selection techniques can be explored instead of the LASSO which is an embedded technique. Also, there are several supervised classification machine learning techniques available which have been employed to solve various healthcare industry problems. But these have not been explored much for mortality prediction.

Determining the remaining life expectancy of a patient by considering their comorbidities and complications would help health practitioners in developing proper and efficient treatment plans for individuals. Also, there is a lack of mortality prediction models specifically for the elderly population. Therefore, the purpose of this research is to propose a multiclass classification methodology with a filter-based feature selection technique and supervised ML algorithm that predicts all-cause mortality subject to a variety of individual-level risk factors in older adults aged 65 years or older having T2DM. Instead of multiple prediction models, the intention is to build just one predictive model. The business significance is that this single mortality prediction model would help clinicians and hospitals in predicting the remaining life of older diabetic patients by considering their comorbidities and other complications. The single prediction model would provide more precise and drilled-down classifications.

## 1.2     Problem Statement

Diabetes is a chronic disease which causes other complications such as blindness, heart attacks, kidney failure, lower limb amputation, and stroke. As reported by the WHO, diabetes caused a total of 1.5 million deaths in the year 2019 with 48% being experienced by people before 70 years of age and more than 95% of diabetic people have T2DM (The World Health Organization, 2022). This disease increases the risk of mortality as well as various microvascular and macrovascular complications (Griffith et al., 2020). But individuals having T2DM can still reduce the mortality risk by achieving treatment goals. Proper treatment plans can be formed by the health practitioners when the individual's remaining life expectancy is considered along with the comorbidities (Vaidya et al., 2022). But predicting the remaining life is a challenge!

Machine Learning (ML) algorithms have been used in past studies to develop prediction models that can predict the occurrence of all-cause mortality in individuals with T2DM. ML models require a vast amount of input data to find patterns and learn from them. Studies conducted by De Cosmo et al., 2013; Copetti et al., 2019, 2021 have achieved good model performances but they have used small datasets for training the ML models. The datasets used have more or less 1000 records only. Now, randomized clinical trial data consists of highly selective populations which are managed in tightly controlled settings (Kim et al., 2018). Whereas, real-world data is the data collected from daily life that are outside the scope of highly controlled settings. A study conducted by Copetti et al., 2019 with the well-established ENFORCE model showed that the model performance lowered on clinical trial settings data and this phenomenon was earlier seen in RECODe model study as well conducted by Basu et al., 2018.

Asians develop diabetes at a younger age compared to other populations and also have a higher tendency of developing its complications (Gujral et al., 2013; Ma and Chan, 2013). As the UKPDS, BRAVO, ENFORCE, and RECODe models are based on the Western populations, these often cannot be directly applied to Asian populations because of the differences in socioeconomic factors, diabetes management approach, and genetics. In another study, an elderly-specific model was developed for the Taiwanese population to predict 5-year mortality risk with 14 clinical predictors. This model by Chang et al., 2017 is significant as older adults are more prone to experience hypoglycaemia or even premature death. The model was developed as there was a lack of prediction models for an older population.

Comorbidities and medication details of patients were not considered as predictors in the study conducted by (Chang et al., 2017). Research conducted by Wells et al., 2008; De Cosmo et al., 2013; Robinson et al., 2015; Wan et al., 2017 have considered very less comorbidities or not considered at all. Griffith et al., 2020 developed 5-year and 10-year mortality prediction models with acceptable performance but stated a limitation in their study. To get better predictive accuracy, the Logistic LASSO Regression methodology used allows bias in the odds ratios. But the dataset used in this study is significantly good. It is large-scale real-world data with more than 200000 records, 68 potential predictors of mortality, and 2 target variables one for 5-year mortality status and the other one for 10-year. The data is of older adults aged 65 years or older and having T2DM.

The dataset used by Griffith et al., 2020 can be used for mortality prediction in a different way compared to the original research. In this dataset, the records of patients shown dead at the 5year mark are also marked dead at the 10-year mark. The 10-year mortality prediction model using binary classification cannot determine whether the patient would die within the next 5 years. It can just predict whether the person will die in the next 10 years or not. A separate 5year mortality prediction model would be required to predict mortality in the upcoming 5 years. But this challenge can be addressed if the binary target variables are converted and made applicable for multiclass classification. Thus, just one multiclass classification model would be able to predict whether the patient's remaining life is "up to 5 years", "more than 5 but up to 10 years" or "more than 10 years".

The mortality prediction models developed in past studies often lack generalizability and transportability to other populations or even to the different broader age groups. There have been various studies conducted for the general age group but there is still a shortage of mortality prediction models for older adults with T2DM. As the human body ages, its immune system weakens. Thus, older adults who are aged 65 years and older are more susceptible to health-related problems. On top of this, the risk of complications and comorbidities increases if they have diabetes. They are more susceptible to hypoglycaemia or premature death. Thus, they need a proper diabetes treatment plan so that the diabetes is manageable and their health doesn't get severely affected. This can prevent hospitalization or at least reduce hospital readmissions. A prediction model is required for the older population aged 65 years or older with Type 2 Diabetes to identify high-risk individuals and prevent premature mortality.

## 1.3 Aim and Objectives

The main aim of this research is to propose a multiclass classification methodology for the prediction of all-cause mortality in older adults having Type 2 Diabetes Mellitus subject to a variety of individual-level risk factors. The goal of this study is to build just one model that estimates the remaining life expectancy of older type 2 diabetic patients by considering their comorbidities and other complications. This can aid health practitioners in developing specific treatment plans for each patient.

The objectives of the research formulated based on the aim of this study are as follows:
- To explore the state-of-the-art models developed for mortality prediction and other supervised multiclass classification problems.
- To apply and compare a filter-based feature selection technique with the LASSO technique for the selection of mortality predictors.
- To employ different supervised multiclass classification algorithms that would classify mortality based on the predictor risk factors.
- To evaluate the performance of these different predictive models and to identify the best model.

## 1.4 Research Questions

The research questions suggested for each of the research objectives mentioned in the previous section are as follows:
1. Instead of separate prediction models for 5-year and 10-year mortality, can only one multiclass classification prediction model be built on the mortality dataset considered for this study?
2. Will the filter-based feature selection techniques be better than the LASSO technique to improve the model performance?
3. Which supervised machine learning algorithm can be employed for multiclass classification other than the base methodology for the chosen problem?
4. Will the performance of the proposed methodology be better than the base methodology?

## 1.5 Scope of the Study

The dataset used in this study is structured data. It consists of only older adults aged 65 years or more. All the patients have Type 2 Diabetes Mellitus (T2DM). Mortality status is followed up at the 5-year as well as at the 10-year mark. The two feature selection techniques to be explored are filter-based. Only the supervised machine learning techniques for multiclass classification would be considered. Also, the study is about predicting all-cause mortality in T2DM patients.

There are some points to be noted as out of the scope of this study. The dataset does not have patients having Type 1 diabetes or Gestational diabetes. External validation of the built models using a different dataset is not considered in this study. Using deep learning algorithms for multiclass classification is also out of scope.

## 1.6 Significance of the Study

Diabetes increases mortality as well as the risk of other complications. Several diabetes clinical practice guidelines state that the patient's comorbidities, complications, and life expectancy should be considered while setting treatment goals (Vaidya et al., 2022). But the uncertainty of life expectancy for patients can make developing treatment plans difficult for health practitioners (Vaidya et al., 2022). A high-performing multiclass classification ML model can help health practitioners as it will predict whether the patient's remaining life is "up to 5 years", "more than 5 but up to 10 years" or "more than 10 years".

The dataset used for this study is very significant as it has a large number of features compared to the datasets used in other studies conducted on all-cause mortality prediction for patients having type 2 diabetes. There is no other dataset having Type 2 diabetes patient details which consider so many potential predictors (68 in total) of mortality. The dataset has details such as patient demographics, medication history, comorbidities, laboratory values and anthropomorphic measurements, procedure codes, and previous health service utilization (Griffith et al., 2020). In this study, four supervised multiclass classification algorithms are used with two filter-based feature selection techniques to predict all-cause mortality risk in older adults having type 2 diabetes. Converting this dataset for a multiclass classification has never been attempted in previous studies. Thus, this novel approach has opened up new frontiers.

## 1.7 Structure of the Study

The structure of the thesis is as follows:

Section 1.1 describes the background of the research in mortality risk prediction in Type 2 Diabetes Mellitus followed by the problem statement in Section 1.2. The aim and objectives of this current study are mentioned in section 1.3 of this section. Further, section 1.4 presents the research questions. Scope of the study wherein the in-scope and out-of-scope of this current work have been mentioned in section 1.5. Then, the significance of this study is mentioned in section 1.6.

Section 2 describes the literature review performed to understand the background of the study and highlights the issues mentioned in Section 1. Section 2.1 describes the relevance of machine learning in the field of healthcare. Further, section 2.2 describes the application of predictive analytics in the healthcare sector. Section 2.3 elaborates on the various works done in the past related to the all-cause mortality prediction for patients with Type 2 Diabetes Mellitus. Then, section 2.4 describes the various feature selection techniques used by researchers for features in medical datasets. Various multiclass classification ML algorithms recently employed in the healthcare sector have been discussed in section 2.5. The key findings from the body of the literature are discussed together in section 2.6 and lastly, a summary of the literature review is given in section 2.7.

Section 3 describes the research approach. Section 3.1 describes the process flow followed in this study. The research design is also depicted. The data selection process has been discussed. Each variable present in the dataset is mentioned and the overall description is given. Further, the data pre-processing and data transformation steps have been elaborated. Feature selection techniques have been discussed in detail. Then the train-test split, class balancing, and cross- validation steps have been elaborated. Section 3.2 describes all the modelling techniques that have been used in this study. First, a benchmark model which is a regression model consisting of Multinomial Logistic Regression with LASSO regularization is explained. Further, all the classifiers such as Multinomial Logistic Regression, Random Forest, XGBoost, and One-Vs-Rest Classifier have also been explained. Section 3.3 gives the details on the evaluation metrics that would be used to check the performance of each model. Lastly, summarizes section 3.

Section 4 gives the details of the model development wherein the sub-sections include the details of the methodology decided for this study. Section 4.1 describes all the steps taken to clean the original dataset. Outlier and Missing values have been identified and handled. Section 4.2 describes the findings from the univariate analysis conducted. The distribution of mortality classes has also been depicted. Section 4.3 gives the details about the dataset after conducting a train-test split. Section 4.4 is about the bivariate analysis conducted using ChiSquare Test in the IBM SPSS statistics software. Section 4.5 is about the dummy encoding and label encoding performed. Section 4.6 details the class balancing process by conducting an under-sampling technique on the training data. Section 4.7 is about the model implementation phase wherein all the configurations applied for each model have been mentioned in detail. Section 4.8 describes the evaluation process for all the models built. Section 4.9 elaborates on the additional feature analysis performed using Chi-Square Statistics in Python. The hardware and software requirements to conduct this study are given in section 4.10. Finally, summarizes the entire section 4.

Section 5 includes the output results and the findings gathered from them. Section 5.1 discusses the performance of the benchmark model. Evaluation of all the remaining classifiers is done in section 5.2. The level of agreement using Cohen's Kappa has been discussed in section 5.3. Additional analysis conducted on the input features has been described in section 5.4. Finally, summarizes all the findings mentioned in section 5.

Section 6 discusses the overall study briefly. Based on the results achieved, conclusions have been drawn and elaborated in section 6.1. Some future recommendations have been mentioned in section 6.2.

2. **LITERATURE REVIEW**

This section will illuminate the recent studies carried out regarding the relevance of Machine Learning and predictive analytics in healthcare. The sections will include summarization of various literatures, beginning with the overall healthcare sector and then channelling down to some research works on mortality prediction for humans more specifically the all-cause mortality in patients having Type 2 Diabetes Mellitus. Then, some latest studies related to feature selection techniques used for medical datasets will also be discussed. In the end, research works on the use of machine learning algorithms for multiclass classification healthcare industry problems, would be seen. Section 2.3 of this section 2 focuses on the previous research works that have been conducted on mortality prediction in patients having Type 2 Diabetes Mellitus (T2DM). Limitations and challenges present in these previous studies would be highlighted.

## 2.1 Relevance of Machine Learning in Healthcare

In the computer science domain, Machine Learning (ML) is considered as a branch of Artificial Intelligence (AI) (Swain et al., 2022). One example is, the clinical documents of patients are mostly hand-written by the health practitioner and only some reports are generated by machines (Swain et al., 2022). Thus, ML techniques have proved to be helpful in automating and improving the performance of major healthcare processes such as clinical workflows, diagnosis, planning treatments, and prognosis (Swain et al., 2022).

## 2.2 Predictive Analytics in Healthcare

In recent years, there has been an increase in the application of predictive analytics in the healthcare sector. Historical medical data of patients is used to analyze and identify certain trends and patterns from the data which can help in predicting events that may occur in the future. The availability of proper detailed healthcare data is very necessary for predictive analytics and therefore high-quality and correct details of the patients should be stored. There have been some researches done in the past using predictive analytics majorly for diagnosis and prognosis of various diseases, mortality prediction, predicting in-hospital length of stay and mortality, and prediction of hospital readmissions. An individual's risk due to an already present disease is calculated by diagnostic prediction models whereas prognostic prediction models calculate the risk of health conditions that can occur in future (van Smeden et al., 2021).

Cancer is the second most cause of death worldwide and the early diagnosis of this disease is necessary as its advanced stage mostly leads to mortality. Painuli et al., 2022 conducted a study to review developments from the year 2016 to 2021 (6 years) in cancer diagnosis using machine learning and deep learning. Detection of six types of cancers namely brain, skin, lung, breast, pancreatic, and liver cancer were considered in this study. The performances of the best state-of-the-art models have given an accuracy of about 97.5% to 99.9%. It has been observed that for cancer detection, the Support Vector Machine (SVM) based ML models and the Convolutional Neural Network (CNN) based DL models gave more accuracy in these past studies. For cancer disease prediction, random forest and decision trees proved to be the best.

Brain-related diseases if detected early and with good accuracy can result in overall less treatment cost as well as a significant reduction in mortality. A review by Ahire et al., 2022 illustrated the various machine learning techniques used to classify Electroencephalography (EEG) signals for the early detection of Dyslexia which is a learning disorder. ML algorithms can process and classify EEG signa to examine and understand brain activities. The Support Vector Machine (SVM) ML algorithm was found to be highly successful in EEG signal classification.

Shorewala, 2021 in their study had explored the effectiveness of ensemble ML techniques in predicting coronary heart disease. The early detection of coronary heart disease is crucial as it is one of the leading causes of death globally. The stacking ensemble consisting of various heterogeneous models was the most effective one giving an accuracy of 75.1% on a dataset that contained 70000 patient records.

Barsasella et al., 2021 in their study, employed ML algorithms to predict the in-hospital Length of Stay (LoS) and mortality in patients having either T2DM or Hypertension (HTN). Insurance claim data of the hospitalized patients were used and the study even aimed to identify how comorbidities related to T2DM and/or HTN affected the LoS and mortality in these patients. LoS was best predicted by the Linear Regression model and mortality was best predicted by Multilayer Perceptron (MLP) model.

Hospital readmissions should be prevented as this challenge puts a financial burden and extra pressure on hospitals as well as the waste of medical resources. A systematic analysis conducted by Chen et al., 2022a. The EHR of T2DM patients suffering from chronic kidney disease (CKD) as well were used in this study (Novitski et al., 2022).

## 2.3 All-Cause Mortality Prediction for Patients with Type 2 Diabetes Mellitus

Determining the remaining life expectancy of a person having diabetes is a challenge as a large number of influencing factors that increase complications have to be taken into account. The development of an all-cause mortality prediction model can be very useful for health practitioners as it can identify risky patients and also determine the remaining life expectancy of their patients. Many models for mortality prediction have been developed in the past. Some of these models estimate the mortality risk for diabetes patients. Thus, this section provides information on the existing studies that have been conducted related to all-cause mortality in patients having T2DM.

### 2.3.1  Mortality Prediction on Clinical Trial Data vs Real-World Data

Randomized clinical trial data (CTD) consists of highly selective populations which are managed in tightly controlled settings (Kim et al., 2018). Whereas, real-world data (RWD) is the data collected from diversified areas of daily life that are outside the scope of highly controlled randomized control trials. These real practical data are collected broadly through the use of various medical devices by users and the medical data accumulated in Electronic Medical Records (EMRs) (Kim et al., 2018).

The method of modelling the progression of T2DM to predict the long-term outcomes of the disease is called computer simulation. The United Kingdom Prospective Diabetes Study (UKPDS) Outcomes Model (UKPDS-OM1) was first published by Clarke et al., 2004. This model has a variety of applications, some of these being the prediction of life expectancy and cost-effectiveness analyses but it was developed using clinical trial data collected until the year 1997 in the UKPDS. There was a need to update this model to include data related to new risks and outcomes.

Thus, the UK Prospective Diabetes Study Outcomes Model version 2 (UKPDS-OM2) was created as an upgrade to the version 1 model UKPDS-OM1. The follow-up data used was for 30 years, the initial 20 years from the original trial and then the survivors were entered into the 10-year post-trial monitoring. These extra 10 years were actually observational data out of clinical settings. This second version had significant advantages over the previous model, as it captured more outcomes, gave greater precision, and captured the progression of diabetes comprehensively (Hayes et al., 2013).

The UKPDS-OM2 equations were externally validated in a different study using two realworld European cohort data. The cohorts considered were the Italian Casale Monferrato Survey (CMS) and the Dutch Hoorn Diabetes Care System (DCS) to analyse outcomes such as all-cause mortality, congestive heart failure (CHF), ischaemic heart disease (IHD), myocardial infarction, and stroke (Pagano et al., 2021). But at follow-up, risk factor data were completely missing in both cohorts. The calculated biases showed that there was a consistent overestimation of 5-, 10-, and 15-year all-cause mortality for CMS and 5- and 10-year allcause mortality for the DCS cohort. Thus, the results of this study indicate that new or updated risk equations might be needed for other populations as the transferability of existing UKPDS-OM2 model equations is unsatisfactory (Pagano et al., 2021).

The Risk Equations for Complications of Type 2 Diabetes (RECODe) in (Basu et al., 2017) are equations that were developed and validated to predict risks due to diabetes complications. Separate equations were developed for each of the micro-vascular outcomes, cardiovascular outcomes, and all-cause mortality as well. These equations were developed as the existing equations were misestimating the risks of diabetes complications. The Action to Control Cardiovascular Risk in Diabetes (ACCORD) study data was used for the development and validation of these risk equations. The external validation was also done on the Diabetes Prevention Program Outcomes Study (DPPOS) data which had micro-vascular events. The external validation of cardiovascular risk equations was done on the Action for Health in Diabetes (Look AHEAD) data.

The predictor variables were demographic characteristics, biomarkers, clinical variables, medications, and comorbidities. Elastic net regularisation was used to select from the predictor variables for the Cox proportional hazards models. Though the 10-year risk estimation of the T2DM complications was done better by RECODe than the older risk equations of UKPDS-OM2/AHA pooled cohorts, the C-statistic value for the all-cause mortality equation is lower than other all-cause mortality equations tested on other populations in (Yang et al., 2008). Also, these RECODe equations may not have good generalizability as the development and validation of the equations are performed on clinical trial data (Basu et al., 2017).

Therefore, these RECODe equations were further externally validated for predicting microvascular and macro-vascular outcomes as well as all-cause mortality in T2DM patients among diverse populations (Basu et al., 2018). The risk predictions from these RECODe equations and the two alternative risk models UKPDS-OM2 and American College of Cardiology (ACC)/American Heart Association (AHA) Pooled Cohort Equations (PCEs) were compared to the observed outcomes in the Multi-Ethnic Study of Atherosclerosis (MESA) and the Jackson Heart Study (JHS) data (Basu et al., 2018). The micro-vascular and macro-vascular RECODe equations did give higher accuracy in 10-year risk estimation than the risk equations of UKPDS-OM2 and ACC/AHA PCE. C-statistics for the all-cause mortality equation was improved compared to that of (Basu et al., 2017). But the external validation is done on small datasets having limited samples.

A short-term 2-year mortality prediction model was developed along with a web-based mortality risk calculator in research. The Gargano Mortality Study (GMS) real-life data was used for training and the Foggia Mortality Study (FMS) real-life data was used for validating the Multivariate Cox proportional hazards regression model (De Cosmo et al., 2013). The continuous Net Reclassification Improvement (cNRI) driven forward variable selection technique selected 9 predictors for the final model. Though the aim was to develop a parsimonious model by considering demographics and easily available clinical biomarkers as variables, comorbidities need to be considered as they play a vital role in increasing mortality risks. The study samples used are relatively small and of Italian ancestry only thus raising questions of the generalizability of the model.

As an update to the previous 2-year model, the Estimation of Mortality Risk in Type 2 Diabetic Patients (ENFORCE) model is a well-established 6-year mortality risk prediction model trained on the GMS real-life samples with 9 final selected predictors that are commonly collected in clinical practice. The external validation is done on three independent samples FMS, Pisa Mortality Study (PMS) and ACCORD (Copetti et al., 2019). One of these validations on ACCORD clinical trial data shows that the model performance lowers on clinical trial settings data which was earlier seen in RECODe study as well. But in this study, the training and the other two validation samples used are restricted only to the Italy population and have a significantly low number of records. Thus there is a need for additional studies to address the model's transportability to different populations (Copetti et al., 2019).

The UKPDS model was originally developed using UK diabetes cohort data from the 1970s but a new risk model was needed to be developed on a US population. This was due to the significant difference in both the populations in terms of ethnicity, treatment algorithms and comorbidities assessments. Hence, the risk engine named Building, Relating, Assessing, and Validating Outcomes (BRAVO) was developed and internally validated on the ACCORD clinical trial data (Shao et al., 2018). This novel risk engine contained three separate modules such as events, risk factors, and mortality with each module having a number of regression equations to predict the occurrence of events, risk factors progression, and mortality (Shao et al., 2018).

BRAVO showed good prediction power of $R^2 = 0.86$ when externally validated on clinical trials data such as the Atorvastatin Study for Prevention of Coronary Heart Disease Endpoints in Non-Insulin-Dependent Diabetes Mellitus (ASPEN), the Action in Diabetes and Vascular Disease: Preterax and Diamicron Modified Release Controlled Evaluation (ADVANCE), and the Collaborative Atorvastatin Diabetes Study (CARDS) (Shao et al., 2018). This risk engine also found that slightly above 7% of glycosylated hemoglobin level is associated with the lowest risk for all-cause mortality (Shao et al., 2018).

### 2.3.2    Mortality Prediction on Western Population vs Asian Population Data

The UKPDS, BRAVO, ENFORCE, and RECODe models are based on Western populations. Asian populations develop diabetes at a younger age when compared with other populations and also have a higher tendency of developing from T2DM and also its complications (Gujral et al., 2013; Ma and Chan, 2013). These models often cannot be directly applied to Asian populations because of the differences in socioeconomic factors, genetics and approaches to diabetes management (Qi et al., 2022). Some studies have been conducted to specifically address this.

A study on the development of a predictive risk model for all-cause mortality in diabetic patients in Hong Kong was conducted by (Lee et al., 2021). Multivariate and score-based Cox proportional hazards models were developed using large-scale data consisting of more than 273000 patients taken from a healthcare database named the Clinical Data Analysis and Reporting System (CDARS). The patient data is managed by the Hong Kong Hospital Authority. These models were improved using Random Survival Forests (RSF) and Deep neural survival learning (DeepSurv) approach. The superiority of the RSF model is seen in this study as it gives the highest accuracy (AUC = 0.86 and C-statistic = 0.87) in prediction as well as good interpretability due to the 'importance ranking' of the predictor variables (Lee et al., 2021).

Large-scale data from the Chinese population aged 40 to 99 years with T2DM were considered in another study to develop a 5-year model for predicting mortality (Qi et al., 2022). The 28 baseline highly accessible predictors consisted of demographics, biochemical indicators, various comorbidities and medications. This risk model was internally validated using 100 bootstrap samples and showed good performance with a mean C-statistic of 0.8113 (Qi et al., 2022). But this model needs to be validated externally with other populations and longer follow-up periods should be considered to form 10-, 15-, and more year risk models as the age spectrum is large.

The UKPDS-OM2 was externally validated on Israeli population data taken from the Maccabi

Healthcare Services (MHS) (Zhuo et al., 2022). T2DM patients aged ≥ 26 years were considered in this study. For different baseline risks, the prediction performance was different between patients. Though the Area Under the Curve (AUC) is 0.86 for all-cause mortality, the Predicted to Observed ratio (P/O) (i.e., predicted to actual 15-year cumulative risk) is 1.32 thus highlighting that the model overpredicted all-cause mortality risk (Zhuo et al., 2022).

**2.3.3    Other Challenges Addressed in Mortality Prediction**

Diabetes and its complications reduce life expectancy. But there was uncertainty about whether mortality can be predicted for patients having T2DM. A study on the Hong Kong Diabetes Registry cohort data of 7583 T2DM patients concluded that mortality prediction can be done using a risk score model developed using common clinical and biochemical variables as predictors. The 5-year all-cause mortality risk score model had an Area Under the Receiver Operating Characteristic (AUROC) Curve of 0.85 (Yang et al., 2008). But this Cox proportional hazards regression model needs to be externally validated on an independent sample to test the validity of the predicted risk score on other population data.

A 6-year mortality risk prediction tool was developed using the Cleveland Clinic electronic health record (EHR) of patients aged at least 18 years and diagnosed with T2DM (Wells et al., 2008). The model was developed on a significantly large cohort using Cox proportional hazards and was internally validated with 10-fold cross-validation. 20 predictor variables and four oral medication classes such as biguanides (BIGs), sulfonylureas (SFUs), thiazolidinediones (TZDs), and meglitinides (MEGs) were considered (Wells et al., 2008). The interaction of each medication class with age, congestive heart failure (CHF) and glomerular filtration rate (GFR) were also investigated. The MEGs were associated with the highest risk of mortality and the BIGs with the lowest risk. Comorbidities considered were very few and the model requires external validation as well.

A 5-year mortality risk score prediction model has been developed using data from patients in New Zealand (NZ) (Robinson et al., 2015). The dataset (n=26864) for training the models was from the NZ Diabetes Cohort Study (NZDCS) and the dataset (n=7610) for testing the models was taken from the Diabetes Care Support Service (DCSS) (Robinson et al., 2015). These cohorts consisted of multi-ethnic patients. Three progressive models were developed as part of this study with the first using predictor variables as demographic information, the second by adding more clinical data and the third with renal disease markers. The best model turned out to be the renal disease model. One major drawback is that apart from CVD, no other comorbidities were considered predictors during model development.

Another study considered large-scale data from Taiwanese diabetic patients aged 65 years or older to develop a 5-year model for predicting mortality risk with 14 clinical predictors. The data used for development and internal validation was collected from Taiwan's Adult Health Screening Program (AHSP) and from the MJ Health Screening database for external validation (Chang et al., 2017). This elderly-specific model is significant as older adults are more prone to experience hypoglycaemia or even premature death (Chang et al., 2017). Harrelle's C-statistic was 0.737 and 0.746 in the development and internal validation datasets, respectively. But the C-statistic value of the model significantly decreased to 0.685 on the external validation dataset. Also, the comorbidities and medication details of patients were not considered predictors in this study.

Further, two gender-specific models for the prediction of 5-year all-cause mortality in Chinese patients were developed by (Wan et al., 2017). Patients aged between 18 and 79 years were considered but the patients having a history of cancer, end-stage renal disease, CVD, and chronic lung disease were excluded from the study. This is a major drawback as the models are only applicable to T2DM patients having no comorbidities. Harrelle's C-statistic was calculated by using 10-fold cross-validation (Wan et al., 2017). Though the study is done on a significantly large number of patient records, still external validation of these models is also required as the development and internal validation are done on the Chinese population only.

Again in another study, two clinical scoring systems were developed for risk estimation, one for all-cause mortality and the second for cardiovascular-specific mortality in T2DM patients (Liu et al., 2021). 9692 patients data with an average follow-up of 8.7 years is present in the retrospective cohort taken from the Diabetes Care Management Program (DCMP) of the China Medical University Hospital (CMUH) in Taiwan (Liu et al., 2021). Missing values were imputed with Multiple Imputation methods. The patient records were randomly divided into training and validation sets by a 2:1 ratio. This study is the first one to predict risk scores based on BP and glycaemic variability (Liu et al., 2021). Still, external validation is recommended as the data is restricted to the Chinese population and in-hospital data.

The all-cause mortality prediction models would be even better if the models could identify high-risk patients from the beginning of the diabetes disease. This challenge has also been addressed in a study. The well-established ENFORCE and RECODe models were successfully applied in the early stage of T2DM thus helping to address the reduction of life expectancy effectively (Copetti et al., 2021). The discovery dataset (n=302) consisted of patients aged 30 years or older who had the onset of T2DM in the last 6 months. The validation set (n=392) consisted of patients from the Italian sample where ENFORCE was updated (Copetti et al., 2021). But this study used small sample sets having only Italian patients.
The 6-year mortality prediction model ENFORCE performs as competently as the RECODe which is a 10-year mortality prediction model with 14 predictors (Copetti et al., 2021). The RECODe model has been even investigated with the population of different countries, under different settings, and validated by employing large samples as well (Copetti et al., 2019). But the baseline predictors used in ENFORCE and RECODe models do not include comorbidities.

A multi-year study has also been conducted on a sample of Taiwanese patients aged 18 years or older. 7-year and 10-year models have been developed using the hospital-based data from the Chang Gung Memorial Hospital in Keelung (CGMH-K) and the mortality data is taken from the nationwide death registry which helped identify all patients that died outside the hospital (Chiu et al., 2021). Patients with missing values were also considered by adopting the missing indicator method. Some comorbidities and drug treatments along with clinical biomarkers are considered as predictors (Chiu et al., 2021). But the hospital patient's data is specific to a city in Taiwan and thus the model prediction may differ among cities and countries. Also, behavioural factors such as alcohol consumption, exercise, and cigarette smoking, are not present in the data.

Regular exercise and a diet rich in fresh fruit play an important role in the prevention as well as management of diabetes. Therefore, a 5-year mortality risk prediction model was developed using the data of 20340 participants from the Comprehensive Research on the Prevention and Control of the Diabetes (CRPCD) project (Chen et al., 2022b).

Now, the dataset considered for the current research work has previously been used in only one research in which two point-based mortality prediction models were developed using Logistic Least Absolute Shrinkage and Selection Operator (LASSO) regressions (Griffith et al., 2020). Separate models for 5-year and 10-year mortality were developed by Griffith et al., 2020 as there are 2 binary target variables in that dataset. The study sample taken is large as it includes 275,190 US veterans aged 65 years or older and having T2DM. In this dataset, there are 68 baseline predictors related to the demographics, comorbidities, medicine prescriptions and various other details. The study identified 37 predictors of mortality out of which 22 were comorbidities or procedures.

The LASSO technique was employed to select the predictors that are independently associated with mortality risk (Griffith et al., 2020). This was done using the training set which had 75% instances of the entire dataset. The test set had the remaining 25% of instances of the dataset. Further, 10-fold cross-validation was also used as this re-sampling procedure helps reduce overfitting problems. Point-based risk scoring systems were developed using Logistic LASSO regression which on the basis of the resulting odds ratios, subtracted 1 from each of the odds ratios and then divided it by 0.2 (Griffith et al., 2020). Good discrimination was shown by the prognostic indices which were proved by the Cstatistics values of 0.74 for the 5-year prediction model and 0.76 for the 10-year prediction model. The balanced accuracy obtained was 0.78 in the 5-year prediction model and 0.77 in the 10-year prediction model (Griffith et al., 2020).

Though the performances of these 5- and 10-year mortality prediction models are acceptable, there is a limitation stated by this research. To get better predictive accuracy, this methodology allows bias in the odds ratios (Griffith et al., 2020). The use of other feature selection techniques may perform better than the LASSO which is an embedded feature selection technique. Apart from this, Griffith et al., 2020 have explored only Logistic LASSO Regression in their study. Thus, a new methodology using other machine learning algorithms needs to be attempted on this mortality dataset and see if multiclass classification is possible.

## 2.4 Feature Selection Techniques for Medical Datasets

Machine Learning algorithms need to be trained only with the most relevant features of a dataset as some features give no contribution to improving the ML algorithm and the presence of such irrelevant features might worsen model performance. Thus, Feature Selection (FS) techniques are very important to find the features that would improve a model's performance. These techniques prevent overfitting, speed up the training process and give useful insights about the data as well. The importance of FS techniques increases even more if the dataset being dealt with is wide and has a high number of features. There are mainly 3 types of FS techniques - filters, wrappers and embedded methods (Cascaro et al., 2019; Ramos-Pérez et al., 2022; Sharma and Mandal, 2023).

Filter methods are used to calculate the relevance of each feature, mainly based on its statistical properties, providing a numerical score for each feature that depends on its contribution to the performance of the algorithm, also called importance (Ramos-Pérez et al., 2022). Figure 2.1 shows the hierarchical representation of the feature selection algorithms.

Agaal and Essgaer, 2022 studied various filters, wrappers, and embedded FS techniques were examined in this study including Generic Univariate, Select K Best, Select Percentile, Mutual Information (MI), Pearson correlation coefficient (PCC), ReliefF, Recursive Feature Elimination (RFE), Recursive Feature Elimination with Cross-Validation (RFECV).

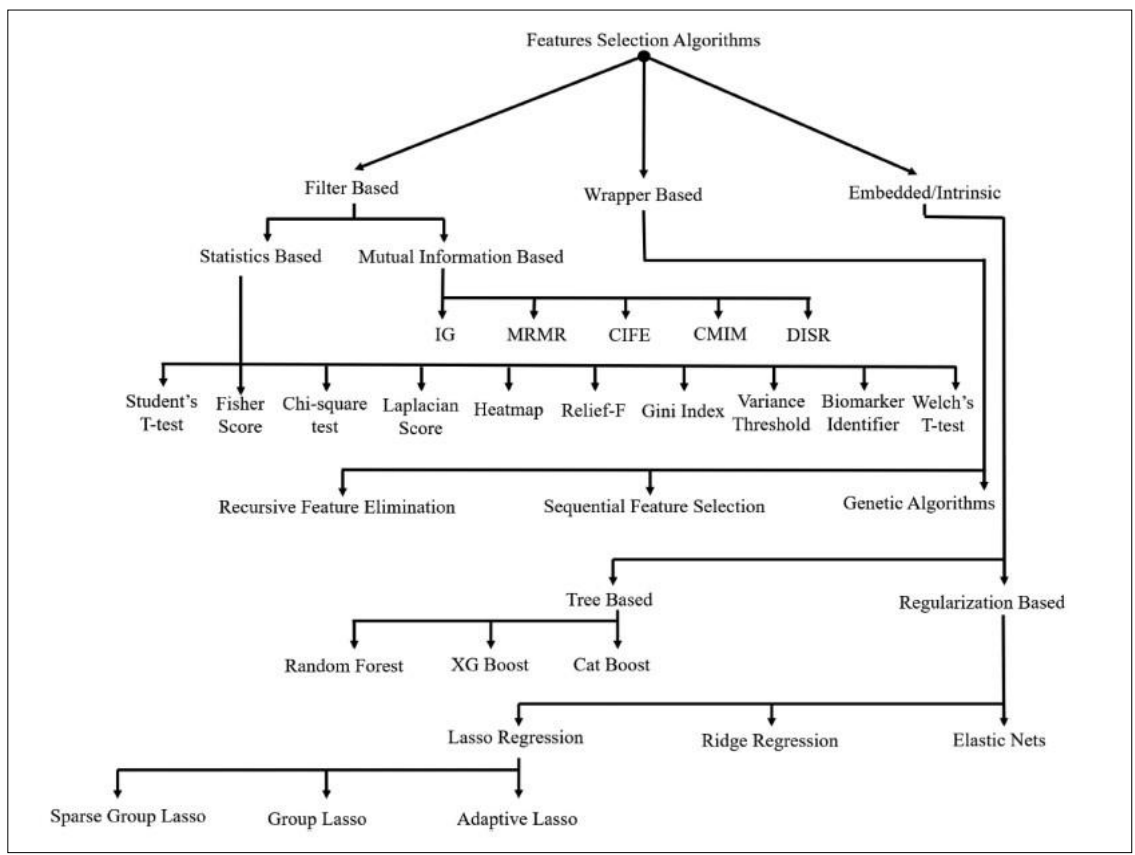

Figure 2.1 A hierarchical representation of the feature selection algorithms (Image taken from Sharma and Mandal, 2023)

## 2.5 Multiclass Classification Machine Learning Algorithms for Healthcare

Machine Learning (ML) algorithms are used to train models using data to extract information and identify patterns present in the data. As seen earlier, ML plays a very important role in the healthcare industry. Supervised Machine Learning techniques are used to predict outcomes by learning, identifying relationships and making decisions from labelled data. These techniques are widely used for classification problems where the target variable is categorical (Malik et al., 2022). Now, if the target variable has two classes it is called binary classification and if the number of classes is more than two then it is deemed as a multiclass classification problem. Some of the healthcare industry problems that have used multiclass classification ML algorithms have been explored in this section.

Biswas et al., 2021 in their research work aimed to provide a ML model to predict personalized medicines for cancer treatment using the genome data of patients. The study concluded Random Forest (RF) as the best prediction model compared to other models that were analyzed such as the Logistic Regression (LR), Support Vector Machine (SVM), Naïve Bayes (NB), K-Nearest Neighbors (K-NN), and Decision Tree (DT) (Biswas et al., 2021). The KNN model performed the worst among these models.

Bárcenas and Fuentes-García, 2022 implemented 3 ML algorithms to assess the patients with COVID-19 risks and to identify the relevant features associated with each COVID-19 severity. The three risk scenarios considered were low (treatment and monitoring), moderate (recovered and non-severe), and high (severe and deaths) (Bárcenas and Fuentes-García, 2022). Extreme Gradient Boosting (XGBoost) with an overall accuracy of 89.97% was the best model compared to RF and Gradient Boosting Machine (GBM). El-Sappagh et al., 2022a built an ensemble model that predicts the progression of Alzheimer's disease 2.5 years in the future. The ensemble consisted of three XGBoost classifiers as the base and the Logistic Regression (LR) as Meta classifier achieving an accuracy of 89.15% (El-Sappagh et al., 2022a).

Alyas et al., 2022 in their study employed various ML algorithms to predict thyroid disease. The algorithms used were KNN, DT, RF, and Artificial Neural Network (ANN) out of which the Random Forest (RF) gave the highest performance with an accuracy of 94.8%. In another study, Guleria et al., 2022 explored a deep learning-based model Artificial Neural Network (ANN) in addition to ML models such as DT, RF, NB and multiclass classifiers. This study was for the early prediction of hypothyroidism and the DT and RF models gave the highest accuracy of 99.5758% and 99.3107% respectively (Guleria et al., 2022).

Abdellatif et al., 2022 proposed a model that would determine the current status of a patient's heart and help in preventing mortality due to heart disease if early treatment is implemented. In the first part of this study, the absence or presence of CVD was predicted using a binary classification approach. Then in the second part, a multiclass classification approach was used to predict the CVD severity level. LR, SGD, Extra Trees (ET), XGBoost, KNN, and SVM were the classifiers analyzed for multiclass classification.

The proposed model integrated the Synthetic Minority Oversampling Technique (SMOTE) for data balancing, hyper-band (HB) for hyperparameter optimization, and ET for classification as ET gave the highest accuracy among all the classifiers analyzed (Abdellatif et al., 2022).

## 2.6    Discussion

Predictive Analytics uses historical medical data of patients to analyze and identify certain trends and patterns present in the vast amount of data which can help in predicting healthrelated events that may occur in the future. In the past few years, researchers have used predictive analytics using ML techniques majorly for diagnosis and prognosis of various diseases, mortality prediction, predicting in-hospital length of stay and mortality, and prediction of hospital readmissions.

The research works related to the all-cause mortality prediction for patients having T2DM have been discussed in section 2.3 and a summarization of these studies is added in Appendix A: Section 2 - Summary of related works. Most of the models are developed using the Cox Proportional Hazards regression. The reason is that in these studies, the target outcome variable suitable is for regression as this target variable contains the number of days till the mortality event occurred. Though the models have performed well in all the mentioned studies, still the impact of clinical trial data and real-world data, the effect of comorbidities and prescriptions, the transportability of models to different age groups, genetics, and ethnicity, and consideration of external validation; these factors need to be explored more and preferably with large-scale data.

Now, let's discuss the mortality dataset which is of interest for this current research work. This dataset has previously been used in only one research conducted by Griffith et al., 2020 in which separate point-based mortality prediction models were developed for 5-year and 10year mortality in older adults having T2DM. The dataset is large as it includes 275,190 US veterans aged 65 years or older and having T2DM. There are 68 potential predictors of mortality related to demographics, comorbidities, medicine prescriptions and various other details. There is a limitation stated in this research that to get better predictive accuracy, this methodology allows bias in the odds ratios (Griffith et al., 2020). Also, Griffith et al., 2020 have explored Logistic LASSO Regression only.

Table 2.1 summarizes some of the works related to FS techniques used on medical datasets that were discussed in the literature mentioned in section 2.4. The filter-based FS techniques are independent of the machine learning classifier used and are simple and fast compared to the embedded and wrapper-based ones (Zhang et al., 2022). Filter-based feature selection techniques have been commonly employed by researchers in the medical field and help to give good model performance. Therefore, filter-based FS techniques such as Chi-Squared or Information Gain may be better than the embedded LASSO FS technique.

Table 2.1 Summary of some recent works related to feature selection techniques used on medical data

| Citation | Main Work | FS Techniques Analysed | Best FS Technique |
|---|---|---|---|
| (Chong et al., 2021) | To identify the appropriate feature subsets for physical activity class prediction | 7 filters - CFS, gain ratio, symmetrical uncertainty, IG, oneR, ReliefF, and MRMR<br><br>2 wrappers - SFS and best first | With ANN - symmetrical uncertainty, gain ratio, and IG<br><br>With SVM - symmetrical uncertainty, IG, and gain ratio<br><br>With RF - gain ratio, MRMR and CFS |
| (Barsasella et al., 2021) | Prediction of length of stay and mortality for hospitalized patients having T2DM and hypertension | Filter - ReliefF (for predicting Length of Stay)<br><br>Filter – Information Gain (for predicting mortality) | ReliefF<br><br>Information Gain |
| (Reddy et al., 2021) | ML model to predict heart disease risk | Filters - ReliefF, correlation-based, and Chi-squared | Chi-squared |
| (MorilloVelepucha et al., 2022) | ML model to predict congestive heart failure risk | Filters - Mutual Information and Chi-square | Chi-squared |
| (El-Sappagh et al., 2022a) | Automatic detection of Alzheimer's disease progression | Filter – Information Gain | Information Gain |
| (El-Sappagh et al., 2022b) | Two-stage deep learning model for Alzheimer's disease detection and prediction | Some filter-based and wrapperbased techniques | Filter-based |
| (Agaal and Essgaer, 2022) | Influence of FS methods on early prediction of breast cancer | 6 filters - Select K Best, Select Percentile, Generic Univariate, PCC, MI, and ReliefF<br><br>4 wrappers - RFE, RFECV, SFS, SBS | Top 3 among all –<br>• RFECV<br>• Select-From-Model<br>• MI |

|  |  | 1 embedded - Select-From-Model |  |
| --- | --- | --- | --- |
| **Citation** | **Main Work** | **FS Techniques Analysed** | **Best FS Technique** |
| (EbiaredohMienye et al., 2022) | A ML method with filter-based FS to improve prediction of CKD | Filter – IG | IG |
| (Bashir et al., 2022) | FS for classification of medical data | Filters, wrappers, and embedded FS techniques | Top 3 among the filters – ReliefF, IG, Correlationbased feature selection |

The dataset used by Griffith et al., 2020 can be used for mortality prediction in a different way compared to the original research. In this dataset, the records of patients shown dead at the 5year mark are also marked dead at the 10-year mark. If a 10-year mortality prediction model is built using binary classification, it cannot determine whether the patient would die within the next 5 years. It can just predict whether the person will die in the next 10 years or not. A separate 5-year mortality prediction model would be required for it. But this challenge can be addressed if the binary target variables are converted and made applicable for multiclass classification. Thus, just one multiclass classification model can predict whether the patient's remaining life is "up to 5 years", "more than 5 but up to 10 years" or "more than 10 years".

Now, table 2.2 depicts a summary of multiclass classification ML algorithms employed in various recent research works related to the healthcare industry. Several ML classifiers have been used for multiclass classifications and ensembles such as RF and XGBoost have outperformed other ML classifiers in most of the studies. Therefore, a new methodology using such supervised machine learning algorithms for multiclass classification can be tried on this mortality dataset to see how the new methodology performs.

In past studies, filter-based feature selection techniques and multiclass classification approaches has never been applied to all-cause mortality prediction for T2DM patients. They have been used in various other healthcare industry problems though. Multiclass classification using ML algorithms has never been done on the mortality dataset of interest which contains large number of patients' records and 68 potential predictors. Such several potential predictors have never been considered in any previous study related to all-cause mortality prediction for T2DM patients.

Table 2.2 Summary of some recent works related to multiclass classification ML algorithms employed in the healthcare industry

| Citation | Main Work | ML Techniques used | Best ML Technique |
|---|---|---|---|
| (Biswas et al., 2021) | Prediction of personalized medicines using genome data for cancer treatment | LR, SVM, NB, KNN, DT, and RF | RF (Accuracy: 69%) |
| (Bárcenas and FuentesGarcía, 2022) | Risk assessment in COVID-19 patients using multiclass classification | RF, GBM, and XGBoost | XGBoost (Accuracy: 89.97%) |
| (Alyas et al., 2022) | ML model for classification of thyroid disease | ANN, RF, DT, and KNN | RF (Accuracy: 94.8%) |
| (Guleria et al., 2022) | Multiclass classification predictive ML and DL for early prediction of hypothyroidism | NB, RF, DT, Multiclass, and ANN | DT (Accuracy: 99.5758%) RF (Accuracy: 99.3107%) |
| (El-Sappagh et al., 2022a) | Automatic detection of Alzheimer's disease progression | Ensemble (3 XGBoost classifiers as base + LR as meta classifier) | Ensemble (Accuracy: 89.15 %) |
| (Abdellatif et al., 2022) | ML model to predict presence of CVD and its severity level | LR, SGD, ET, XGBoost, KNN, and SVM | ET (Accuracy: 84.53%) XGBoost (Accuracy: 84.26%) Proposed SMOTE+ET+HB (Accuracy: 95.73%) |

Motivated by these discussions, the current research aims to propose a multiclass classification methodology for mortality prediction in older adults aged 65 years and older who have Type 2 Diabetes Mellitus (T2DM). This new methodology consists of a filter-based feature selection technique and the best-performing algorithm among the ML classifiers and ensembles. Thus, the proposed methodology is novel and contributes to the existing research conducted by Griffith et al., 2020 who built separate 5-year and 10-year mortality risk prediction models on this dataset. The novelty of the proposed methodology is in taking the multiclass classification approach and building just one model.

**2.7 Summary of the literature review**

This section presented some of the recent studies carried out regarding the relevance of Machine Learning and predictive analytics in healthcare. Further, various works of literature about all-cause mortality in patients having T2DM were discussed and summarized. It was seen that most studies have employed the Cox Proportional Hazards Regression models for mortality prediction. The literature on all-cause mortality prediction concluded that the impact of clinical trial data and real-world data, the inclusion of comorbidities and prescriptions, the transportability of models to different age groups, genetics and cultural backgrounds, and consideration of external validation; these are still needed to be explored more and preferably with large-scale data.

The previous research conducted on the mortality dataset used in the current research has used only the Logistic LASSO Regressions to predict 5-year and 10-year mortality in older patients with T2DM. Therefore, some recent studies related to feature selection techniques used for medical datasets were discussed which concluded that various filter-based feature selection techniques have been used by researchers. These are deemed better than wrapper-based or embedded ones. In the end, some recent research works regarding the use of ML algorithms for multiclass classification in healthcare industry problems were also seen. Ensembles such as Random Forest and XGBoost have given high performance.

**3. RESEARCH METHODOLOGY**

This section sheds light on the research approach taken for this study. All the technical jargons that are used throughout the study have been mentioned and are supported with theoretical details. Each step in the research methodology is explained in detail. The sections in this section focus on details such as the data selection and its description, data preprocessing, process of data transformation, feature selection techniques, cross-validation, machine learning techniques employed for modelling, and lastly the evaluation metrics for assessing the performance of various models.

Different feature selection techniques such as the embedded-based LASSO and filter-based Information Gain and Chi-Squared, have been briefed. Various supervised machine learning algorithms employed for multiclass classification have been explained by theoretical concepts. First, a benchmark model consisting of Multinomial Logistic Regression with LASSO regularization is explained. This model will perform multiclass classification using regression. The remaining models have used classifiers. Multinomial Logistic Regression is a non-ensemble classifier that has been considered. In ensembles, Random Forest and XGBoost have been elaborated. Another classifier which can be considered as an ensemble is the OneVs-Rest Classifier and it has also been explained.

## 3.1	Research Approach

The methodology used in this study involves various processes such as target data selection, data pre-processing, data transformation, feature selection, employing various supervised machine learning techniques, and finally, evaluating the performance of each machine learning model using evaluation metrics. By performing these steps, various insights can be driven out from the dataset that has been used. By the end of this study, a prediction model is developed that performs multiclass classification and predicts the mortality risk in patients having T2DM. The process flow executed to achieve the objectives is depicted in Figure 3.1.

As shown in Figure 3.1, there are several steps and processes involved in the development of a mortality risk prediction model for T2DM patients. The study has two main parts. The first part is to build a benchmark model with Multinomial Logistic Regression using feature selection as the LASSO technique. The second part is to build various supervised multiclass classification models and ensemble models using filter-based feature selection techniques.

Each process has been explained in the following sections.

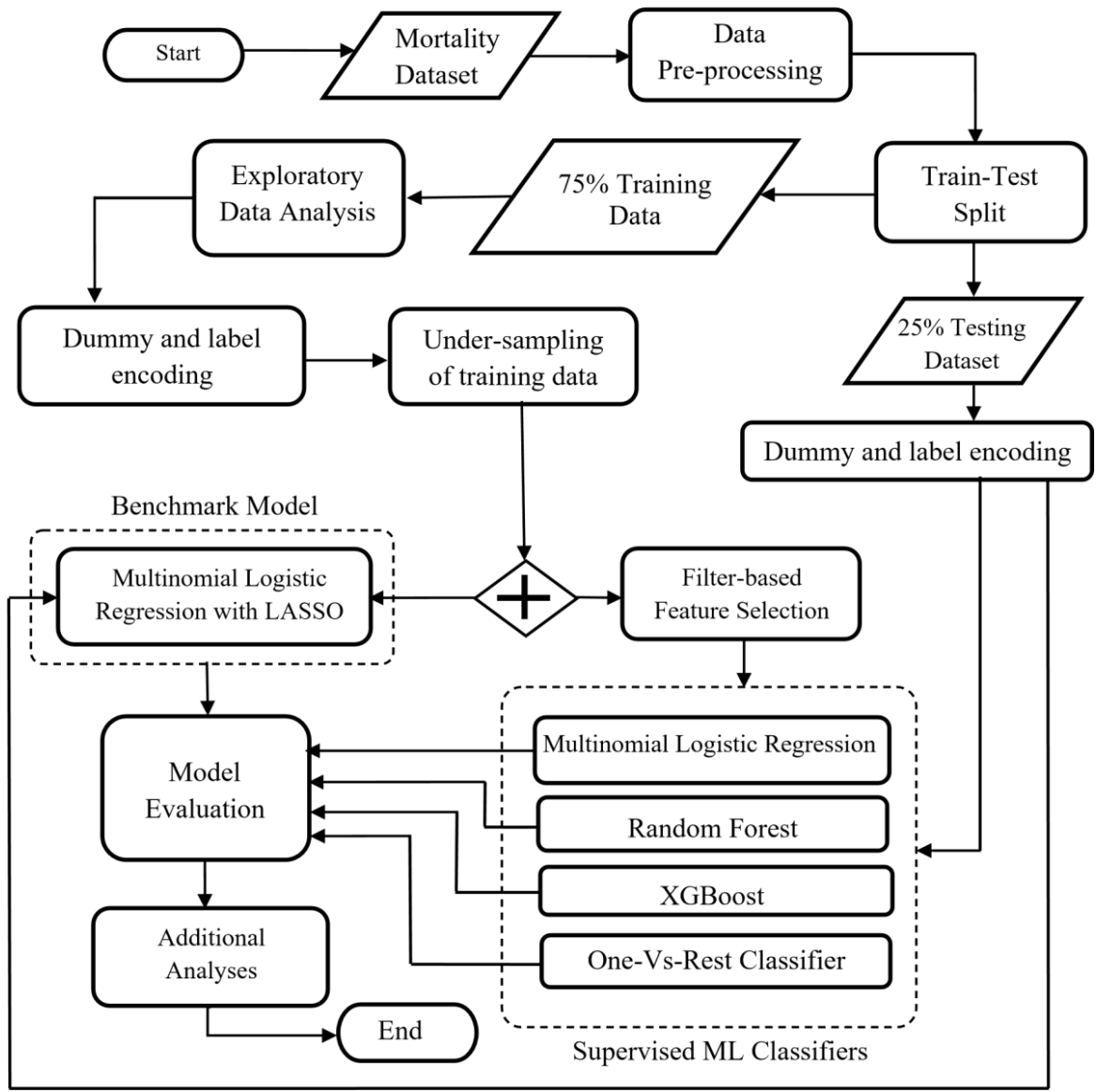

Figure 3.1 Research approach taken for this study

### 3.1.1  Data Selection and Description

The data considered for this study to build an all-cause mortality prediction model has been selected as it contains comorbidities details along with the common demographics and regular clinical biomarkers. The dataset considered is taken from a CSV file publicly accessible through the Mendeley Data Repository (Griffith, 2021). As it is licensed under a Creative Commons (CC) Attribution 4.0 International license, data can be copied, modified and shared as long as appropriate credit and dataset link has been given.

This dataset includes real-world samples of U.S. Military Veterans who were aged 65 years or older, enrolled in the Veterans Health Administration (VHA) as well as traditional Medicare, and had a previous diagnosis of diabetes (Type 2 Diabetes Mellitus) as of December 2005 (Vaidya et al., 2022). This entire data is originally accessed from the VHA CDW (VHA Corporate Data Warehouse) (Vaidya et al., 2022) from the year 2004 to 2016 and then deidentified by researchers before making it publicly available.

The originally extracted RAW data has 907,507 records of patients with one inpatient visit or two outpatient visits with the ICD-9 diabetes code or having a diabetes medication prescription but excluding metformin-only, in the year 2004 to 2005 (Vaidya et al., 2022). Then, the final sample of 275,190 records is obtained by limiting the patients aged 65 or more and alive as of 31 December 2005; with Medicare enrolment; during the year 2004-2005 were enrolled in traditional Medicare with no Medicare advantage; and lastly had a minimum of one primary care visit with routine biomarkers records like body mass index, haemoglobin A1c and blood pressure (Vaidya et al., 2022).

In the final dataset (i.e., the CSV file) taken for this study, there are 275,190 rows and 70 columns out of which 68 columns are potential predictor features listed in Table 3.1 and the remaining 2 are binary target variables. The potential predictors of mortality are patient demographics, medication history, comorbidities, laboratory values and anthropomorphic measurements, procedure codes, and previous health service utilization (Griffith et al., 2020). The dataset details and data dictionary are available in this research paper (Vaidya et al., 2022). Lastly, there are 2 target variables (dependent variables):

- DEATH_5 – if alive as of December 31, 2011 then the value is 0, otherwise 1
- DEATH_10 – if alive as of December 31, 2016 then the value is 0, otherwise 1

Table 3.1 Potential predictors considered in the dataset

| # | Category | Attributes | Variable Name |
|---|---|---|---|
| 1 | Demographics | Age (in years) | AGE |
| 2 | | Gender | SEX |
| 3 | | Marital Status | MARRIED |
| 4 | | Race | RACE |
| 5 | | Enrolment priority | PRIORITY |
| 6 | Clinical Biomarkers | Body Mass Index (BMI) | BMI |
| 7 | | Diastolic blood pressure | DIASTOLIC |
| 8 | | Systolic blood pressure | SYSTOLIC |
| 9 | | HDL Cholesterol | HDL |
| 10 | | LDL Cholesterol | LDL |
| 11 | | Hemoglobin A1c | A1C |
| 12 | | Triglycerides | TRI |
| 13 | | Urine microalbumin | MICROALB |
| 14 | | Serum albumin | SERUMALB |

| # | Category | Attributes | Variable Name |
|---|---|---|---|
| 15 | | Serum creatinine | SERUMCRE |
| 16 | **Prescriptions of Diabetes Medication** | Alpha-glucosidase inhibitor prescription | ALPHA |
| 17 | | Biguanide prescription | BIGUAN |
| 18 | | Insulin prescription | INSULIN |
| 19 | | Sulfonylurea prescription | SULF |
| 20 | | Thiazolidinedione prescription | TZD |
| 21 | | Blood Pressure prescription | BP_RX |
| 22 | | Other diabetes medication | OTHER_MED |
| 23 | **Comorbidities** | Acute myocardial infarction | AMI |
| 24 | | AIDS/HIV | HIV |
| 25 | | Alcohol abuse | ALCOHOL |
| 26 | | Ankle-brachial index | ABI |
| 27 | | Blood loss anemia | BLOODLOSS |
| 28 | | Cardiac arrhythmias | ARRHYTHMIA |
| 29 | | Chronic pulmonary disease | PULMONARY |

| # | Category | Attributes | Variable Name |
|---|---|---|---|
| 30 | **Comorbidities** | Coagulopathy | COAG |
| 31 | | Congestive heart failure | CHF |
| 32 | | Coronary arterial disease | CAD |
| 33 | | Coronary Artery Bypass Grafting | CABG |
| 34 | | Deficiency anemia | ANEMIA |
| 35 | | Depression | DEPRESSION |
| 36 | | Diabetes, complicated | DMCX |
| 37 | | Diabetic foot infections | FEET |
| 38 | | Drug abuse | DRUGS |
| 39 | | End-stage liver disease for cirrhosis or alcoholic fatty liver | ESLD |
| 40 | | Fluid & electrolyte disorders | FLUIDSLYTES |
| 41 | | Hyperglycemia | HYPERG |
| 42 | | Hypertension, complicated | HTNCX |

| #  | Category | Attributes | Variable Name |
|----|----------|------------|---------------|
| 43 |          | Hypertension, uncomplicated | HTN |
| 44 |          | Hypothyroidism | HYPOTHYROID |
| 45 |          | Liver disease | LIVER |
| 46 |          | Lower limb amputation | AMPUTATION |
| 47 |          | Lymphoma | LYMPHOMA |
| 48 |          | Metastatic cancer | METS |
| 49 |          | Obesity | OBESITY |
| 50 |          | Other neurological disorders | NEUROOTHER |
| 51 |          | Paralysis | PARALYSIS |
| 52 |          | Peptic ulcer disease excluding bleeding | PUD |
| 53 |          | Percutaneous coronary interventions | PCI |
| 54 |          | Peripheral vascular disorders | PVD |
| 55 |          | Psychoses | PSYCHOSES |
| 56 |          | Pulmonary circulation disorders | PHTN |
| 57 |          | Renal failure | RENAL |
| **#** | **Category** | **Attributes** | **Variable Name** |
| 58 | **Comorbidities** | Retinopathy | RETINOPATHY |
| 59 |          | Retinopathy screening | RETSCREEN |
| 60 |          | Rheumatoid arthritis/collagen vascular diseases | RHEUMATIC |
| 61 |          | Severe depression | SEVERE_DEP |
| 62 |          | Smoking | SMOKER |
| 63 |          | Solid tumor without metastasis | TUMOR |
| 64 |          | Valvular disease | VALVULAR |
| 65 |          | Weight loss | WEIGHTLOSS |
| 66 | **Others** | Frailty Index | FRAILTY |
| 67 |          | Inpatient days | N_IP |
| 68 |          | Outpatient visits | N_OP |

Table 3.1 lists all the 68 potential predictor variables/attributes present in the dataset. All of these variables have been considered in this study. These variables have been segregated into categories such as Demographics, Clinical Biomarkers, Prescriptions of Diabetes Medication, Comorbidities, and Others. Frailty Index is a variable that indicates whether the patient has 30 deficits in health (Vaidya et al., 2022).

### 3.1.2 Data Pre-processing

Data pre-processing is an important part of the data mining process. It consists of preparing data before giving it to the ML classifiers. RAW data is converted to a desirable form to derive useful insights from it. The data is cleaned by taking appropriate measures for handling missing and noisy data. These steps are necessary as the data is used to train the models so that they can predict well on the unseen data after learning from the input data. Originally, the dataset had separate 5-year and 10-year Mortality outcome variables (DEATH_5 and DEATH_10). As this study is about multiclass classification, these two target variables are combined to form one target variable (Mortality) and the original DEATH_5 and DEATH_10 columns are dropped. Table 3.2 depicts the scenarios for each possible combination of values in the 'DEATH_5' and 'DEATH_10' target columns.

Table 3.2 Possible scenarios after converting the problem to multiclass classification

| Scenario | DEATH_5 | DEATH_10 | Mortality |
|---|---|---|---|
| The remaining life is up to 5 years | 1 | 1 | 11 |
| The remaining life is more than 5 but up to 10 years | 0 | 1 | 01 |
| The remaining life is more than 10 years | 0 | 0 | 00 |

Combination of 'DEATH_5' as 1 and 'DEATH_10' as 0 is impossible because a person dead at the 5-year mark cannot be alive at the 10-year mark. As seen in table 3.3, there are only 3 classes. The mortality value "11" is labelled as 'Class 1', "01" as 'Class 2', and "00" as 'Class 3'.

Table 3.3 Classes in new Target Variable (Mortality)

| Mortality | Labels |
|---|---|
| 11 | Class 1 |
| 01 | Class 2 |
| 00 | Class 3 |

Now, medical data often has missing values. The missing values in this dataset have been labelled as 'Missing' to adopt the missing indicator method. Keeping the missing values seems to be appropriate as imputing the null values with mean/median/zeros can make the medical data reporting incorrect and deleting the data can lead to loss of data. Therefore, to avoid these issues, missing values are considered as a separate category 'Missing'. Also, adopting the missing-indicator method allows the inclusion of such patients for the complete data analysis and it may further help in capturing compliance or health awareness (Chiu et al., 2021).

### 3.1.3 Data Transformation

The numeric columns in the dataset have several distinct values. There is a need to reduce the high cardinality of continuous data. The binning method is used to convert values of numeric columns into categories. It groups several continuous values into a smaller number of "bins" or intervals (Mahboob Alam et al., 2019). Binning or discretization can help handle outliers (Emam et al., 2023). As a model requires a number as input, there is a need to encode the categorical features. One-Hot Encoding is a process by which every unique value in the categorical variable is added as a feature variable in the dataset (Mukherjee et al., 2022). Category features are then represented in terms of the binary value of 0 or 1 (Meng and Xing, 2022). But the disadvantage of one-hot encoding is it leads to a dummy variable trap. Another technique called dummy encoding avoids this dummy variable trap as one of the dummy variables created for every feature is dropped (Mukherjee et al., 2022).

### 3.1.4 Feature Selection

The irrelevant features tend to cause overfitting and reduce the performance of the models (Cascaro et al., 2019). The feature selection (FS) process identifies the most important features and eliminates the irrelevant as well as the redundant ones. As seen in the body of literature, the three main types of FS techniques are filter, wrapper and embedded (Cascaro et al., 2019). Now, the performance of the model depends considerably on both the FS approach and the classifier but merely a change of FS technique used in the classification model impacts the model performance considerably (Sharma and Mandal, 2023). Therefore, feature selection is required.

Filter-based feature selection methods have an advantage over the wrapper and embedded methods as the filter-based methods are ML classifier-independent. Filter-based methods have low computational costs and rank features according to the relevance of that feature to the class (Cascaro et al., 2019). In this study, the filter feature selection methods used with the ML algorithms mentioned in section 3.2.2 are Information Gain (IG) and Chi-Squared Statistics. These two techniques have given good results in previous studies as seen in the body of literature.

**3.1.4.1 Least Absolute Shrinkage and Selection Operator (LASSO)**

The Least Absolute Shrinkage and Selection Operator (LASSO) also called L1 regularization is an embedded feature selection technique. LASSO selects a subset from a pool of variables that removes any possible collinearity and minimizes error in prediction (Griffith et al., 2020). The β-coefficients least associated with target outcome shrink towards 0 compared to the other β-coefficients that are more strongly associated (Griffith et al., 2020). In the case of highly correlated variables, the β-coefficients of the least associated ones shrink until it drops out of the model. In predictive research applications, it is an advantage if unbiasedness is traded off for lower variance and this is achievable through the LASSO technique (Griffith et al., 2020).

The shrinking parameter is λ on which the LASSO technique is dependent on (Pankajavalli and Karthick, 2022). Thus, the noisy features that are to be ignored are assigned a value of zero and the features with a non-zero coefficient are further chosen to be part of the model (Nibbering and Hastie, 2022). With the elimination of irrelevant features, the prediction errors are reduced which results in the prediction accuracy to be increased. LASSO feature selection technique has been employed with multiclass classification ML algorithms in various medical research works (Pankajavalli and Karthick, 2022; Joo et al., 2023).

### 3.1.4.2 Information Gain (IG)

Information Gain (IG) or Info-Gain, is a measure that is commonly used in the decision trees classifiers to find the most informative feature to be used for splitting each node (RamosPérez et al., 2022). The IG measures the importance of each feature separately. The gain of each feature is evaluated with respect to the outcome feature, meaning as per their relevance to the outcome variable (Ebiaredoh-Mienye et al., 2022). Features that have the least IG score are removed and if the value of IG is more than a threshold then that feature is considered relevant and employed while training the model. This technique is basically able to calculate the ability of a predictor variable to classify an outcome variable. The statistical dependence between two variables is calculated by IG and it can be formulated as shown in eq 3.1.

$$IG(X|Y) = H(X) - H(X|Y), \quad \text{……………. (Eq. 3.1)}$$

where the entropy for variable X is H(X) and the representation of the conditional entropy for X given Y is done by H(X|Y) (Ebiaredoh-Mienye et al., 2022). For computing a predictor feature's IG value, the target variable's entropy is ids calculated for the whole dataset and then the conditional entropy H(X|Y) of every potential value of the predictor variable is subtracted (Ebiaredoh-Mienye et al., 2022).

$$H(X) = - \sum_{x \in X} P(x) log_2(x), \quad \text{……………. (Eq. 3.2)}$$

$$H(X|Y) = - \sum_{x \in X} P(x) \sum_{y \in Y} P(x|y) log_2(P(x|y)), \quad \text{…………….. (Eq. 3.3)}$$

So, consider *X* and *Z* as the two variables. The variable *Y* is more significantly correlated to the variable *X* than the variable *Z* if *IG(X|Y) > IG(Z|Y)* (Ebiaredoh-Mienye et al., 2022). Information Gain FS technique has been used for selecting relevant features from various medical datasets. This technique was also used to obtain the best features for mortality prediction (Barsasella et al., 2021).

### 3.1.4.3 Chi-Squared Statistical Test

The Chi-square ($\chi^2$) test is a non-parametric powerful statistical tool for testing hypotheses between two categorical variables (Mchugh, 2013). The relationship between two categorical variables can be checked as the Chi-square test of independence determines whether there is a statistically significant association between two categorical variables (Hess and Hess, 2017). Traditionally, the significance boundary is set by a p-value of less than 0.05 (Hess and Hess, 2017). If the p-value is less than 0.05 (5% level of significance) then there is an association between the variables compared (both the variables are dependent). If the p-value is not less than 0.05 then there is no association between the variables that are compared (both the variables are independent). Thus, when the chi-squared statistic test is used as a feature selection technique, it helps to find the most relevant features that can contribute to the model's performance.

### 3.1.5   Train-Test Split

ML models are trained and tested using a dataset. As per the name, the training dataset is used to train the ML models. The test dataset is used to evaluate the performance of the ML models on unseen data. The train-test split technique is used by which the entire dataset is randomly divided into a train set and a hold-out test set. 75% of the samples are used for training the models and 25% of the samples are for testing the models. The distribution of instances in each class is uneven. So, to ensure that the proportion of the classes is maintained in the train as well as the test set, the stratify parameter is also considered during the 75% train and 25% test split.

### 3.1.6   Class Balancing

The distribution of each class in the target variable may vary in the dataset. To balance them out, class balancing techniques can be used. When the number of samples in one class is significantly higher compared to other classes, the model trained on such data tends to be biased towards the majority class. This leads to poor prediction performance of the minority class. The distribution of classes can be re-balanced by over-sampling or under-sampling strategy (Martins et al., 2023).

In the case of oversampling, the minority classes are replicated until the distribution of all the classes is equal (Vanacore et al., 2023). But this strategy has disadvantages. One it increases the overfitting chances as the model learns too much from the data and second if the original dataset already has a huge number of instances then the computation time increases while training the model (Vanacore et al., 2023). Under-sampling on the other hand removes the instances belonging to the majority class until the distribution of all the classes is equal (Martins et al., 2023).

### 3.1.7   Cross-validation

Stratified k-fold cross-validation with a hold-out approach is considered in this study by keeping 'k' as 10. The 25% hold-out test data will not be used in stratified k-fold crossvalidation. K-fold cross-validation splits the training data into 'k' sets with an equal number of samples. In every iteration, (k-1) folds are used for training and the remaining fold is used for validation. This ensures that every sample from the main training set appears as training as well as validation data. Additionally, stratification sees to it that the folds are made by preserving the distribution of samples for each class (Demirkıran et al., 2022).

## 3.2 Modelling Techniques

Supervised Machine Learning techniques are used to predict outcomes by learning, identifying relationships and making decisions from labelled data (Malik et al., 2022). The models after being trained and tested predict the value of the outcome based on the input features. These techniques are widely used for classification problems where the target variable is categorical. As the problem in this study involves classifying the instances into one of the three classes, it is a supervised multiclass or multinomial classification (Malik et al., 2022). The study is divided into two parts. The first is about building a benchmark model using feature selection as the LASSO technique. The second part is about building various supervised multiclass classification models but with only filter-based feature selection techniques.

### 3.2.1 Benchmark Model

Griffith et al., 2020 have built Logistic LASSO Regression models using this mortality dataset with the original binary target variables, one each for 5-year and 10-year mortality prediction for T2DM patients. The two models are regression models built with LASSO regularization using the 'glmnet' R package. The results of the models built in this study, cannot be compared with the previous research conducted on this dataset as that research had produced two separate prediction models and were not regarding multiclass classification.

The Generalized Linear Models (GLMNET) with Elastic-Net and LASSO regularizations, is an efficient R package. It is used to fit various models such as the linear, cox, poisons, logistic as well as multinomial regression models (Wang et al., 2023). Therefore, in this study, the same 'glmnet' R package is considered to build the benchmark model but the family parameter is set as 'multinomial' while using the 'glmnet' function. This benchmark model is built using the Multinomial Logistic Regression with LASSO regularization.

### 3.2.2 Various Other Supervised ML Classifiers

Several ML techniques can be used for supervised multiclass classification but one of the reasons the ML techniques chosen for this study and mentioned in this section are because these are the most prominent ones as seen in the body of literature. Multinomial Logistic Regression (MLR) is a simple non-ensemble ML classifier that is considered in this study. The remaining three classifiers are ensembles. An ensemble is a collection of classifiers that are trained one after the other and then their decisions are combined.

Homogeneous ensembles are a pool of the same base classifiers and can be further classified into bagging and boosting methods (Shobana and Nandhini, 2022).

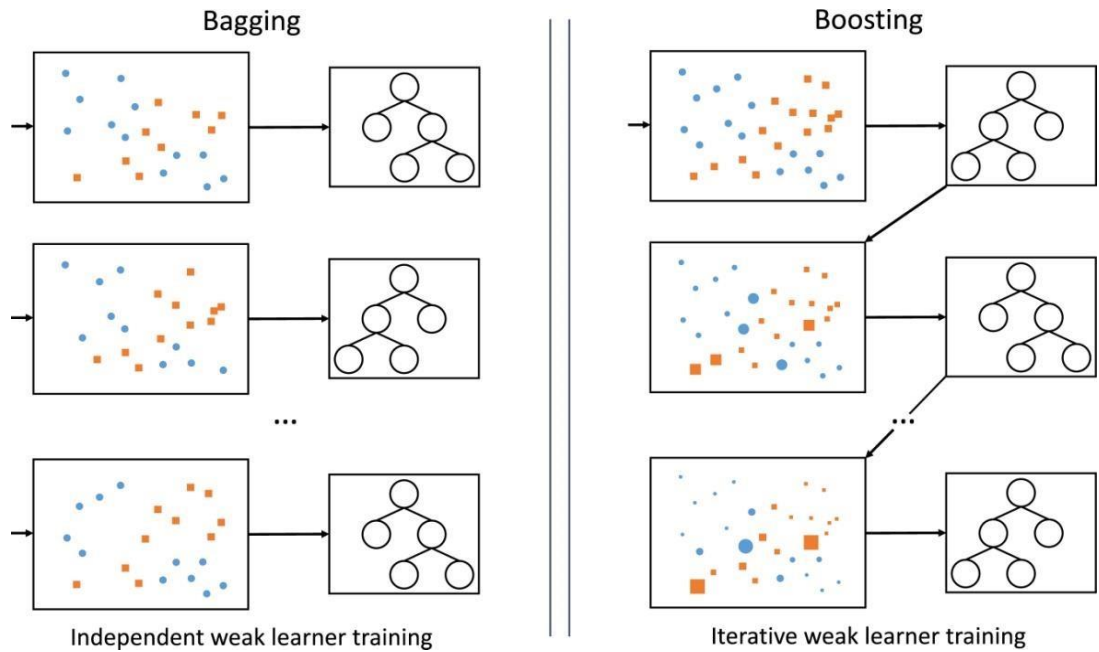

Figure 3.2 Workflow of bagging and boosting algorithms

(Image taken from González et al., 2020)

Bagging is a machine learning technique wherein multiple versions of a predictor classifier are generated and used to get an aggregated one (Shorewala, 2021). The Random Forest (RF) which consists of multiple decision trees is explored. Boosting is a technique where the errors occurring due to a model are corrected by the subsequent models. This method reduces bias by using a weighting strategy (El-Sappagh et al., 2022a). In bagging, multiple models predict outcomes in parallel and in boosting the predictor models are generated sequentially. Figure 3.2 depicts the difference in the workflow of bagging and boosting algorithms. Extreme Gradient Boosting (XGBoost) is the classifier that is also included in this study. Lastly, the One-Vs-Rest Classifier has also been considered to perform the multiclass classification. This classifier works differently and it can be considered as an ensemble of multiple binary classifiers.

**3.2.2.1 Multinomial Logistic Regression (MLR)**

Logistic Regression (LR) is used in classification problems having several independent variables for the prediction of a class from the dependent or target variable which is categorical. The core function used in this algorithm is the sigmoid function. Any output value from the range 0 and 1 can be taken by this sigmoid function (Biswas et al., 2021). It is the simplest ML classifier for binary classification which can estimate the probability of the occurrence of an event. The sigmoid or the logistic function which is an S-shaped curve can be defined as the equation (3.4) wherein the sigmoid function is defined as $f(x)$, Euler's number is $e$, the slope of the curve is defined by $k$, and the sigmoid midpoint's $x$ value is given by $x_0$ and lastly, the curve's maximum value is $L$ (de Andrade et al., 2023).

Despite the name, logistic regression is more of a classifier than a regression technique and this is especially true when the focus is on prediction in medicine and the scientific field (de Andrade et al., 2023). In this study, the extended form of LR that is used for multiclass classification problems is considered. It is the Logistic Regression classifier with the multiclass parameter (denoted as multi_class), set as 'multinomial'. Consider $C$ as the number of target classes. When $C > 2$, the model gives a $C$-dimensional vector output in which the elements sum up to 1 and then the *argmax* function predicts the class having the maximum probability value (Sipper, 2022).

**3.2.2.2 Random Forest (RF)**

Random Forest (RF) is another supervised ML algorithm. It is a group or ensemble formed by several decision trees (DT). In the decision tree, a hierarchical tree structure is formed due to the decisions made by the algorithm. This tree consists of internal nodes which are used to form decisions and thus they are the decision nodes. The attributes are the internal nodes and attribute values are represented by the branches (Guleria et al., 2022). The classification is performed at the leaf nodes. Figure 3.3 depicts the simple structure of a decision tree.

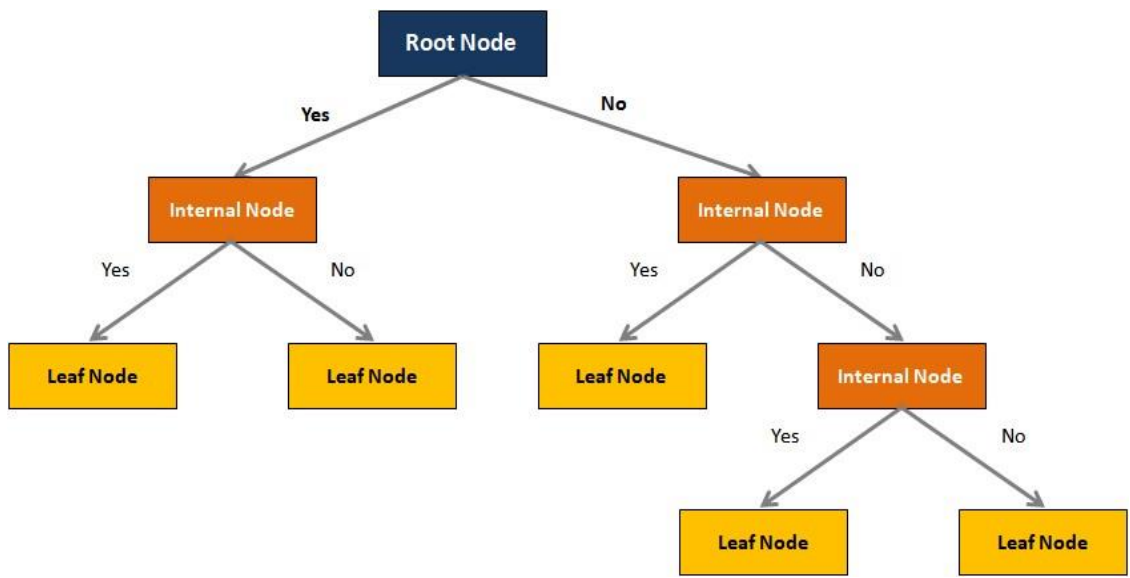

Figure 3.3 Structure of a decision tree

In RF, bagging is one of the main embedded computational procedures and the other one is the Classification And Regression Tree (CART) criterion (de Andrade et al., 2023). For large datasets, the bagging technique is advantageous as it can improve unstable estimates. It has an aggregation scheme by which subsamples are generated from the main dataset and a predictor is built for each re-sampling followed by a decision using mean. Individual trees are constructed by using the CART algorithm as well as the CART-split derivative. In every node of a tree, the best amount is selected and for classification, the criterion for CART split is optimized based on the Gini impurity idea (de Andrade et al., 2023).

Basically, the RF algorithm can perform classification as well as regression. It performs better when classification problems have more than two classes in the target or output variable and is well-suited for classification problems in the healthcare industry (Guleria et al., 2022). The bagging method, on which the RF is based, basically generates multiple versions of predictor models and then an aggregated predictor is formed using the other predictors. The performance of multiple weak classifiers is increased as these homogeneous models are employed in parallel (Shorewala, 2021).

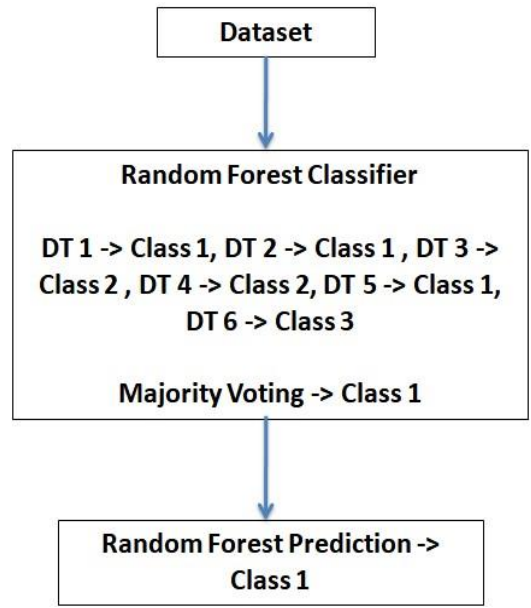

Figure 3.4 Prediction process of a Random Forest algorithm

The process of predicting the outcome by using the RF algorithm is shown in Figure 3.4. As mentioned earlier, RF is a group of multiple decision trees (DT) and the outcome from each tree is stored. The total votes for each individual class are calculated and the final outcome is declared by using majority voting. So, in the example shown in Figure 3.4, three out of a total of six trees have given the outcome as 'Class 1' for a sample and hence the final outcome predicted is 'Class 1' by the Random Forest for that particular sample.

**3.2.2.3 Extreme Gradient Boosting (XGBoost)**

Boosting method in ensembles is also used for supervised ML problems. In this technique, sequential learning is used. In an ensemble of weak models, subsequent models correct the errors that occurred in the previous models (Shobana and Nandhini, 2022). Initially, all the data points from a subset of the original data are given equal weights and a base model is created which is used to perform predictions on this subset. Now, errors would be calculated by comparing the predicted values with the actual values and the wrongly predicted data points are given the highest weights. The next model is created which tries to correct the previous model's errors.

After several iterations, multiple models would be created with the final model being the combination of all the previous models' weighted mean (Shobana and Nandhini, 2022). Thus, a combination of multiple weak models would produce a strong one that would be the ensemble. Every model of the ensemble will boost the performance and hence the method is termed as boosting. Boosting method in ensembles is also used for supervised ML problems. In this technique, sequential learning is used. In an ensemble of weak models, subsequent models correct the errors that occurred in the previous models (Shobana and Nandhini, 2022). The generalized form of a boosting machine model is the Gradient Boosting Machine (GBM) and during model fitting the loss function is optimized by the application of the gradient descent method (Bárcenas and Fuentes-García, 2022).

An extension to GBM is the Extreme Gradient Boosting (XGBoost) which is a portable, flexible, and highly efficient classifier. Problems are solved accurately and in a faster way as this classifier provides a parallel tree boosting. Its advantage over other algorithms can be maximized by parameter tuning. The overfitting problem is also reduced by employing the XGBoost algorithm. XGBoost is a regularized boosting algorithm which works even with the encoded categorical data (Bárcenas and Fuentes-García, 2022).

### 3.2.2.4 One-Vs-Rest Classifier

Multiclass classification can also be performed by an ensemble of binary classifiers. For this, two decomposition strategies that are well-known, one-vs-one (OVO) and one-vs-all (OVA) (Galar et al., 2011). In OVO, each binary classifier learns to differentiate between a pair of classes and then output is predicted after combining their outputs. In OVA which is also known as One-Vs-Rest, each binary classifier learns about one class as the positive class by treating the remaining classes as the negative class (Galar et al., 2011). The final class is predicted by taking the highest score out of all classifiers (Yashodhar and Kini, 2021). Thus, the One-Vs-Rest Classifier follows the one-vs-rest scheme and converts the multiclass classification into multiple binary classifications. As the target variable has 3 classes, there are 3 binary classifiers trained by the One-Vs-Rest classifier, one for each class. Thus, it uses a base estimator per target class. In this study, the base estimator is a Logistic Regression classifier for the One-Vs-Rest Classifier.

### 3.3    Evaluation Metrics

The evaluation of the performance of any classification model does not have any universal standards. It depends on the problem and the requirement of the user (Ali et al., 2022). In this study, the correct prediction of all three classes is important, especially the classes with a smaller number of instances. The evaluation metrics used for assessing model performance are the confusion matrix, accuracy, recall, and Cohen's kappa coefficient.

### 3.3.1    Confusion Matrix

The confusion matrix helps to see the comparison between the actual label and the predicted label for each class. It is the easiest way to understand the number of correct and incorrect predictions for each class as it summarizes the model's performance in the form of a visualization. Consider there are *N* classes in the target variable that are classified by the prediction model. A *N* x *N* table (or matrix) is formed to summarize how the model has classified the instances.

In a confusion matrix, the result for each data instance can be one out of the four categories namely; True Positive (TP), False Positive (FP), True Negative (TN), and False Negative (FN). For multiclass classification, the class label for which the calculation is being done is the positive class and the remaining class labels are the negative class (Ali et al., 2022). Whenever the model correctly predicts the positive class, it is True Positive (TP). If the negative class is incorrectly predicted as a positive class, then it is False Positive (FP). When the model correctly predicts the negative class, it is True Negative (TN) and if the positive class is incorrectly predicted as a negative class, then it is False Negative (FN). Figure 3.5 depicts the confusion matrix for class 2 (positive class) in a multiclass classification problem. Class 1 and Class 3 are negative classes in this case.

|  |  | Predicted | | |
|---|---|---|---|---|
|  | Classes | 1 | 2 | 3 |
| Actual | 1 | TN | FP | TN |
|  | 2 | FN | TP | FN |
|  | 3 | TN | FP | TN |

Figure 3.5 Example of a confusion matrix for 'Class 2' (Multiclass Classification)

### 3.3.2 Accuracy

Accuracy is the percentage of predictions the model got right out of the total (Ali et al., 2022). It describes how the model performs across all the classes. The accuracy can be expressed as Eq. 3.4.

$$Accuracy = \frac{TP + TN}{TP + FP + FN + TN}$$

……………….. (Eq. 3.4)

### 3.3.3 Recall

The recall metric indicates the percentage of actual instances of a class that are correctly predicted as that class (Ali et al., 2022; Jayasri and Aruna, 2022). It is the True Positive Rate. So, for a class, recall can be expressed as Eq. 3.5. The number of false negatives (FN) needs to be low so that recall of a class is high.

$$Recall = \frac{TP}{TP + FN}$$

…………………..(Eq. 3.5)

### 3.3.4 Cohen's Kappa Coefficient

Cohen's Kappa (κ) score can be used to see the level of agreement between two predictors. The classification model's performance can also be evaluated using this metric (Dewangan et al., 2022). Here, $P_o$ indicates the actual agreement and $P_e$ indicates the chance agreement (McHugh, 2012).

The kappa value ranges from -1 to +1 and thus the value ≤ 0 means no agreement (McHugh, 2012). Other agreement levels are, 0.01 - 0.20 means none to slight, fair is indicated by 0.21 - 0.40, moderate is 0.41 - 0.60, substantial is 0.61–0.80, and 0.81 - 1.00 means almost perfect (McHugh, 2012).

This section contains a detailed explanation of the various steps involved in research methodology. The data selected in this study on mortality risk prediction for T2DM patients is real-world data containing medical records of U.S. Military Veterans aged 65 years or older and having T2DM. The dataset contains 68 potential predictors of mortality risk in T2DM patients. A single target variable 'Mortality' is formed by combining the two target variables originally present in the dataset. Missing values are considered as a new category by employing the missing indicator method. These steps are part of data pre-processing. The data transformation part consists of converting the values of variables present to a more compatible format for machine learning classifiers.

The medical data considered for this study is not highly imbalanced but still a stratified approach in train-test splitting of the entire dataset is considered to maintain the uneven distribution of all the classes in each set. Further, stratified 10-fold cross-validation with a holdout test set has been decided to be used. Filter-based feature selection techniques, Information Gain and Chi-Squared, have been briefed. First, a benchmark model consisting of Multinomial Logistic Regression with LASSO regularization has been discussed. Then all the supervised ML classifiers that are employed in this study, have been elaborated. Multinomial Logistic Regression is a non-ensemble classifier that has been considered. In ensembles, Random Forest and XGBoost have been considered. One-Vs-Rest Classifier has also been employed in this study. Lastly, evaluation metrics such as accuracy, recall score, Cohen's kappa and confusion matrix have been considered for assessing model performance.

## 4. ANALYSIS AND DESIGNS

This section gives the details of the model development wherein the sub-sections include the details of the methodology decided for this study. Every data preparation step involved has been included and the data has been analysed to get insights. The two original target variables have been transformed into a single target variable so that the dataset can be employed for multiclass classification. Records having the same values in all the variables have also been identified and removed as these records are duplicates. Missing values and outliers have been identified and handled in a way that preserves the integrity of the data. The continuous numeric variables have been transformed into categorical variables. Univariate analysis has been done to see the distribution of the values in each of the variables present in the complete cleaned dataset.

Next, the cleaned mortality dataset has been divided into training and testing sets based on the ratio reported in the previous study conducted on this dataset. Further, the bivariate analysis has been done on the training dataset. For this, the Chi-square test of independence has been performed using the IBM SPSS software and the results have been interpreted. This test is used to discover whether the relationship between some of the independent variables and the target variable is significant or not. Then, the steps to conduct under-sampling are elaborated for class balancing as the number of samples per target class of the training dataset is uneven. Then a benchmark model which is a regression model using the base methodology, has been explained and is built using R programming in the RStudio desktop application. The rest of the models are built using Python programming in Jupyter Notebooks with the Anaconda Navigator desktop graphical user interface (GUI).

As part of the proposed methodology, filter-based feature selection techniques such as Chisquared statistics and Information Gain (or Mutual Information) have been used to select the features most relevant for classification. Further, various classifiers such as the Multinomial Logistic Regression, Random Forest, XGBoost, and the One-Vs-Rest Classifier, have been explained with their configurations for multiclass classification. The configurations of the grid search cross-validation for hyperparameter tuning employed in whichever models have also been mentioned. 10-fold cross-validation is also used and its details have been elaborated. Further, various metrics are considered to evaluate the performance of the models constructed. Lastly, an additional analysis has also been conducted using the Chi-Square Statistics in Python to check the association of every independent input feature with the target variable to gain more insights.

## 4.1 Data Preparation

The mortality dataset has 275,190 records and 70 variables of which two are target variables. The rest 68 variables are the potential input variables/features. In this section, the steps taken to prepare the data for further analysis have been described. The data underwent a thorough cleaning process to ensure data quality and consistency. Data needs to be suitable for various modelling techniques employed to meet the research objectives of this study. The entire data preparation process has been done using the Python programming language in Jupyter Notebook available through the Anaconda Navigator desktop GUI. At the start, commonly required Python libraries such as numpy and pandas have been imported. The dataset is then imported into a Python data frame using the read_csv() pandas function in Python. The CSV file containing the mortality data is placed at the same location as the Jupyter Notebook and only the filename has been provided as a parameter in the read_csv() function.

### 4.1.1 Conversion of the problem to Multiclass Classification

A new target variable named 'Mortality' is created by concatenating the values present in the original target columns 'DEATH_5' and 'DEATH_10'. The '+' operator has been used to perform this string concatenation. Table 4.1 depicts the scenarios for each possible combination of values in the 'DEATH_5' and 'DEATH_10' target columns.

Table 4.1 Possible scenarios with respective total instances after converting the problem to multiclass classification

| Scenario | DEATH_5 | DEATH_10 | Mortality | Number of Instances |
|---|---|---|---|---|
| Remaining life is up to 5 years | 1 | 1 | 11 | 65,171 |
| Remaining life is more than 5 but up to 10 years | 0 | 1 | 01 | 92,449 |
| Remaining life is more than 10 years | 0 | 0 | 00 | 117,570 |
| Total | | | | 275,190 |

Further, the mortality variable value "11" is coded as "Class 1", "01" as "Class 2", and "00" as "Class 3", as shown in Table 4.2. This conversion is performed by first defining a dictionary and then the Python string replace() method has been used. Thus, the new target variable has 3 classes. Also, the two original target variables have been eliminated as they are unnecessary.

Table 4.2 Classes in the new Target Variable (Mortality) and their respective total instances

| Mortality | Target Class | Number of Instances |
|---|---|---|
| 11 | Class 1 | 65,171 |
| 01 | Class 2 | 92,449 |
| 00 | Class 3 | 117,570 |

Figure 4.1 depicts the comparison between the original target variables with the new target variable. Patients who were reported as dead at the 5-year follow-up mark (DEATH_5 = 1) are now represented as Class 1 in the new target variable 'Mortality'. Patients who were reported as dead at the 10-year follow-up mark (DEATH_10 = 1) also included the patients who were reported dead at the 5-year mark. But Class 2 in the new target variable 'Mortality' now shows the count of patients who died between the 5-year and 10-year mark.

Patients who were alive at the 10-year follow-up mark (DEATH_10 = 0) are now represented as Class 3 in the new target variable 'Mortality'. This dataset is to be used for mortality prediction and hence each class represents the remaining life expectancy. Thus, a total of 65,171 patients have a remaining life of up to 5 years. 92,449 patients have a remaining life of more than 5 but up to 10 years. The remaining life is more than 10 years for a total of 117,570 patients.

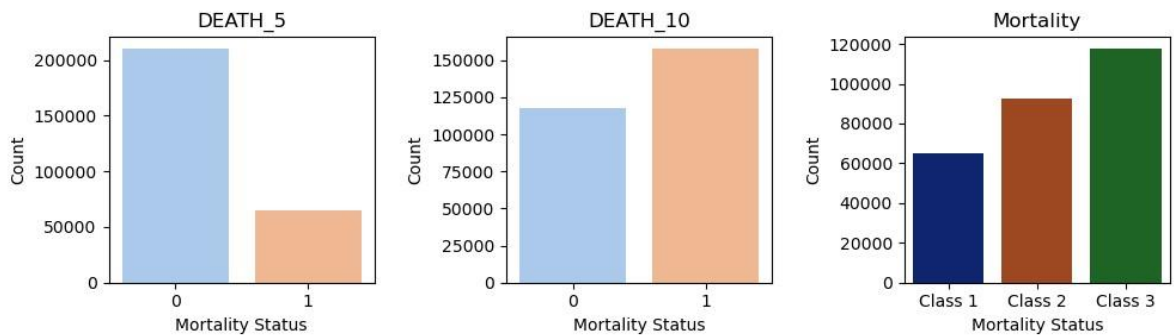

Figure 4.1 Comparison of the original target variables 'DEATH_5' and 'DEATH_10' with the new target variable 'Mortality

### 4.1.2 Removal of duplicate entries

The duplicated() method from the pandas Python library is used to find all the records in the dataset that are duplicates of other entries with every column having the same value. There are 4650 duplicate records. Another method drop_duplicates() with the parameter keep set to 'last' is used to remove the duplicate rows keeping only the last occurrence of that entry. Also, the parameter 'inplace' was set to 'True' to modify the same data frame. The index of each row in the data frame has been reset using the reset_index() Python method. The drop parameter in reset_index() is set to 'True' to avoid retention of the old indexes as a column. The parameter 'inplace' is set to 'True' to operate on the same data frame. After removing the duplicates, the total records in the dataset are 270,540.

### 4.1.3 Identifying Missing Values

In this mortality dataset, each feature is examined to identify the missing values and this has been done by using the isnull() Python method. Table 4.3 shows the six input variables that have missing values along with the percentage of missing values in the dataset. The variable MICROALB has been dropped because the percentage of missing values is very high for this variable. For the remaining five features, it has been decided to first see the outliers in the dataset and then decide how to handle these missing values. The target variable 'Mortality' has no missing values.

Table 4.3 Count and percentage of Missing Values in the dataset

| Variable | Total | Missing Values | % Missing Values |
|---|---|---|---|
| MICROALB | 270540 | 200455 | 74.09 % |
| SERUMALB | 270540 | 82512 | 30.50 % |
| LDL | 270540 | 17018 | 6.29 % |
| SERUMCRE | 270540 | 11676 | 4.32 % |
| HDL | 270540 | 7632 | 2.82 % |
| TRI | 270540 | 7211 | 2.67 % |

### 4.1.4 Outlier Detection and Treatment

Outliers are extreme values that deviate significantly from the majority of the data points. The continuous numeric variables are identified by first checking the number of unique values in each variable using the nunique() Python method and then all the variables which have more than 10 unique values have been considered as continuous numeric variables and stored in a Python list. There are 13 such variables. The outliers have been detected by plotting boxplots for these 13 continuous variables. These boxplots are graphical representations which consist of a box and two vertical lines known as whiskers. Outliers are the data points that fall beyond the range defined by the whiskers. Let's consider the boxplot shown in Figure 4.2 for the variable, AGE.

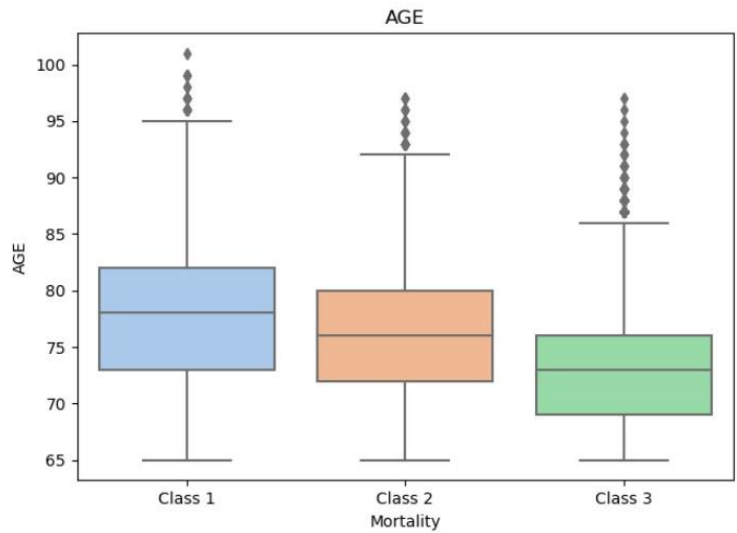

Figure 4.2 Outlier detection using boxplot for input variable 'AGE'

As we can see, there are outliers in all the mortality classes. Now the data points that are falling beyond the whiskers are not incorrect data. These data points are just rare occurrences when compared to the frequently occurring values which are in the range of the whiskers. To treat the outliers, removing such data points will lead to loss of information and capping these down to (or replacing these with) the extreme values represented by the respective whiskers
(such as 95 for 'Class 1'), will change the original data and there would be loss of information as well.

All the box plots produced are available in Appendix B: Section 4 - Additional Results. It is seen that the outliers are present in all 13 variables. These extreme values in medical data may contain valuable information and therefore it is decided to keep these outliers and handle them by the process of binning (also known as data discretization). This method was also used in the previous study on this dataset conducted by Griffith et al., 2020 and it seemed right. Binning divides the continuous data into distinct intervals known as bins. In this study, the process is performed using the cut() Python method. This method requires the parameters such as the column data to be binned, the interval edges of the bins, and the labels to be given for each bin. One more parameter 'include_lowest' has been set to 'True' so that the lowest value in the first interval is also included while binning. Also, the missing or nan (Not a number) values are stored separately as nan categorical bin by the cut() method.

Table 4.4 Binning details for the 13 continuous variables converted to categorical variables

| Variable | New Variable | Binning intervals defined | Binning Categories |
|---|---|---|---|
| AGE | AGE_GROUP | [65, 69, 74, 79, 84, 89, np.inf] | 65-69, 70-74, 75-79, 80-84, 85-89, >=90 |
| BMI | BMI_RANGE | [10, 18.4, 24.9, 39.9, 49.9, np.inf] | <18.5, 18.5-24.9, 25-39.9, 40-49.9, >=50 |
| A1C | A1C_RANGE | [-np.inf, 7.9, 9.0, np.inf] | <8, 8-9, >9 |
| SERUMALB | SERUMALB_RANGE | [-np.inf, 3.49, np.inf] | <3.5, >=3.5 |
| SERUMCRE | SERUMCRE_RANGE | [-np.inf, 1.49, 3.00, np.inf] | <1.5, 1.5-3.0, >3.0 |
| N_IP | N_IP_RANGE | [0, 5, np.inf] | 0-5, >5 |
| N_OP | N_OP_RANGE | [0, 5, 30, np.inf] | 0-5, 6-30, >30 |
| SYSTOLIC | SYSTOLIC_RANGE | [-np.inf, 119, 129, 139, 179, np.inf] | <120, 120-129, 130-139, 140-179, >=180 |
| DIASTOLIC | DIASTOLIC_RANGE | [-np.inf, 79, 89, np.inf] | <80, 80-89, >=90 |
| TRI | TRI_RANGE | [-np.inf, 149.99, 199.99, np.inf] | <150, 150-199.99, >=200 |
| LDL | LDL_RANGE | [-np.inf, 99.99, 129.99, 159.99, 189.99, np.inf] | <100, 100-129.99, 130-159.99, 160-189.99, >=190 |
| HDL | HDL_RANGE | [-np.inf, 39.99, 59.99, np.inf] | <40, 40-59.99, >=60 |
| FRAILTY | FRAILTY_GROUP | [0.00, 0.10, 0.20, 0.30, 0.40, np.inf] | Non-frail, Pre-frail, Mild, Moderate, Severe |

The bins have been formed with the help of the categories mentioned in appendix table 2 in the supplementary document provided in Vaidya et al., 2022. The binning details for the 13 continuous variables have been depicted in Table 4.4. The process of binning simplifies the data as the number of unique values is reduced due to the creation of discrete categories.

These categories are easier to interpret and analyse large datasets. So, in this dataset, the original 13 continuous variables have been dropped as their respective new categorical variables will be further used in the study.

### 4.1.5   Handling the Missing Values

The five categorical variables that still have the same percentage of missing values as the respective continuous variables had, are the 'SERUMALB_RANGE', 'LDL_RANGE', 'SERUMCRE_RANGE', 'HDL_RANGE', and 'TRI_RANGE'. This is because the binning process keeps the nan or null values as a separate category. Now, these null values are replaced with the string 'Missing' by using the fillna() Python method with the required string in it. Missing values are intentionally kept as a separate category. This preserves the information that the data was absent. It is better than changing the medical data by replacing the null values with the frequently occurring value (mode value) for categorical data. The absence of data might hold valuable insights.

### 4.1.6   Data Decoding

The variable 'RACE' was decoded. Decoding a variable means transforming encoded data into meaningful values. This is done by defining a Python dictionary with 1 as 'White', 2 as 'Black' and 3 as 'Other'. Then replace() Python method is employed to replace the values in the 'RACE' column as given in the dictionary. The cleaned data in the Python data frame is exported as a .csv file using the to_csv() Python method. The path where the .csv file is to be stored is mentioned as a parameter and the index parameter is set to 'False' to avoid the data frame index being saved in the .csv file. The filename is mentioned in the location path itself.

## 4.2   Univariate Analysis

Univariate analysis is a type of analysis wherein only one variable is focused at a time and analysed in isolation. Here, the distribution of data for that variable can be seen without taking its relationship with other variables into consideration. The value_counts() method in Python is used to find the count of each distinct value in all the 68 variables present in the complete cleaned dataset which has the missing values as a separate category in five variables
('SERUMALB_RANGE', 'LDL_RANGE', 'SERUMCRE_RANGE', 'HDL_RANGE', and
'TRI_RANGE') as discussed previously. The count of each distinct value (category) gives an overview of the prevalence of different categories in every variable. This method is used inside a for loop to iterate over a Python list that contains names of all 67 variables currently present. The count output has been added in Appendix B: Section 4 - Additional Results.

### 4.2.1   Input Variables

There are a total of 67 input variables. A simple frequency distribution gives insights into which category is prevalent in a particular variable. Let's see some of the variables related to demographics from the output generated using the value_counts() method. This dataset is male-dominated as the count of males is 267,454 and for females, it is 3086. Most of the patients fall in the 70-74 years old age group having a count of 84,165 while the least number of patients fall in the >=90 years age group with a count of 1,406.

Most of the Veteran patients (103,303) belong to Priority Group 5 and 86 patients fall in the unknown category. In the analysis of some bio measures, it is seen that the BMI of most patients (222,999) fall in the range of 25-39.9 kg/m$^2$. Only 467 patients have a BMI of less than 18.5 kg/m$^2$. The hemoglobin A1C of most of the patients (222,304) is less than 8% and the least number of patients (16,955) have more than 9% of A1C.

### 4.2.2    Target Variable

The target variable 'Mortality' has 3 classes. Class 1 has 64,190 records; Class 2 has 90,952 and Class 3 has 115,398 records. The class distribution in terms of percentage is shown in Figure 4.3. Of the three classes, the most prevalent is Class 3 and the least frequent is Class 1. In terms of mortality prediction, Class 1 represents that the remaining life is up to 5 years. Class 2 represents that the remaining life is more than 5 but up to 10 years. Class 3 represents that the remaining life is more than 10 years. The pie chart is produced using the pie() after importing the matplotlib Python library.

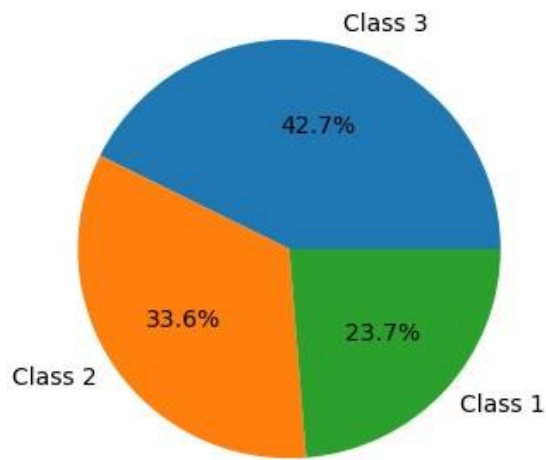

Figure 4.3 Distribution of mortality classes

### 4.3    Train-Test Split

To accurately evaluate the performance of the models, the dataset has been split into training and testing sets. The training set, comprising 75% of the data, is first used in bivariate analysis and then for model development. The remaining 25% of the data has been kept unknown to the models and thus it serves as an independent testing set to assess the predictive ability of the models. This splitting ratio was even considered in the previous study done on this dataset and thus it has been taken in this study. The train-test split is performed using the train_test_split() function present in the 'sklearn.model_selection' package in Python. As seen in Figure 4.3, the distribution of target classes is uneven. Therefore, stratify parameter has been used during the train-test split to ensure that the same proportion of target variable classes is preserved in both datasets. The target variable is assigned to the 'stratify' parameter.

For further analysis, the training and testing datasets have been exported as separate .csv files using the to_csv() Python method. The path where the .csv file is to be stored is mentioned as a parameter and the index parameter is set to 'False' to avoid the data frame index being saved in the .csv file. The filename is mentioned in the location path itself. Below mentioned is the number of records in each dataset.

1. Complete cleaned dataset – 270,540 records
2. Training dataset – 202,905 records (75%)
3. Testing dataset – 67,635 records (25%)

## 4.4 Bivariate Analysis

Bivariate analysis is the process of analysing two variables at a time. This type of analysis helps to gain insights into the existing association and dependency between the two variables. Here, the relationship between each input variable of interest and the target variable can be found by doing a bivariate analysis (Shukla et al., 2021). As the independent input variables and the dependent target variable in the dataset are categorical, the chi-square test is used for bivariate analysis.

### 4.4.1 Chi-Square Test between Input Variable and Target Variable

The Chi-square ($\chi^2$) test is a non-parametric powerful statistical tool for testing hypotheses between two categorical variables (Mchugh, 2013). The relationship between two categorical variables can be checked as the Chi-square test of independence determines whether there is a statistically significant association between two categorical variables (Hess and Hess, 2017). Out of the 67 input variables, only the demographics and clinical biomarkers are considered here for bivariate analysis. The association between the selected 17 input categorical variables and the target categorical variable present in the training dataset is checked in this study. The Chi-square test calculations were performed using the IBM SPSS Statistics software.

The chi-square test can be performed by first importing the training dataset in the SPSS software and then clicking on the 'Analyze' option on the main menu, followed by 'Descriptive Statistics', and then the 'Crosstabs' option. Select the input variable of interest (as Row(s)) and the target variable (as Column(s)) in the crosstab box with Chi-square Test as statistics. The cross-tabulation provides a tabular summary of the joint distribution of the two categorical variables being checked along with the percentage values if that is also selected in the Crosstabs: Cell Display. And then this crosstabulation table is followed by the Chi-square test results table.

Now, each of the 17 input variables has been cross-tabulated with the target variable 'Mortality'. The target variable 'Mortality' depicts the remaining life (in years) the patient will have. Thus, the remaining life is up to 5 years only for the patients that belong to Class 1, the remaining life is more than 5 but up to 10 years only for the patients that belong to Class 2, and the remaining life is more than 10 years for the patients that belong to Class 3. Hypothesis testing that is considered while analysing chi-square test results is as follows:

- Null Hypothesis (H$_0$): There is no association between the two variables (both the variables are independent)
- Alternate Hypothesis (H$_1$): There is an association between the two variables (both the variables are dependent)

1. PRIORITY

First of all, the count value in each cell of the crosstabulation shown in Table 4.5 is more than 5 and hence it meets the assumption for the Chi-square test. The highest percentage of 38.2% (77,417 records) is reported by priority GROUP 5 within the mortality variable.

Table 4.5 Crosstabulation between PRIORITY and Mortality variables

| | | | Mortality | | | |
|---|---|---|---|---|---|---|
| | | | Class 1 | Class 2 | Class 3 | Total |
| **PRIORITY** | GROUP 1 | Count | 5378 | 6712 | 7388 | 19478 |
| | | % within PRIORITY | 27.6% | 34.5% | 37.9% | 100.0% |
| | | % within Mortality | 11.2% | 9.8% | 8.5% | 9.6% |
| | GROUP 2 | Count | 1940 | 2483 | 3232 | 7655 |
| | | % within PRIORITY | 25.3% | 32.4% | 42.2% | 100.0% |
| | | % within Mortality | 4.0% | 3.6% | 3.7% | 3.8% |
| | GROUP 3 | Count | 3682 | 5154 | 6226 | 15062 |
| | | % within PRIORITY | 24.4% | 34.2% | 41.3% | 100.0% |
| | | % within Mortality | 7.6% | 7.6% | 7.2% | 7.4% |
| | GROUP 4 | Count | 2213 | 1643 | 735 | 4591 |
| | | % within PRIORITY | 48.2% | 35.8% | 16.0% | 100.0% |
| | | % within Mortality | 4.6% | 2.4% | 0.8% | 2.3% |
| | GROUP 5 | Count | 20007 | 26965 | 30445 | **77417** |
| | | % within PRIORITY | 25.8% | 34.8% | 39.3% | 100.0% |
| | | % within Mortality | **41.6%** | **39.5%** | 35.2% | **38.2%** |
| | GROUP 6 | Count | 172 | 245 | 395 | 812 |
| | | % within PRIORITY | 21.2% | 30.2% | 48.6% | 100.0% |

|   |   |   | % within Mortality | 0.4% | 0.4% | 0.5% | 0.4% |
|---|---|---|---|---|---|---|---|
|   | GROUP 7 | Count |   | 1587 | 2482 | 3511 | 7580 |
|   |   | % within PRIORITY |   | 20.9% | 32.7% | 46.3% | 100.0% |
|   |   | % within Mortality |   | 3.3% | 3.6% | 4.1% | 3.7% |
|   | GROUP 8 | Count |   | 13152 | 22514 | 34573 | 70239 |
|   |   | % within PRIORITY |   | 18.7% | 32.1% | 49.2% | 100.0% |
|   |   | % within Mortality |   | 27.3% | 33.0% | **39.9%** | 34.6% |
|   | Unknown | Count |   | 11 | 16 | 44 | **71** |
|   |   | % within PRIORITY |   | 15.5% | 22.5% | 62.0% | 100.0% |
|   |   | % within Mortality |   | **0.0%** | **0.0%** | **0.1%** | **0.0%** |
| Total |   | Count |   | 48142 | 68214 | 86549 | 202905 |
|   |   | % within PRIORITY |   | 23.7% | 33.6% | 42.7% | 100.0% |
|   |   | % within Mortality |   | 100.0% | 100.0% | 100.0% | 100.0% |

By column-wise, the highest percentage of 41.6% is again reported by priority GROUP 5 for the Class 1 mortality category. For the Class 2 mortality category, the highest percentage of 39.5% is again under priority GROUP 5. For the Class 3 mortality category, the highest percentage of 39.9% is reported by priority GROUP 8. Priority GROUP 5 comprises those patients having economic hardships. Priority GROUP 8 patients are those who do not have economic hardships as household income is above the threshold or service-connected disability (Vaidya et al., 2022). Thus, the chances of remaining life to be more than 10 years (Class 3) is more for the patients belonging to GROUP 8 than GROUP 5. The lowest percentage is reported by the unknown priority category for all mortality classes as the total count is just 71.

Table 4.6 Chi-square test between PRIORITY and Mortality variables

| **Chi-Square Tests** | | | |
|---|---|---|---|
|   | **Value** | **Df** | **Asymptotic Significance (2-sided)** |
| Pearson Chi-Square | 4139.033[a] | 16 | <.001 |
| Likelihood Ratio | 4135.267 | 16 | <.001 |
| N of Valid Cases | 202905 |   |   |
| a. 0 cells (0.0%) have expected count less than 5. The minimum expected count is 16.85. | | | |

As seen in Table 4.6, the Pearson Chi-Square value is 4139.033 with the degree of freedom as 16 at a p-value of <0.001. The null hypothesis ($H_0$) is rejected as the p-value is less than the significance level (0.05). This concludes that there is a significant association between the variables PRIORITY and Mortality.

2. MARRIED

First of all, the count value in each cell of the crosstabulation shown in Table 4.7 is more than 5 and hence it meets the assumption for the Chi-square test. In Table 4.7, the highest percentage of 65.0% (131,953 records) is reported by the MARRIED category within the mortality variable. By column-wise, the highest percentage of 64.2% is again reported by the MARRIED category for the Class 1 mortality category. For the Class 2 mortality category, the highest percentage of 62.9% is again under the MARRIED category. For the Class 3 mortality category, the highest percentage of 67.1% is also reported by the MARRIED category.

Table 4.7 Crosstabulation between MARRIED and Mortality variables

| Crosstabulation | | | | | | |
|---|---|---|---|---|---|---|
| | | | Mortality | | | |
| | | | Class 1 | Class 2 | Class 3 | Total |
| **MARRIED** | MARRIED | Count | 30925 | 42939 | 58089 | **131953** |
| | | % within MARRIED | 23.4% | 32.5% | 44.0% | 100.0% |
| | | % within Mortality | **64.2%** | **62.9%** | **67.1%** | **65.0%** |
| | SINGLE | Count | 7860 | 10622 | 13294 | 31776 |
| | | % within MARRIED | 24.7% | 33.4% | 41.8% | 100.0% |
| | | % within Mortality | **16.3%** | **15.6%** | **15.4%** | 15.7% |
| | WIDOWED | Count | 9357 | 14653 | 15166 | 39176 |
| | | % within MARRIED | 23.9% | 37.4% | 38.7% | 100.0% |
| | | % within Mortality | 19.4% | 21.5% | 17.5% | 19.3% |
| Total | | Count | 48142 | 68214 | 86549 | 202905 |
| | | % within MARRIED | 23.7% | 33.6% | 42.7% | 100.0% |
| | | % within Mortality | 100.0% | 100.0% | 100.0% | 100.0% |

The lowest percentage is reported by the SINGLE category for all mortality classes with the lowest being 15.4% under the Class 3 category of Mortality. Thus, the chances of remaining life to be more than 10 years (Class 3) are more for the patients who are married. The chances of remaining life to be more than 10 years (Class 3) is the least for the single patients.

Table 4.8 Chi-square test between MARRIED and Mortality variables

| Chi-Square Tests | | | |
|---|---|---|---|
| | Value | Df | Asymptotic Significance (2-sided) |
| Pearson Chi-Square | 437.131[a] | 4 | <.001 |
| Likelihood Ratio | 435.561 | 4 | <.001 |
| N of Valid Cases | 202905 | | |
| a. 0 cells (0.0%) have expected count less than 5. The minimum expected count is 7539.29. | | | |

As seen in Table 4.8, the Pearson Chi-Square value is 437.131 with the degree of freedom as 4 at a p-value of <0.001. The null hypothesis ($H_0$) is rejected as the p-value is less than the significance level (0.05). This concludes that there is a significant association between the variables MARRIED and Mortality.

3. SEX

First of all, the count value in each cell of the crosstabulation shown in Table 4.9 is more than 5 and hence it meets the assumption for the Chi-square test. In Table 4.9, the male category is denoted by 1 and the female category is denoted by 0. The highest percentage of 98.8% (200,547 records) is reported under the Male category within the mortality variable. By column-wise as well, the highest percentage is again reported under the male category for all the Classes of mortality. The data is male dominated as there are very few female records.

Table 4.9 Crosstabulation between SEX and Mortality variables

| Crosstabulation | | | | | | | |
|---|---|---|---|---|---|---|---|
| | | | | Mortality | | | |
| | | | | Class 1 | Class 2 | Class 3 | Total |
| SEX | 0 | | Count | 540 | 783 | 1035 | 2358 |
| | | | % within SEX | 22.9% | 33.2% | 43.9% | 100.0% |
| | | | % within Mortality | 1.1% | 1.1% | 1.2% | 1.2% |

| | | | Class 1 | Class 2 | Class 3 | Total |
|---|---|---|---|---|---|---|
| | 1 | Count | 47602 | 67431 | 85514 | **200547** |
| | | % within SEX | 23.7% | 33.6% | 42.6% | 100.0% |
| | | % within Mortality | **98.9%** | **98.9%** | **98.8%** | **98.8%** |
| Total | | Count | 48142 | 68214 | 86549 | 202905 |
| | | % within SEX | 23.7% | 33.6% | 42.7% | 100.0% |
| | | % within Mortality | 100.0% | 100.0% | 100.0% | 100.0% |

Table 4.10 Chi-square test between SEX and Mortality variables

| Chi-Square Tests | | | |
|---|---|---|---|
| | Value | Df | Asymptotic Significance (2-sided) |
| Pearson Chi-Square | 1.664[a] | 2 | .435 |
| Likelihood Ratio | 1.664 | 2 | .435 |
| N of Valid Cases | 202905 | | |
| a. 0 cells (0.0%) have expected count less than 5. The minimum expected count is 559.47. | | | |

As seen in Table 4.10, the Pearson Chi-Square value is 1.664 with the degree of freedom as 2 at p-value = 0.435. We fail to reject the null hypothesis ($H_0$) as the p-value is more than the significance level (0.05). This concludes that there is no significant association between the variables SEX and Mortality.

4. RACE

First of all, the count value in each cell of the crosstabulation shown in Table 4.11 is more than 5 and hence it meets the assumption for the Chi-square test. In Table 4.11, the highest percentage of 86.6% (175,703 records) is reported under the 'White' category within the mortality variable. By column-wise as well, the highest percentage is again reported by the

White category for all the classes of mortality. The lowest percentage is reported by the

'Other' category for all the classes of mortality.

Table 4.11 Crosstabulation between RACE and Mortality variables

| Crosstabulation | | | | | | |
|---|---|---|---|---|---|---|
| | | | Mortality | | | |
| | | | Class 1 | Class 2 | Class 3 | Total |
| **RACE** | Black | Count | 4833 | 6381 | 9569 | 20783 |
| | | % within RACE | 23.3% | 30.7% | 46.0% | 100.0% |
| | | % within Mortality | 10.0% | 9.4% | 11.1% | 10.2% |

|  | | Count | 1443 | 2048 | 2928 | 6419 |
|---|---|---|---|---|---|---|
| | Other | % within RACE | 22.5% | 31.9% | 45.6% | 100.0% |
| | | % within Mortality | **3.0%** | **3.0%** | **3.4%** | 3.2% |
| | White | Count | 41866 | 59785 | 74052 | **175703** |
| | | % within RACE | 23.8% | 34.0% | 42.1% | 100.0% |
| | | % within Mortality | **87.0%** | **87.6%** | **85.6%** | **86.6%** |
| Total | | Count | 48142 | 68214 | 86549 | 202905 |
| | | % within RACE | 23.7% | 33.6% | 42.7% | 100.0% |
| | | % within Mortality | 100.0% | 100.0% | 100.0% | 100.0% |

As seen in Table 4.12, the Pearson Chi-Square value is 153.508 with the degree of freedom as 4 at a p-value of <0.001. The null hypothesis ($H_0$) is rejected as the p-value is less than the significance level (0.05). This concludes that there is a significant association between the variables RACE and Mortality.

Table 4.12 Chi-square test between RACE and Mortality variables

| **Chi-Square Tests** | | | |
|---|---|---|---|
| | **Value** | **Df** | **Asymptotic Significance (2-sided)** |
| Pearson Chi-Square | 153.508[a] | 4 | <.001 |
| Likelihood Ratio | 153.571 | 4 | <.001 |
| N of Valid Cases | 202905 | | |
| a. 0 cells (0.0%) have expected count less than 5. The minimum expected count is 1523.00. | | | |

5. SERUMALB_RANGE

First of all, the count value in each cell of the crosstabulation shown in Table 4.13 is more than 5 and hence it meets the assumption for the Chi-square test. The SERUMALB_RANGE is a variable that denotes the range of serum albumin (in g/dL). In Table 4.13, the highest percentage of 64.8% (131,522 records) is reported under the '>=3.5' category of the SERUMALB_RANGE variable within the mortality variable. By column-wise, the highest percentage of 65.9% is again reported by the '>=3.5' category for the Class 3 mortality category. The lowest percentage of 2.1% is reported under the '<3.5' category for the Class 3.

Table 4.13 Crosstabulation between SERUMALB_RANGE and Mortality variables

| **Crosstabulation** |
|---|

|  |  |  | Mortality |  |  |  |
|---|---|---|---|---|---|---|
|  |  |  | Class 1 | Class 2 | Class 3 | Total |
| **SERUMALB_RANGE** | <3.5 | Count | 4358 | 3250 | 1858 | 9466 |
|  |  | % within SERUMALB_RANGE | 46.0% | 34.3% | 19.6% | 100.0% |
|  |  | % within Mortality | **9.1%** | **4.8%** | **2.1%** | 4.7% |
|  | >=3.5 | Count | 30271 | 44241 | 57010 | **131522** |
|  |  | % within SERUMALB_RANGE | 23.0% | 33.6% | 43.3% | 100.0% |
|  |  | % within Mortality | **62.9%** | **64.9%** | **65.9%** | **64.8%** |
|  | Missing | Count | 13513 | 20723 | 27681 | 61917 |
|  |  | % within SERUMALB_RANGE | 21.8% | 33.5% | 44.7% | 100.0% |
|  |  | % within Mortality | 28.1% | 30.4% | 32.0% | 30.5% |
| Total |  | Count | 48142 | 68214 | 86549 | 202905 |
|  |  | % within SERUMALB_RANGE | 23.7% | 33.6% | 42.7% | 100.0% |
|  |  | % within Mortality | 100.0% | 100.0% | 100.0% | 100.0% |

Table 4.14 Chi-square test between SERUMALB_RANGE and Mortality variables

| Chi-Square Tests | | | |
|---|---|---|---|
|  | **Value** | **df** | **Asymptotic Significance (2-sided)** |
| Pearson Chi-Square | 3362.954[a] | 4 | <.001 |
| Likelihood Ratio | 3230.738 | 4 | <.001 |
| N of Valid Cases | 202905 |  |  |
| a. 0 cells (0.0%) have expected count less than 5. The minimum expected count is 2245.94. | | | |

As seen in Table 4.14, the Pearson Chi-Square value is 3362.954 with the degree of freedom as 4 at a p-value of <0.001. The null hypothesis ($H_0$) is rejected as the p-value is less than the significance level (0.05). This concludes that there is a significant association between the variables SERUMALB_RANGE and Mortality at the 5% level of significance.

6. FRAILTY_GROUP

First of all, the count value in each cell of the crosstabulation shown in Table 4.15 is more than 5 and hence it meets the assumption for the Chi-square test. The FRAILTY_GROUP is a variable that denotes the frailty index groups. In Table 4.15, the highest percentage of 52.6% (106,752 records) is reported under the 'Severe' category of the FRAILTY_GROUP variable within the mortality variable. By column-wise, the highest percentage of 56.8% is again reported by the 'Severe' category for the Class 1 mortality category. The lowest percentage of 12.4% is reported under the 'Moderate' category for the Class 1 as well as Class 2 mortality categories.

Table 4.15 Crosstabulation between FRAILTY_GROUP and Mortality variables

| Crosstabulation | | | Mortality | | | Total |
|---|---|---|---|---|---|---|
| | | | Class 1 | Class 2 | Class 3 | |
| **FRAILTY_GROUP** Moderate | Count | | 5952 | 8443 | 11459 | 25854 |
| | % within FRAILTY_GROUP | | 23.0% | 32.7% | 44.3% | 100.0% |
| | % within Mortality | | **12.4%** | **12.4%** | **13.2%** | 12.7% |
| Non-frail | Count | | 14848 | 22132 | 33319 | 70299 |
| | % within FRAILTY_GROUP | | 21.1% | 31.5% | 47.4% | 100.0% |
| | % within Mortality | | 30.8% | 32.4% | 38.5% | 34.6% |
| Severe | Count | | 27342 | 37639 | 41771 | **106752** |
| | % within FRAILTY_GROUP | | 25.6% | 35.3% | 39.1% | 100.0% |
| | % within Mortality | | **56.8%** | **55.2%** | **48.3%** | **52.6%** |
| Total | Count | | 48142 | 68214 | 86549 | 202905 |
| | % within FRAILTY_GROUP | | 23.7% | 33.6% | 42.7% | 100.0% |
| | % within Mortality | | 100.0% | 100.0% | 100.0% | 100.0% |

Table 4.16 Chi-square test between FRAILTY_GROUP and Mortality variables

| Chi-Square Tests | | | |
|---|---|---|---|
| | Value | Df | Asymptotic Significance (2-sided) |
| Pearson Chi-Square | 1252.936[a] | 4 | <.001 |
| Likelihood Ratio | 1253.223 | 4 | <.001 |
| N of Valid Cases | 202905 | | |

> a. 0 cells (0.0%) have expected count less than 5. The minimum expected count is 6134.22.

As seen in Table 4.16, the Pearson Chi-Square value is 1252.936 with the degree of freedom as 4 at a p-value of <0.001. The null hypothesis ($H_0$) is rejected as the p-value is less than the significance level (0.05). This concludes that there is a significant association between the variables FRAILTY_GROUP and Mortality at the 5% level of significance.

7. AGE_GROUP

First of all, the count value in each cell of the crosstabulation shown in Table 4.17 is more than 5 and hence it meets the assumption for the Chi-square test. The AGE_GROUP is a variable that denotes the age group (in years). In Table 4.17, the highest percentage of 31.1% (63,199 records) is reported under the '70-74' category of the AGE_GROUP variable within the mortality variable. By column-wise, the highest percentage of 28.0% is reported under the

'75-79' age group category for the Class 1 mortality category. The highest percentage of 29.4% is reported under the '75-79' age group category for the Class 2 mortality category. The highest percentage of 38.2% is reported under the '70-74' age group category for the Class 3 mortality category. The lowest percentage is reported under the '>=90' age group category for all the mortality classes.

Table 4.17 Crosstabulation between AGE_GROUP and Mortality variables

| Crosstabulation | | | Mortality | | | Total |
|---|---|---|---|---|---|---|
| | | | Class 1 | Class 2 | Class 3 | |
| AGE_GROUP | >=90 | Count | 646 | 372 | 41 | 1059 |
| | | % within AGE_GROUP | 61.0% | 35.1% | 3.9% | 100.0% |
| | | % within Mortality | **1.3%** | **0.5%** | **0.0%** | 0.5% |
| | 65-69 | Count | 4828 | 8660 | 22880 | 36368 |
| | | % within AGE_GROUP | 13.3% | 23.8% | 62.9% | 100.0% |
| | | % within Mortality | 10.0% | 12.7% | 26.4% | 17.9% |
| | 70-74 | Count | 11394 | 18724 | 33081 | **63199** |
| | | % within AGE_GROUP | 18.0% | 29.6% | 52.3% | 100.0% |
|  |  | % within Mortality | 23.7% | 27.4% | **38.2%** | **31.1%** |
|  | 75-79 | Count | 13479 | 20046 | 20631 | 54156 |
|  |  | % within AGE_GROUP | 24.9% | 37.0% | 38.1% | 100.0% |
|  |  | % within Mortality | **28.0%** | **29.4%** | 23.8% | 26.7% |
|  | 80-84 | Count | 12772 | 16097 | 8749 | 37618 |
|  |  | % within AGE_GROUP | 34.0% | 42.8% | 23.3% | 100.0% |
|  |  | % within Mortality | 26.5% | 23.6% | 10.1% | 18.5% |
|  | 85-89 | Count | 5023 | 4315 | 1167 | 10505 |
|  |  | % within AGE_GROUP | 47.8% | 41.1% | 11.1% | 100.0% |
|  |  | % within Mortality | 10.4% | 6.3% | 1.3% | 5.2% |
| Total |  | Count | 48142 | 68214 | 86549 | 202905 |
|  |  | % within AGE_GROUP | 23.7% | 33.6% | 42.7% | 100.0% |
|  |  | % within Mortality | 100.0% | 100.0% | 100.0% | 100.0% |

Table 4.18 Chi-square test between AGE_GROUP and Mortality variables

| Chi-Square Tests | | | |
|---|---|---|---|
|  | Value | df | Asymptotic Significance (2-sided) |
| Pearson Chi-Square | 21354.703[a] | 10 | <.001 |
| Likelihood Ratio | 22230.046 | 10 | <.001 |
| N of Valid Cases | 202905 |  |  |
| a. 0 cells (0.0%) have expected count less than 5. The minimum expected count is 251.26. | | | |

As seen in Table 4.18, the Pearson Chi-Square value is 21354.703 with the degree of freedom as 10 at a p-value of <0.001. The null hypothesis ($H_0$) is rejected as the p-value is less than the significance level (0.05). This concludes that there is a significant association between the variables AGE_GROUP and Mortality.

8. DIASTOLIC_RANGE

First of all, the count value in each cell of the crosstabulation shown in Table 4.19 is more than 5 and hence it meets the assumption for the Chi-square test. The DIASTOLIC_RANGE is a variable that denotes the range of diastolic blood pressure (in mm Hg). In Table 4.19, the highest percentage of 88.2% (178,961 records) is reported under the '<80' category of the DIASTOLIC_RANGE variable within the mortality variable. By column-wise, the highest percentage of 91.3% is reported under the '<80' DIASTOLIC_RANGE category for the Class 1 mortality category. The highest percentage of 89.2% is reported again under the '<80' DIASTOLIC_RANGE category for the Class 2 mortality category. The highest percentage of

85.7% is reported under the '<80' DIASTOLIC_RANGE category for the Class 3 mortality category. The lowest percentage is reported under the '>=90' category for all the mortality classes.

Table 4.19 Crosstabulation between DIASTOLIC_RANGE and Mortality variables

| | | Crosstabulation | | | |
|---|---|---|---|---|---|
| | | Mortality | | | |
| | | Class 1 | Class 2 | Class 3 | Total |
| **DIASTOLIC_RANGE** <80 | Count | 43955 | 60856 | 74150 | **178961** |
| | % within DIASTOLIC_RANGE | 24.6% | 34.0% | 41.4% | 100.0% |
| | % within Mortality | **91.3%** | **89.2%** | **85.7%** | **88.2%** |
| >=90 | Count | 353 | 637 | 956 | 1946 |
| | % within DIASTOLIC_RANGE | 18.1% | 32.7% | 49.1% | 100.0% |
| | % within Mortality | **0.7%** | **0.9%** | **1.1%** | 1.0% |
| 80-89 | Count | 3834 | 6721 | 11443 | 21998 |
| | % within DIASTOLIC_RANGE | 17.4% | 30.6% | 52.0% | 100.0% |
| | % within Mortality | 8.0% | 9.9% | 13.2% | 10.8% |
| Total | Count | 48142 | 68214 | 86549 | 202905 |
| | % within DIASTOLIC_RANGE | 23.7% | 33.6% | 42.7% | 100.0% |
| | % within Mortality | 100.0% | 100.0% | 100.0% | 100.0% |

Table 4.20 Chi-square test between DIASTOLIC_RANGE and Mortality variables

| Chi-Square Tests | | | |
|---|---|---|---|
| | Value | Df | Asymptotic Significance (2-sided) |
| Pearson Chi-Square | 1049.617[a] | 4 | <.001 |
| Likelihood Ratio | 1060.753 | 4 | <.001 |
| N of Valid Cases | 202905 | | |
| a. 0 cells (0.0%) have expected count less than 5. The minimum expected count is 461.72. | | | |

As seen in Table 4.20, the Pearson Chi-Square value is 1049.617 with the degree of freedom as 4 at a p-value of <0.001. The null hypothesis ($H_0$) is rejected as the p-value is less than the significance level (0.05). This concludes that there is a significant association between the variables DIASTOLIC_RANGE and Mortality.

9. SYSTOLIC_RANGE

First of all, the count value in each cell of the crosstabulation shown in Table 4.21 is more than 5 and hence it meets the assumption for the Chi-square test. The SYSTOLIC_RANGE is a variable that denotes the range of systolic blood pressure (in mm Hg). In Table 4.21, the highest percentage of 39.9% (80,959 records) is reported under the '140-179' category of the SYSTOLIC_RANGE variable within the mortality variable. By column-wise, the highest percentage of 41.8% is reported under the '140-179' SYSTOLIC_RANGE category for the Class 2 mortality category. The highest percentage of 39.7% is reported again under the '140-179' SYSTOLIC_RANGE category for the Class 1 mortality category. The highest percentage of 38.5% is reported under the '140-179' SYSTOLIC_RANGE category for the Class 3 mortality category. The lowest percentage is reported under the '>=180' category for all the mortality classes.

Table 4.21 Crosstabulation between SYSTOLIC_RANGE and Mortality variables

| Crosstabulation | | | | | | |
|---|---|---|---|---|---|---|
| | | | Mortality | | | |
| | | | Class 1 | Class 2 | Class 3 | Total |
| SYSTOLIC_RANGE <120 | Count | | 5174 | 5809 | 6827 | 17810 |
| | % within SYSTOLIC_RANGE | | 29.1% | 32.6% | 38.3% | 100.0% |
| | | % within Mortality | 10.7% | 8.5% | 7.9% | 8.8% |
| | >=180 | Count | 234 | 296 | 242 | 772 |
| | | % within SYSTOLIC_RANGE | 30.3% | 38.3% | 31.3% | 100.0% |
| | | % within Mortality | **0.5%** | **0.4%** | **0.3%** | **0.4%** |
| | 120-129 | Count | 9510 | 12900 | 17791 | 40201 |
| | | % within SYSTOLIC_RANGE | 23.7% | 32.1% | 44.3% | 100.0% |
| | | % within Mortality | 19.8% | 18.9% | 20.6% | 19.8% |
| | 130-139 | Count | 14105 | 20697 | 28361 | 63163 |
| | | % within SYSTOLIC_RANGE | 22.3% | 32.8% | 44.9% | 100.0% |
| | | % within Mortality | 29.3% | 30.3% | 32.8% | 31.1% |
| | 140-179 | Count | 19119 | 28512 | 33328 | **80959** |
| | | % within SYSTOLIC_RANGE | 23.6% | 35.2% | 41.2% | 100.0% |
| | | % within Mortality | **39.7%** | **41.8%** | **38.5%** | **39.9%** |
| Total | | Count | 48142 | 68214 | 86549 | 202905 |
| | | % within SYSTOLIC_RANGE | 23.7% | 33.6% | 42.7% | 100.0% |
| | | % within Mortality | 100.0% | 100.0% | 100.0% | 100.0% |

Table 4.22 Chi-square test between SYSTOLIC_RANGE and Mortality variables

| Chi-Square Tests | | | |
|---|---|---|---|
| | **Value** | **Df** | **Asymptotic Significance (2-sided)** |
| Pearson Chi-Square | 634.945[a] | 8 | <.001 |
| Likelihood Ratio | 623.995 | 8 | <.001 |
| N of Valid Cases | 202905 | | |
| a. 0 cells (0.0%) have expected count less than 5. The minimum expected count is 183.17. | | | |

As seen in Table 4.22, the Pearson Chi-Square value is 634.945 with the degree of freedom as 8 at a p-value of <0.001. The null hypothesis ($H_0$) is rejected as the p-value is less than the significance level (0.05). This concludes that there is a significant association between the variables SYSTOLIC_RANGE and Mortality.

10. N_IP_RANGE

First of all, the count value in each cell of the crosstabulation shown in Table 4.23 is more than 5 and hence it meets the assumption for the Chi-square test. The N_IP_RANGE is a variable that denotes the number of inpatient days. In Table 4.23, the highest percentage of 93.8% (190,230 records) is reported under the '0-5' category of the N_IP_RANGE variable within the mortality variable. By column-wise, the highest percentage of 88.3% is reported under the '0-5' N_IP_RANGE category for the Class 1 mortality category. The highest percentage of 93.6% is reported again under the '0-5' N_IP_RANGE category for the Class 2 mortality category. The highest percentage of 96.9% is reported under the '0-5' N_IP_RANGE category for the Class 3 mortality category. The lowest percentage of 3.1% is reported under the '>5' category for the Class 3 mortality category.

Table 4.23 Crosstabulation between N_IP_RANGE and Mortality variables

| Crosstabulation | | | Mortality | | | |
|---|---|---|---|---|---|---|
| | | | Class 1 | Class 2 | Class 3 | Total |
| N_IP_RANGE | >5 | Count | 5617 | 4336 | 2722 | 12675 |
| | | % within N_IP_RANGE | 44.3% | 34.2% | 21.5% | 100.0% |
| | | % within Mortality | 11.7% | 6.4% | **3.1%** | 6.2% |
| | 0-5 | Count | 42525 | 63878 | 83827 | **190230** |
| | | % within N_IP_RANGE | 22.4% | 33.6% | 44.1% | 100.0% |
| | | % within Mortality | **88.3%** | **93.6%** | **96.9%** | **93.8%** |
| Total | | Count | 48142 | 68214 | 86549 | 202905 |
| | | % within N_IP_RANGE | 23.7% | 33.6% | 42.7% | 100.0% |
| | | % within Mortality | 100.0% | 100.0% | 100.0% | 100.0% |

Table 4.24 Chi-square test between N_IP_RANGE and Mortality variables

| Chi-Square Tests | | | |
|---|---|---|---|
| | Value | df | Asymptotic Significance (2-sided) |
| Pearson Chi-Square | 3838.680[a] | 2 | <.001 |

| Likelihood Ratio | 3675.348 | 2 | <.001 |
| N of Valid Cases | 202905 | | |

a. 0 cells (0.0%) have expected count less than 5. The minimum expected count is 3007.32.

As seen in Table 4.24, the Pearson Chi-Square value is 3838.680 with the degree of freedom as 2 at a p-value of <0.001. The null hypothesis ($H_0$) is rejected as the p-value is less than the significance level (0.05). This concludes that there is a significant association between the variables N_IP_RANGE and Mortality.

11. N_OP_RANGE

First of all, the count value in each cell of the crosstabulation shown in Table 4.25 is more than 5 and hence it meets the assumption for the Chi-square test. The N_OP_RANGE is a variable that denotes the number of outpatient visits. In Table 4.25, the highest percentage of 64.2% (130,255 records) is reported under the '6-30' category of N_OP_RANGE variable within the mortality variable. By column-wise, the highest percentage of 53.3% is reported under the '6-30' N_OP_RANGE category for the Class 1 mortality category. The highest percentage of 63.1% is reported again under the '6-30' N_OP_RANGE category for the Class 2 mortality category. The highest percentage of 71.1% is reported under the '6-30' N_OP_RANGE category for the Class 3 mortality category. The lowest percentage of 0.8% is reported under the '0-5' category for the Class 1 mortality category.

Table 4.25 Crosstabulation between N_OP_RANGE and Mortality variables

| Crosstabulation | | | | | | |
|---|---|---|---|---|---|---|
| | | | Mortality | | | |
| | | | Class 1 | Class 2 | Class 3 | Total |
| N_OP_RANGE | >30 | Count | 22099 | 24355 | 23749 | 70203 |
| | | % within N_OP_RANGE | 31.5% | 34.7% | 33.8% | 100.0% |
| | | % within Mortality | 45.9% | 35.7% | 27.4% | 34.6% |
| | 0-5 | Count | 366 | 837 | 1244 | 2447 |
| | | % within N_OP_RANGE | 15.0% | 34.2% | 50.8% | 100.0% |
| | | % within Mortality | **0.8%** | **1.2%** | **1.4%** | 1.2% |
| | 6-30 | Count | 25677 | 43022 | 61556 | **130255** |
| | | % within N_OP_RANGE | 19.7% | 33.0% | 47.3% | 100.0% |

|       | % within Mortality | **53.3%** | **63.1%** | **71.1%** | **64.2%** |
|-------|--------------------|-----------|-----------|-----------|-----------|
| Total | Count              | 48142     | 68214     | 86549     | 202905    |
|       | % within N_OP_RANGE | 23.7%    | 33.6%     | 42.7%     | 100.0%    |
|       | % within Mortality | 100.0%    | 100.0%    | 100.0%    | 100.0%    |

Table 4.26 Chi-square test between N_OP_RANGE and Mortality variables

| Chi-Square Tests | | | |
|---|---|---|---|
| | **Value** | **df** | **Asymptotic Significance (2-sided)** |
| Pearson Chi-Square | 4747.159[a] | 4 | <.001 |
| Likelihood Ratio | 4721.075 | 4 | <.001 |
| N of Valid Cases | 202905 | | |
| a. 0 cells (0.0%) have expected count less than 5. The minimum expected count is 580.58. | | | |

As seen in Table 4.26, the Pearson Chi-Square value is 4747.159 with the degree of freedom as 4 at a p-value of <0.001. The null hypothesis ($H_0$) is rejected as the p-value is less than the significance level (0.05). This concludes that there is a significant association between the variables N_OP_RANGE and Mortality.

12. BMI_RANGE

First of all, the count value in each cell of the crosstabulation shown in Table 4.27 is more than 5 and hence it meets the assumption for the Chi-square test. The BMI_RANGE is a variable that denotes the range of body mass index (in $kg/m^2$). In Table 4.27, the highest percentage of 82.1% (166,642 records) is reported under the '25-39.9' category of the BMI_RANGE variable within the mortality variable. By column-wise, the highest percentage of 77.3% is reported under the '25-39.9' category of BMI_RANGE for the Class 1 mortality category. The highest percentage of 81.2% is reported again under the '25-39.9' category of BMI_RANGE for the Class 2 mortality category. The highest percentage of 85.6% is also reported under the '25-39.9' category of BMI_RANGE for the Class 3 mortality category.

Table 4.27 Crosstabulation between BMI_RANGE and Mortality variables

| **Crosstabulation** |
|---|

|  |  |  | Mortality | | | Total |
|---|---|---|---|---|---|---|
|  |  |  | **Class 1** | **Class 2** | **Class 3** |  |
| **BMI_RANGE** | <18.5 | Count | 169 | 108 | 71 | 348 |
|  |  | % within BMI_RANGE | 48.6% | 31.0% | 20.4% | 100.0% |
|  |  | % within Mortality | **0.4%** | **0.2%** | **0.1%** | 0.2% |
|  | >=50 | Count | 189 | 229 | 195 | 613 |
|  |  | % within BMI_RANGE | 30.8% | 37.4% | 31.8% | 100.0% |
|  |  | % within Mortality | **0.4%** | 0.3% | 0.2% | 0.3% |
|  | 18.5-24.9 | Count | 8762 | 9753 | 9094 | 27609 |
|  |  | % within BMI_RANGE | 31.7% | 35.3% | 32.9% | 100.0% |
|  |  | % within Mortality | 18.2% | 14.3% | 10.5% | 13.6% |
|  | 25-39.9 | Count | 37214 | 55369 | 74059 | **166642** |
|  |  | % within BMI_RANGE | 22.3% | 33.2% | 44.4% | 100.0% |
|  |  | % within Mortality | **77.3%** | **81.2%** | **85.6%** | **82.1%** |
|  | 40-49.9 | Count | 1808 | 2755 | 3130 | 7693 |
|  |  | % within BMI_RANGE | 23.5% | 35.8% | 40.7% | 100.0% |
|  |  | % within Mortality | 3.8% | 4.0% | 3.6% | 3.8% |
| Total |  | Count | 48142 | 68214 | 86549 | 202905 |
|  |  | % within BMI_RANGE | 23.7% | 33.6% | 42.7% | 100.0% |
|  |  | % within Mortality | 100.0% | 100.0% | 100.0% | 100.0% |

The lowest percentage of 0.4% is reported under the '<18.5' and '>=50' BMI categories for the Class 1 mortality category. The lowest percentage of 0.2% is reported under the '<18.5' category for the Class 2 mortality category. The lowest percentage of 0.1% is reported again under the '<18.5' category for the Class 3 mortality category.

Table 4.28 Chi-square test between BMI_RANGE and Mortality variables

| **Chi-Square Tests** |
|---|

|  | Value | df | Asymptotic Significance (2-sided) |
|---|---|---|---|
| Pearson Chi-Square | 1832.751[a] | 8 | <.001 |
| Likelihood Ratio | 1802.633 | 8 | <.001 |
| N of Valid Cases | 202905 |  |  |

a. 0 cells (0.0%) have expected count less than 5. The minimum expected count is 82.57.

As seen in Table 4.28, the Pearson Chi-Square value is 1832.751 with the degree of freedom as 8 at a p-value of <0.001. The null hypothesis ($H_0$) is rejected as the p-value is less than the significance level (0.05). This concludes that there is a significant association between the variables BMI_RANGE and Mortality.

13. A1C_RANGE

First of all, the count value in each cell of the crosstabulation shown in Table 4.29 is more than 5 and hence it meets the assumption for the Chi-square test. The A1C_RANGE is a variable that denotes the percentage of hemoglobin A1C. In Table 4.29, the highest percentage of 82.2% (166,699 records) is reported under the '<8' category of the A1C_RANGE variable within the mortality variable. By column-wise, the highest percentage of 78.5% is reported under the '<8' A1C_RANGE category for the Class 1 mortality category. The highest percentage of 81.5% is reported again under the '<8' category for the Class 2 mortality category. The highest percentage of 84.7% is also reported under the '<8' category of A1C_RANGE for the Class 3 mortality category. The lowest percentage are reported under the '>9' A1C category for all the mortality categories. The lowest percentage of 5.1% is reported under the '>9' category for the Class 3 mortality category.

Table 4.29 Crosstabulation between A1C_RANGE and Mortality variables

| Crosstabulation | | | | | | |
|---|---|---|---|---|---|---|
|  |  |  | Mortality | | | |
|  |  |  | Class 1 | Class 2 | Class 3 | Total |
| A1C_RANGE | <8 | Count | 37784 | 55610 | 73305 | **166699** |
|  |  | % within A1C_RANGE | 22.7% | 33.4% | 44.0% | 100.0% |
|  |  | % within Mortality | **78.5%** | **81.5%** | **84.7%** | **82.2%** |
|  | >9 | Count | 3868 | 4487 | 4428 | 12783 |
|  |  | % within A1C_RANGE | 30.3% | 35.1% | 34.6% | 100.0% |
|   |   | % within Mortality | **8.0%** | **6.6%** | **5.1%** | 6.3% |
|   | 8-9 | Count | 6490 | 8117 | 8816 | 23423 |
|   |   | % within A1C_RANGE | 27.7% | 34.7% | 37.6% | 100.0% |
|   |   | % within Mortality | 13.5% | 11.9% | 10.2% | 11.5% |
| Total |   | Count | 48142 | 68214 | 86549 | 202905 |
|   |   | % within A1C_RANGE | 23.7% | 33.6% | 42.7% | 100.0% |
|   |   | % within Mortality | 100.0% | 100.0% | 100.0% | 100.0% |

Table 4.30 Chi-square test between A1C_RANGE and Mortality variables

| Chi-Square Tests | | | |
|---|---|---|---|
|   | **Value** | **Df** | **Asymptotic Significance (2-sided)** |
| Pearson Chi-Square | 883.335[a] | 4 | <.001 |
| Likelihood Ratio | 876.073 | 4 | <.001 |
| N of Valid Cases | 202905 |   |   |
| a. 0 cells (0.0%) have expected count less than 5. The minimum expected count is 3032.94. | | | |

As seen in Table 4.30, the Pearson Chi-Square value is 883.335 with the degree of freedom as 4 at a p-value of <0.001. The null hypothesis ($H_0$) is rejected as the p-value is less than the significance level (0.05). This concludes that there is a significant association between the variables A1C_RANGE and Mortality.

14. SERUMCRE_RANGE

First of all, the count value in each cell of the crosstabulation shown in Table 4.31 is more than 5 and hence it meets the assumption for the Chi-square test. The SERUMCRE_RANGE is a variable that denotes the range of serum creatinine levels (in mg/dL). In Table 4.31, the highest percentage of 78.2% (158,662 records) is reported under the '<1.5' category of SERUMCRE_RANGE variable within the mortality variable. By column-wise, the highest percentage of 68.4% is reported under the '<1.5' category of SERUMCRE_RANGE for the

Class 1 mortality category. The highest percentage of 76.0% is reported again under the

'<1.5' category for the Class 2 mortality category. The highest percentage of 85.4% is also reported under the same '<1.5' category for the Class 3 mortality category. The lowest percentages are reported under the '>3.0' serum creatinine category for all the mortality categories. The lowest percentage of 0.2% is reported under the '>3.0' category for the Class 3 mortality category.

Table 4.31 Crosstabulation between SERUMCRE_RANGE and Mortality variables

| Crosstabulation | | | Mortality | | | Total |
|---|---|---|---|---|---|---|
| | | | Class 1 | Class 2 | Class 3 | |
| **SERUMCRE_RANGE** <1.5 | | Count | 32937 | 51841 | 73884 | **158662** |
| | | % within SERUMCRE_RANGE | 20.8% | 32.7% | 46.6% | 100.0% |
| | | % within Mortality | **68.4%** | **76.0%** | **85.4%** | **78.2%** |
| | >3.0 | Count | 1068 | 579 | 182 | 1829 |
| | | % within SERUMCRE_RANGE | 58.4% | 31.7% | 10.0% | 100.0% |
| | | % within Mortality | **2.2%** | **0.8%** | **0.2%** | 0.9% |
| | 1.5-3.0 | Count | 12303 | 12797 | 8636 | 33736 |
| | | % within SERUMCRE_RANGE | 36.5% | 37.9% | 25.6% | 100.0% |
| | | % within Mortality | 25.6% | 18.8% | 10.0% | 16.6% |
| | Missing | Count | 1834 | 2997 | 3847 | 8678 |
| | | % within SERUMCRE_RANGE | 21.1% | 34.5% | 44.3% | 100.0% |
| | | % within Mortality | 3.8% | 4.4% | 4.4% | 4.3% |
| Total | | Count | 48142 | 68214 | 86549 | 202905 |
| | | % within SERUMCRE_RANGE | 23.7% | 33.6% | 42.7% | 100.0% |
| | | % within Mortality | 100.0% | 100.0% | 100.0% | 100.0% |

Table 4.32 Chi-square test between SERUMCRE_RANGE and Mortality variables

| Chi-Square Tests | | | |
|---|---|---|---|
| | Value | df | Asymptotic Significance (2sided) |
| Pearson Chi-Square | 7415.854[a] | 6 | <.001 |
| Likelihood Ratio | 7415.768 | 6 | <.001 |

| N of Valid Cases | 202905 | | |

a. 0 cells (0.0%) have expected count less than 5. The minimum expected count is 433.96.

As seen in Table 4.32, the Pearson Chi-Square value is 7415.854 with the degree of freedom as 6 at a p-value of <0.001. The null hypothesis ($H_0$) is rejected as the p-value is less than the significance level (0.05). This concludes that there is a significant association between the two variables Mortality and SERUMCRE_RANGE.

## 15. HDL_RANGE

First of all, the count value in each cell of the crosstabulation shown in Table 4.33 is more than 5 and hence it meets the assumption for the Chi-square test. The HDL_RANGE is a variable that denotes the range of HDL cholesterol (in mg/dL). In Table 4.33, the highest percentage of 50.8% (102,986 records) is reported under the '<40' category of HDL_RANGE variable within the mortality variable. By column-wise, the highest percentage of 53.5% is reported under the '<40' category of HDL_RANGE for the Class 1 mortality category. The highest percentage of 51.7% is reported again under the '<40' category for the Class 2 mortality category. The highest percentage of 48.5% is also reported under the same '<40' category for the Class 3 mortality category. The lowest percentages are reported under the 'Missing' category for all the mortality classes. The lowest percentage of 2.5% is reported under the 'Missing' category for the Class 3 mortality category.

Table 4.33 Crosstabulation between HDL_RANGE and Mortality variables

| Crosstabulation | | | Mortality | | | Total |
|---|---|---|---|---|---|---|
| | | | Class 1 | Class 2 | Class 3 | |
| HDL_RANGE | <40 | Count | 25737 | 35270 | 41979 | **102986** |
| | | % within HDL_RANGE | 25.0% | 34.2% | 40.8% | 100.0% |
| | | % within Mortality | **53.5%** | **51.7%** | **48.5%** | **50.8%** |
| | >=60 | Count | 2532 | 3477 | 4525 | 10534 |
| | | % within HDL_RANGE | 24.0% | 33.0% | 43.0% | 100.0% |
| | | % within Mortality | 5.3% | 5.1% | 5.2% | 5.2% |
| | 40-59.99 | Count | 18437 | 27420 | 37876 | 83733 |
| | | % within HDL_RANGE | 22.0% | 32.7% | 45.2% | 100.0% |

|  |  | % within Mortality | 38.3% | 40.2% | 43.8% | 41.3% |
|---|---|---|---|---|---|---|
|  | Missing | Count | 1436 | 2047 | 2169 | 5652 |
|  |  | % within HDL_RANGE | 25.4% | 36.2% | 38.4% | 100.0% |
|  |  | % within Mortality | **3.0%** | **3.0%** | **2.5%** | **2.8%** |
| Total |  | Count | 48142 | 68214 | 86549 | 202905 |
|  |  | % within HDL_RANGE | 23.7% | 33.6% | 42.7% | 100.0% |
|  |  | % within Mortality | 100.0% | 100.0% | 100.0% | 100.0% |

Table 4.34 Chi-square test between HDL_RANGE and Mortality variables

| Chi-Square Tests | | | |
|---|---|---|---|
|  | Value | Df | Asymptotic Significance (2-sided) |
| Pearson Chi-Square | 464.621[a] | 6 | <.001 |
| Likelihood Ratio | 465.251 | 6 | <.001 |
| N of Valid Cases | 202905 |  |  |
| a. 0 cells (0.0%) have expected count less than 5. The minimum expected count is 1341.01. | | | |

As seen in Table 4.34, the Pearson Chi-Square value is 464.621 with the degree of freedom as 6 at p-value of <0.001. The null hypothesis ($H_0$) is rejected as the p-value is less than the significance level (0.05). This concludes that there is a significant association between the two variables HDL_RANGE and Mortality.

16. LDL_RANGE

First of all, the count value in each cell of the crosstabulation shown in Table 4.35 is more than 5 and hence it meets the assumption for the Chi-square test. The LDL_RANGE is a variable that denotes the range of LDL cholesterol (in mg/dL). In Table 4.35, the highest percentage of 61.2% (124,263 records) is reported under the '<100' category of the LDL_RANGE variable within the mortality variable. By column-wise, the highest percentage of 62.9% is reported under the '<100' LDL_RANGE category for the Class 1 mortality category. The highest percentage of 62.0% is reported again under the '<100' category for the

Class 2 mortality category. The highest percentage of 59.7% is also reported under the same

'<100' category for the Class 3 mortality category. The lowest percentages are reported under the '>=190' category for all the mortality classes. The lowest percentage of 0.1% is reported under the '>=190' category for Class 2 and Class 3 mortality categories. For Class 1 mortality, the lowest percentage is reported as 0.2% under the '>=190' category.

Table 4.35 Crosstabulation between LDL_RANGE and Mortality variables

| Crosstabulation | | | | | | |
|---|---|---|---|---|---|---|
| | | | Mortality | | | Total |
| | | | Class 1 | Class 2 | Class 3 | |
| LDL_RANGE | <100 | Count | 30263 | 42320 | 51680 | **124263** |
| | | % within LDL_RANGE | 24.4% | 34.1% | 41.6% | 100.0% |
| | | % within Mortality | **62.9%** | **62.0%** | **59.7%** | **61.2%** |
| | >=190 | Count | 75 | 98 | 126 | 299 |
| | | % within LDL_RANGE | 25.1% | 32.8% | 42.1% | 100.0% |
| | | % within Mortality | **0.2%** | **0.1%** | **0.1%** | 0.1% |
| | 100-129.99 | Count | 11526 | 16615 | 23339 | 51480 |
| | | % within LDL_RANGE | 22.4% | 32.3% | 45.3% | 100.0% |
| | | % within Mortality | 23.9% | 24.4% | 27.0% | 25.4% |
| | 130-159.99 | Count | 2750 | 3965 | 5345 | 12060 |
| | | % within LDL_RANGE | 22.8% | 32.9% | 44.3% | 100.0% |
| | | % within Mortality | 5.7% | 5.8% | 6.2% | 5.9% |
| | 160-189.99 | Count | 512 | 700 | 941 | 2153 |
| | | % within LDL_RANGE | 23.8% | 32.5% | 43.7% | 100.0% |
| | | % within Mortality | 1.1% | 1.0% | 1.1% | 1.1% |
| | Missing | Count | 3016 | 4516 | 5118 | 12650 |
| | | % within LDL_RANGE | 23.8% | 35.7% | 40.5% | 100.0% |
| | | % within Mortality | 6.3% | 6.6% | 5.9% | 6.2% |
| Total | | Count | 48142 | 68214 | 86549 | 202905 |
| | | % within LDL_RANGE | 23.7% | 33.6% | 42.7% | 100.0% |
| | | % within Mortality | 100.0% | 100.0% | 100.0% | 100.0% |

Table 4.36 Chi-square test between LDL_RANGE and Mortality variables

| Chi-Square Tests | | | |
|---|---|---|---|
| | Value | Df | Asymptotic Significance (2-sided) |
| Pearson Chi-Square | 260.499[a] | 10 | <.001 |
| Likelihood Ratio | 259.830 | 10 | <.001 |
| N of Valid Cases | 202905 | | |

a. 0 cells (0.0%) have expected count less than 5. The minimum expected count is 70.94.

As seen in Table 4.36, the Pearson Chi-Square value is 260.499 with the degree of freedom as 10 at a p-value of <0.001. The null hypothesis ($H_0$) is rejected as the p-value is less than the significance level (0.05). This concludes that there is a significant association between the two variables LDL_RANGE and Mortality.

17. TRI_RANGE

First of all, the count value in each cell of the crosstabulation shown in Table 4.37 is more than 5 and hence it meets the assumption for the Chi-square test. The TRI_RANGE is a variable that denotes the range of Triglycerides (in mg/dL). In Table 4.37, the highest percentage of 53.0% (107,560 records) is reported under the '<150' category of the TRI_RANGE variable within the mortality variable. By column-wise, the highest percentage of 53.2% is reported under the '<150' category of the TRI_RANGE for the Class 1 mortality category. The highest percentage of 52.2% is reported again under the '<150' category for the

Class 2 mortality category. The highest percentage of 53.5% is also reported under the same

'<150' category for the Class 3 mortality category. The lowest percentages are reported under the 'Missing' category for all the mortality classes. The lowest percentage of 2.4% is reported under the 'Missing' category for Class 3 mortality category.

Table 4.37 Crosstabulation between TRI_RANGE and Mortality variables

| Crosstabulation | | | | | | |
|---|---|---|---|---|---|---|
| | | | Mortality | | | |
| | | | Class 1 | Class 2 | Class 3 | Total |
| TRI_RANGE <150 | | Count | 25632 | 35626 | 46302 | **107560** |
| | | % within TRI_RANGE | 23.8% | 33.1% | 43.0% | 100.0% |
| | | % within Mortality | **53.2%** | **52.2%** | **53.5%** | **53.0%** |
| | >=200 | Count | 11858 | 17117 | 20718 | 49693 |
| | | % within TRI_RANGE | 23.9% | 34.4% | 41.7% | 100.0% |
| | | % within Mortality | 24.6% | 25.1% | 23.9% | 24.5% |
| | 150-199.99 | Count | 9327 | 13534 | 17478 | 40339 |
| | | % within TRI_RANGE | 23.1% | 33.6% | 43.3% | 100.0% |
| | | % within Mortality | 19.4% | 19.8% | 20.2% | 19.9% |
| | Missing | Count | 1325 | 1937 | 2051 | 5313 |
| | | % within TRI_RANGE | 24.9% | 36.5% | 38.6% | 100.0% |
| | | % within Mortality | **2.8%** | **2.8%** | **2.4%** | 2.6% |
| Total | | Count | 48142 | 68214 | 86549 | 202905 |
| | | % within TRI_RANGE | 23.7% | 33.6% | 42.7% | 100.0% |
| | | % within Mortality | 100.0% | 100.0% | 100.0% | 100.0% |

Table 4.38 Chi-square test between TRI_RANGE and Mortality variables

| Chi-Square Tests | | | |
|---|---|---|---|
| | **Value** | **Df** | **Asymptotic Significance (2-sided)** |
| Pearson Chi-Square | 80.600[a] | 6 | <.001 |
| Likelihood Ratio | 80.939 | 6 | <.001 |
| N of Valid Cases | 202905 | | |
| a. 0 cells (0.0%) have expected count less than 5. The minimum expected count is 1260.58. | | | |

As seen in Table 4.38, the Pearson Chi-Square value is 80.600 with the degree of freedom as 6 at a p-value of <0.001. The null hypothesis ($H_0$) is rejected as the p-value is less than the significance level (0.05). This concludes that there is a significant association between the two variables TRI_RANGE and Mortality at the 5% level of significance.

Thus, among the 17 input variables analysed with the target variable 'Mortality' using the chisquare test, it was concluded that only the 'SEX' variable is not associated with the target variable 'Mortality'. Therefore, this variable would be dropped.

### 4.4.2 Additional Chi-Squared Test between Two Input Variables

Two additional chi-square tests are being conducted for the variables mentioned in Table 4.39. The hypothesis testing that is considered while analysing these chi-square test results is as follows:

- Null Hypothesis ($H_0$): There is no association between the two variables (both the variables are independent)
- Alternate Hypothesis ($H_1$): There is an association between the two variables (both the variables are dependent)

Table 4.39 Additional Case Processing Summary for 4 input variables

|  | Cases | | | | | |
|---|---|---|---|---|---|---|
|  | Valid | | Missing | | Total | |
|  | N | Percent | N | Percent | N | Percent |
| PRIORITY * FRAILTY_GROUP | 202905 | 100.0% | 0 | 0.0% | 202905 | 100.0% |
| MARRIED * RACE | 202905 | 100.0% | 0 | 0.0% | 202905 | 100.0% |

1. PRIORITY and FRAILTY_GROUP

As seen in Table 4.40, the Pearson Chi-Square value is 6576.316 with the degree of freedom as 16 at a p-value of <0.001. The null hypothesis ($H_0$) is rejected as the p-value is less than the significance level (0.05). This concludes that there is a significant association between the two variables PRIORITY and FRAILTY_GROUP. These two input variables are dependent and hence one of them needs to be dropped.

Table 4.40 Chi-square test between PRIORITY and FRAILTY_GROUP variables

| **Chi-Square Tests** | | | |
|---|---|---|---|
|  | **Value** | **Df** | **Asymptotic Significance (2-sided)** |
| Pearson Chi-Square | 6576.316[a] | 16 | <.001 |
| Likelihood Ratio | 6511.355 | 16 | <.001 |
| N of Valid Cases | 202905 | | |
| a. 0 cells (0.0%) have expected count less than 5. The minimum expected count is 9.05. | | | |

As seen previously in Table 4.6, for the PRIORITY variable with the Mortality target variable the Pearson Chi-Square value is 4139.033. Similarly, as seen in Table 4.16, for the FRAILTY_GROUP variable with the Mortality target variable the Pearson Chi-Square value is 1252.936. As the chi-square value of PRIORITY is higher, the PRIORITY variable has been kept and the FRAILTY_GROUP variable has been decided to be dropped.

2.  MARRIED and RACE

As seen in Table 4.41, the Pearson Chi-Square value is 3138.669 with the degree of freedom as 4 at a p-value of <0.001. The null hypothesis ($H_0$) is rejected as the p-value is less than the significance level (0.05). This concludes that there is a significant association between the two variables MARRIED and RACE. These two input variables are dependent and hence one of them needs to be dropped.

Table 4.41 Chi-square test between MARRIED and RACE variables

| **Chi-Square Tests** | | | |
| --- | --- | --- | --- |
| | Value | Df | Asymptotic Significance (2-sided) |
| Pearson Chi-Square | 3138.669[a] | 4 | <.001 |
| Likelihood Ratio | 2825.862 | 4 | <.001 |
| N of Valid Cases | 202905 | | |
| a. 0 cells (0.0%) have expected count less than 5. The minimum expected count is 1005.25. | | | |

As seen previously in Table 4.8, for the MARRIED variable with the Mortality target variable the Pearson Chi-Square value is 437.131. Similarly, as seen in Table 4.12, for the RACE variable with the Mortality target variable the Pearson Chi-Square value is 153.508. As the chi-square value of the MARRIED variable is higher, the MARRIED variable has been kept and the RACE variable has been decided to be dropped.

Finally, after performing bivariate analysis, it is concluded that the 3 input categorical variables - SEX, RACE, and FRAILTY_GROUP, would be dropped from the dataset.

**4.5  Data Encoding**

Categorical variables related to the demographics, clinical biomarkers and the target variable itself in this study, have non-numeric categories as seen during the univariate and bivariate analysis in previous sections. There are in total 14 such input categorical variables. The categories in these variables need to be represented as numerical so that they (Meng and Xing, 2022)can be given as input to various machine learning algorithms. Numerical representation is necessary so that machine learning algorithms can process categorical data effectively.

### 4.5.1 Dummy Encoding

One-hot encoding technique is also known as the process of creating dummy variables. By using this technique, categorical variables are transformed into binary vectors and each category is represented by a binary indicator (0 or 1) (Meng and Xing, 2022). Every category present in a categorical variable becomes a separate variable (or feature) in the dataset. Thus, if there are k categories in a categorical variable, k resultant columns get added to the dataset. This is a disadvantage as it leads to a 'Dummy Variable Trap' (Mukherjee et al., 2022).

Instead, the dummy encoding technique helps to avoid the 'Dummy Variable Trap' as only k-1 variables are created from k categories in a variable (Mukherjee et al., 2022). This conversion helps to interpret each category's impact on the predictions done by the model. This approach has been considered for this dataset even in the previous study, as Griffith et al., 2020 have mentioned the adjusted odds ratios for the relevant variables associated with mortality in their research paper.

Dummy variables are created by using the get_dummies() method in Python. The dataset and the columns to be dummy encoded are given as parameters in this method. There is another parameter 'drop_first' which is by default set to 'False'. If this parameter is set to 'True' it drops the first level and thus k-1 resultant columns get added for k categories. But in this study, this parameter is kept as the default value (false) to manually select the dummy columns to be dropped rather than dropping the first dummy column. The variable names mentioned below are removed from the dataset.

- After bivariate analysis – SEX, RACE, FRAILTY_GROUP
- After dummy encoding, the following dummy variables have been dropped due to the mentioned reasons.
- Very less number of instances - PRIORITY_Unknown
- A high number of instances in all target classes - MARRIED_MARRIED, N_OP_RANGE_6-30, BMI_RANGE_25-39.9 o Lowest range - A1C_RANGE_<8, AGE_GROUP_65-69, N_IP_RANGE_0-5 o Normal range - DIASTOLIC_RANGE_<80, SYSTOLIC_RANGE_<120, SERUMCRE_RANGE_<1.5, HDL_RANGE_40-59.99, LDL_RANGE_<100, TRI_RANGE_<150, SERUMALB_RANGE_>=3.5

The creation of dummy variables and removing the above-mentioned variables has been performed on both the training as well as testing dataset. Finally, there are 96 input variables and a target variable 'Mortality', in both the training as well as the testing dataset.

### 4.5.2 Label Encoding

Label encoding of the target variable is used when the objective is to convert a categorical target variable having non-numeric values into numerical form. This process is to assign a unique numerical label to each category present in the target variable. In this study, a dictionary has been created in Python with the encoding details and target variable values are encoded using the replace() method with the dictionary as a parameter. 'Class 1' is now 0,

'Class 2' is 1, and 'Class 3' is 2. Again, this operation has been performed on both the training as well as the testing dataset.

## 4.6 Class Balancing

Class balancing was done on the training dataset which has 202,905 records. The reason for doing this is to balance out the classes present in the target variable 'Mortality' so that the model gets trained for all the classes equally and effectively. Under-sampling of majority classes has been performed using the 'RandomUnderSampler' from the Python library imblearn.under_sampling. The parameter sampling_strategy is set as 'not minority' so that the minority class is not under-sampled. Table 4.42 shows the distribution of classes in the training dataset before and after under-sampling. Under-sampling has been done on the training dataset only. Under-sampling helps reduce the number of samples belonging to the majority classes (Class 2 and Class 3) and focus on the minority class (Class 1).

Table 4.42 Distribution of classes in the training dataset before and after under-sampling

|  | Training Dataset | Class 1 (0) | Class 2 (1) | Class 3 (2) |
|---|---|---|---|---|
| **Before under-sampling** | 202,905 | 48,142 (23.7%) | 68,214 (33.6%) | 86,549 (42.7%) |
| **After under-sampling** | 144,426 | 48,142 (33.33%) | 48142 (33.33%) | 48142 (33.33%) |

To further use the training dataset for the development of various models, the under-sampled training dataset having 96 input variables and a target variable has been exported as a separate .csv file using the to_csv() Python method. The testing dataset is also exported as a separate .csv file. The testing dataset is not under-sampled.

## 4.7 Model Implementation

First, a benchmark model has been implemented using the R programming language in RStudio software. This benchmark model consists of the Least Absolute Shrinkage and Selection Operator (LASSO) feature selection technique. Every other model implementation has been done with the filter-based feature selection techniques (Chi-Square Statistics Test and Information Gain) using the Python programming language in separate Jupyter notebooks launched through Anaconda software. The machine learning algorithms and the configurations used for each model are explained in the following sections.

Additionally, it is to be noted that before the models were built using the under-sampled training data (144,426 records), some other configurations were also tried. The models were trained on default parameters of classifiers but gave poor results. Also, instead of the undersampled training dataset, initially, the complete training data (202,905 records) was used to train the models. The overall performance results for multiclass classification given by these models, were slightly lower than what the models mentioned in the following sections have given. Hence, under-sampled data was finalized to train the models with the configurations mentioned in the following sections.

### 4.7.1 Benchmark Model – Multinomial Logistic Regression with LASSO

The working directory is set using the setwd() function in RStudio software. The file path where the training and testing datasets (exported as mentioned in section 4.6) are placed is given as the function argument. These datasets are imported into separate data frames using the read.csv() function in R. In the training data frame, there are 97 attributes (96 input variables and 1 target variable 'Mortality') with 144,426 records. In the testing data frame, there are 97 attributes (96 input variables and 1 target variable 'Mortality') with 67,635 records. The 96 input variables present in the training data frame have been assigned to a matrix of variables and the target variable is assigned to another, to differentiate the target variable from the rest of the input variables. Similarly, this has been performed with the testing data frame.

The method cv.glmnet() from the glmnet R package is used to build and train a regression model. The separated input and target training data are the first two arguments in cv.glmnet(). The rest of the arguments are configured as, family is set as "multinomial", parallel is TRUE, type.measure is "class", and type.multinomial is "grouped". The LASSO regularization is performed by default as the default alpha value is 1. The model is then run with 10-fold crossvalidation as the number of folds is 10 by default. The coef() is used to get the coefficients of each feature and see which features have been penalized to 0 by LASSO in this model. Prediction of the testing data is done directly while computing the confusion matrix, as described in section 4.8.

### 4.7.2 Models Using Multiclass Classifiers

The training and testing data files exported as mentioned in section 4.6, are first imported into separate data frames using the read_csv() pandas function in Python. In the training data frame, there are 97 attributes (96 input variables and 1 target variable 'Mortality') with 144,426 records. In the testing data frame, there are 97 attributes (96 input variables and 1 target variable 'Mortality') with 67,635 records. The 96 input variables present in the training data frame have been assigned to 'X_train' data frame and the target variable is assigned to the 'y_train' data frame to differentiate the target variable from the rest of the input variables. Similarly, this has been performed with the testing data frame. The 'X_test' data frame now has the 96 input variables and 'y_test' has the target variable from the testing data frame.

Now, as the number of input variables is quite high, feature selection is performed to select a certain number of variables that are most relevant and helpful in predicting the target variable.

The filter-based feature selection techniques, Chi-square statistics and information gain (or mutual information) have been employed. To perform the feature selection, 'SelectKBest',

'chi2', and 'mutual_info_classif' are imported from sklearn.feature_selection Python module.

The parameter score_func is assigned 'chi2' for feature selection using the Chi-Square Statistics technique in Python. The 'SelectKBest' function selects the k features as per the k highest chi-squared scores (Puneet and Chauhan, 2020). To select features using the information gain technique, parameter score_func is assigned 'mutual_info_classif'. Both times the k parameter in the SelectKBest() is assigned the value of 66 to select the top 66 features out of 96. The value '66' has been decided as 66 features were selected by LASSO mentioned in the previous section.

Finally, the chi-square feature selection function is fit to the training data and a transformed version with the reduced features is returned by the fit_transform(X_train,y_train) and assigned to 'X_train_ch'. Only transform(X_test) is done to directly reduce the features present in X_test to the selected features and assign them to 'X_test_ch' without fitting the testing data. Similar steps have been performed with the information gain (or mutual information) function. The training data with reduced features is assigned to 'X_train_ig' and the testing data with reduced feature is assigned to 'X_test_ig'. These new train and test sets have been used further while training and testing various models.

Hyperparameter tuning is performed individually for the Multinomial Logistic Regression, Random Forest, and XGBoost, using the GridSearchCV() function after importing it from the sklearn.model_selection module in Python. The values given to the estimator and the param_grid parameters have been discussed for each model in the respective sections below.

The scoring parameter is set as 'accuracy'. The verbose is set as 0 to avoid the display of messages. The parameter cv which is for cross-validation has been explained next.

StratifiedKFold() function imported again from the sklearn.model_selection module is used with parameter n_splits as 2 while tuning the hyperparameters using grid search crossvalidation. In the StratifiedKFold() function, parameter n_splits is assigned the value of 10 while training the models. Other parameters in StratifiedKFold() are set as, random_state as 1 and shuffle as True. In this study, the stratified k-fold function provides the indices to split the training sets (X_train_ch and X_train_ig) into train and validation sets with each of the kfolds having the same percentage of samples of each class. Thus, 10 folds get created during model training from which the model is trained on 9 folds and validated on the remaining fold. This is repeated until all the folds have been used as a validation set in the respective iteration.

### 4.7.2.1 Multinomial Logistic Regression

While building the Multinomial Logistic Regression model, the best parameters are searched using the grid search cross-validation discussed previously. LogisticRegression() classifier imported from sklearn.linear_model Python module is used as an estimator and the n_jobs parameter is set as '-1' to use all processors, in GridSearchCV(). In this classifier, the multi_class parameter is set to 'multinomial' and random_state is set to 42. The rest of the parameters take the default value even if not mentioned explicitly. Then multiple parameters are passed as a Python dictionary to the param_grid parameter for grid search CV.

- For building a model with features selected by the Chi-Square Statistic technique, parameters = {'penalty': ['l2', None], 'C': [0.1,1.0,10.0], 'solver': ['newton-cg', 'sag', 'saga', 'lbfgs'], 'class_weight': ['balanced', None]}

Then, after the grid search CV is performed, the best parameters out of the parameters dictionary are found by using best_params_ attribute. The best parameters are considered finally for building a model. So, the parameters multi_class as 'multinomial', C as 1.0, class_weight as 'balanced', penalty as 'l2', solver as 'newton-cg', and random_state as 42, are then given to the LogisticRegression() classifier. Rest of the parameters take the default value even if not mentioned explicitly. Further, the classifier is trained by fitting on the 9 folds and it is then validated on the remaining fold. This happens 10 times with different fold validated every time. The predict() function is then used with the trained model to predict the outcomes for the samples in the hold-out test set (X_test_ch).

- For building a model with features selected by the information gain (or mutual information) technique, parameters = {'penalty': ['l2', None], 'C': [0.01,0.1,1.0], 'solver': ['newton-cg', 'sag', 'saga', 'lbfgs'], 'class_weight': ['balanced', None]}

Then, after the grid search CV is performed, the best parameters out of the parameters dictionary are found by using best_params_ attribute. The best parameters are considered finally for building a model. So, the parameters multi_class as 'multinomial', C as 0.1, class_weight as 'balanced', penalty as 'l2', solver as 'lbfgs', and random_state as10, are then given to the LogisticRegression() classifier. The rest of the parameters take the default value even if not mentioned explicitly. Further, the classifier is trained by fitting on the 9 folds and it is then validated on the remaining folds. This happens 10 times with a different fold validated every time. The predict() function is then used with the trained model to predict the outcomes for the samples in the hold-out test set (X_test_ig).

**4.7.2.2 Random Forest**

While building the Random Forest model, the best parameters are searched using the grid search cross-validation discussed previously. RandomForestClassifier() imported from sklearn.ensemble Python module, is used as estimator in GridSearchCV(). In this classifier, the random_state parameter is set to 42. Rest of the parameters take the default value even if not mentioned explicitly. Then, multiple parameters are passed as a Python dictionary to the param_grid parameter for grid search CV.

- For building a model with features selected by the Chi-Square Statistic technique, parameters = {'n_estimators': [300, 500], 'max_depth': [6,10], 'min_samples_leaf': [3, 4, 5]}

Then, after the grid search CV is performed, the best parameters out of the parameters dictionary are found by using best_params_ attribute. The best parameters are considered finally for building a model.

So, the parameters n_estimators as 500, max_depth as 10, min_samples_leaf as 4, and random_state as 42, are then given to the RandomForestClassifier() classifier. The rest of the parameters take the default value even if not mentioned explicitly. Further, the classifier is trained by fitting on the 9 folds and it is then validated on the remaining fold. This happens 10 times with different fold validated every time. The predict() function is then used with the trained model to predict the outcomes for the samples in the hold-out test set (X_test_ch).

- For building a model with features selected by the information gain (or mutual information) technique, parameters = {'n_estimators': [500], 'max_depth': [6,10]}

Then, after the grid search CV is performed, the best parameters out of the parameters dictionary are found by using best_params_ attribute. The best parameters are considered finally for building a model. So, the parameters max_depth as 10, n_estimators as 500, and random_state as 42, are then given to the RandomForestClassifier() classifier. The rest of the parameters take the default value even if not mentioned explicitly. Further, the classifier is trained by fitting on the 9 folds and it is then validated on the remaining folds. This happens 10 times with different fold validated every time. The predict() function is then used with the trained model to predict the outcomes for the samples in the hold-out test set (X_test_ig).

**4.7.2.3 XGBoost**

XGBoost models have been built, trained, and tested on Google Colab. While building the XGBoost model, the best parameters are searched using the grid search cross-validation discussed previously. XGBClassifier() imported from xgboost module, is used as estimator in GridSearchCV(). In this classifier, the tree_method parameter is set as 'gpu_hist' and the random_state parameter is set to 42. Rest of the parameters take the default value even if not mentioned explicitly. So, the estimator used by XGBoost by default is Then, multiple parameters are passed as a Python dictionary to the param_grid parameter for grid search CV.

- For building a model with features selected by the Chi-Square Statistic technique, parameters = {'learning_rate': [0.01, 0.1], 'colsample_bytree': [0.8, 1.0], 'n_estimators': [500,1000]}

Then, after the grid search CV is performed, the best parameters out of the parameters dictionary are found by using best_params_ attribute. The best parameters are considered finally for building a model. So, the parameters tree_method as "gpu_hist", colsample_bytree as 0.8, learning_rate as 0.01, n_estimators as 1000, and random_state as 42, are then given to the XGBClassifier() classifier. Rest of the parameters take the default value even if not mentioned explicitly. Further, the classifier is trained by fitting on the 9 folds and it is then validated on the remaining fold. This happens 10 times with different fold validated every time. The predict() function is then used with the trained model to predict the outcomes for the samples in the hold-out test set (X_test_ch).

- For building a model with features selected by the information gain (or mutual information) technique, parameters = {'learning_rate': [0.01, 0.1], 'colsample_bytree': [0.8, 1.0], 'n_estimators': [500,1000]}

Then, after the grid search CV is performed, the best parameters out of the parameters dictionary are found by using best_params_ attribute. The best parameters are considered finally for building a model. So, the parameters tree_method as "gpu_hist", colsample_bytree as 0.8, learning_rate as 0.01, n_estimators as 1000, and random_state as 42, are then given to the XGBClassifier() classifier. The rest of the parameters take the default value even if not mentioned explicitly. Further, the classifier is trained by fitting on the 9 folds and it is then validated on the remaining folds. This happens 10 times with a different fold validated every time. The predict() function is then used with the trained model to predict the outcomes for the samples in the hold-out test set (X_test_ig).

**4.7.2.4 One-Vs-Rest Classifier**

The OneVsRestClassifier() is a classifier that builds individual binary classifiers per target class. While building the One vs Rest model, the target variable labels in both the training as well as testing data have been first binarized using the label_binarize() function. This function is imported from the sklearn.preprocessing Python module. This function binarizes the integer labels present in the target variable data which is passed as a parameter along with the array of classes as [0, 1, 2], used in this study. Thus, 0 is denoted as [1, 0, 0]. 1 is denoted as [0, 1, 0] and 2 is denoted as [0, 0, 1].

The binary classifier for a particular class considers that class as a positive class and the rest of the classes are considered as the negative class. This is done to extend the binary classification to multiclass classification by using the one-vs-the-rest strategy. This strategy is implemented by the OneVsRestClassifier() classifier imported from sklearn.multiclass Python module. For each class, this classifier fits an individual separate classifier given as an estimator.

In this classifier, the estimator is the Logistic Regression classifier. The parameters used in this estimator are class_weight as 'balanced', penalty as None, and random_state as 42. The rest of the parameters take the default value even if not mentioned explicitly. The 10-fold cross-validation is employed using the KFold() function imported from the sklearn.model_selection Python module. The parameters given are n_splits as 10, random_state as1, and shuffle as True. The One-Vs-Rest Classifier is fitted on each training fold data and validated on the validation fold. The distance of each sample data from the class's decision boundary is returned by the decision_function() with input test data as the parameter in it. The decision scores computed for the input test data and the true labels of the test data are then used by the roc_curve() function. This has been explained in section 4.8.

Further, the predict_proba() function with the input testing data given as a parameter, predicted the probability of every test data sample for all the 3 classes. Then the predict() function with the fitted model is used by passing the input testing data as the parameter in the function. As binary classification is carried out, there are chances that some test data samples might get classified as a positive class by more than one binary classifier in the One-Vs-Rest Classifier. There can also be a possibility that the test data sample is not classified as a positive class by any of the binary classifiers. Therefore, the final class is predicted using the argmax() Python function by passing the parameters, predicted probabilities for the testing data and the axis as 1. This function gives the position index of the highest probability among the 3 probabilities computed by the respective binary classifier for every testing data sample.

## 4.8   Performance Evaluation of Models

The performance of all the models that have been built is evaluated using certain evaluation metrics. For the benchmark model, confusion.glmnet() method has been used for creating the confusion matrix. The trained model, the input testing data (newx), and the actual target values of testing data (newy) are given as parameters in this function. It returns a table of confusion matrix along with the percent correct score which has been displayed using the print() method. The predicted labels with their total are represented by the rows and the true (actual) labels with their total are represented by the columns. The entry in the $i^{th}$ row and $j^{th}$ column shows the number of samples that have the actual label as $j^{th}$ class and are predicted as $i^{th}$ class.

For the other models built, accuracy_score, confusion_matrix, and classification_report are imported from sklearn.metrics Python module. The accuracy score is then calculated by accuracy_score() function. The confusion matrix which shows how the predicted classes are classified is formed using the confusion_matrix() function. This function returns an array of dimensions (3,3) as there are three target classes. The i rows are the true labels and the j columns are the predicted labels. The entry in the $i^{th}$ row and $j^{th}$ column shows the number of samples that have the actual label as $i^{th}$ class and are predicted as $j^{th}$ class. This array is converted to a data frame and then a heatmap is plotted using the Seaborn library to visualize the confusion matrix.

The classification report which shows the main classification metrics such as precision, recall, f1-score, and accuracy as well as the macro and weighted averages is built by using the classification_report() function. In all three functions, accuracy_score, confusion_matrix, and classification_report, the first parameter is given as the y_test which has the actual (correct) outcomes of the testing dataset and the second parameter is given as the predicted outcomes by a model on the testing dataset.

Additionally, for the model built using the One-Vs-Rest Classifier, the Receiver Operating Characteristic (ROC) curves for each target class is plotted before computing the accuracy, confusion matrix, and classification report. The roc_curve() and auc() functions have been imported from the sklearn.metrics Python module. The true outcomes of testing data and the decision scores calculated by the model are passed as parameters to the roc_curve() function. The false positive rate (FPR) and the true positive rate (TPR) returned by the roc_curve() function are stored in Python dictionaries. The FPR and TPR are passed to the auc() function to calculate the ROC. This is computed for each target class by iterating over a 'for' loop. The ROC, accuracy, confusion matrix, and classification report are calculated for each class by considering the other two classes as the 'rest' class in the one vs rest implementation by the One-Vs-Rest Classifier.

Further, cohen's kappa value is calculated for the multiclass classification models, first between the actual class labels of the testing dataset and the predicted labels by the models. Then, the level of agreement has been checked between the models selected after evaluating the other performance metrics. Then, these kappa values have been interpreted. To calculate the kappa score, the cohen_kappa_score() function is imported from the sklearn.metrics (Parida et al., 2021).

## 4.9    Feature Analysis in Python using Chi-Square Statistics

Features analysis using the chi2() feature selection Python function has been conducted to see the association between the 96 independent input features and the dependent target variable 'Mortality', present in the training data. A data frame is created with all the 96 input feature names as the column names and just one row with the index 'Mortality'. If the p-value returned by the chi2() function for the chi-square test conducted between an input variable and the target variable is not less than 0.05 then the p-value is directly updated as 1 for that input feature in the data frame to denote no association between the variables. Otherwise, the p-value is updated in the data frame as it is for that input feature to denote the association between the variables. This has been done to differentiate the associated and the nonassociated variables when a data frame is viewed in the form of a heatmap.

But every independent feature may not be relevant in predicting all the target classes. Therefore, the chi2() function is also used to check the association between each of the independent input features and each class of the target variable 'Mortality'. To do this, first of all, three copies of the training data frame have been created using the copy() Python function.

Then in the first copy, the 'Class 1' in the target variable which is actually denoted as '0', is replaced as 1 and the remaining two classes have been replaced as 0. Then in the second copy, the 'Class 2' in the target variable which is actually denoted as '1', is kept as 1 only and the remaining two classes have been replaced as 0. Similarly, in the third copy, the 'Class 3' in the target variable which is actually denoted as '2', is replaced as 1 and the remaining two classes have been replaced as 0. The p-values have been computed and updated for every class in the respective data frame, in the same way as mentioned earlier. The association of every feature with the target is analysed.

The chi2() function also returns a chi-squared score which indicates how high the association is between the variables being analysed. Thus, another data frame is created with the input feature names as row indexes and the target variable, 'Class 1', Class 2', and 'Class 3', as the column names. The chi-squared value for every combination of the input feature and the target is updated in this data frame. Then a heatmap is plotted to visualize this data frame and see the chi-squared score for every input feature. The parameter 'annot' is set as True to see the chi-squared scores on the colour and parameter 'fmt' is set as 'g' to display the value without scientific notation. The higher the chi-squared score, the higher the association level between the variables that are compared (Tislenko et al., 2022). The lower the chi-squared score, the lower the association level between the variables that are compared. A heatmap is a color-coded matrix which helps to easily visualize and interpret the difference in the values due to the colours.

**4.10    Resources Required**

This study has been implemented on a laptop having a 64-bit Windows 11 operating system. It has a 12th Generation Intel i5 Core (1.30 GHz) processor, 16 GB RAM, and Intel IRIS Xe Graphics. So, this is the hardware requirement.

IBM SPSS Statistics 29.0 has been installed for conducting Chi-Square Test as part of bivariate analysis. The RStudio desktop application has been installed and used for building the benchmark model using R programming. The Anaconda distribution is used to install Jupyter and Python. The rest of the models are built using the Python programming language in Jupyter Notebook (version 6.5.2) with a Python 3.10.9 version. The Jupyter Notebook is launched by Anaconda Navigator (version 2.4.0). Python machine learning packages and libraries such as NumPy, Pandas, Matplotlib, Seaborn, and Scikit-learn are already available and just need to be imported. Only 'xgboost' needs to be installed if using Anaconda. In this study, the XGBoost model has been run on Google Colaboratory (or Google Colab) for GPU and hence installation of 'xgboost' was not required.

Overall, this section gives information about the analysis of the mortality dataset and the designing of various models for multiclass classification. At the start of the analysis, there were two target variables which have been converted to a single target. In total, there were 68 independent features and 275,190 records. After checking for duplicates, 4650 duplicate records were deleted and therefore 270,540 records remained. Six features had missing values but only the 'MICROALB' feature has been dropped as it had a very high percentage (74.09%) of missing values. Missing values and outliers were identified and handled by transforming the continuous numeric variables into categorical variables. This process is called binning. Thus, the missing values in the remaining five features have been kept as a separate category.

Univariate analysis was conducted to see the distribution of the values in each of the variables present in the complete cleaned dataset which had 270,540 records. Next, the cleaned mortality dataset was divided into training and testing sets, with the stratify parameter as the target variable to preserve the same distribution of the target classes. The training dataset had 202,905 records (75% of the total) and the testing dataset had 67,635 records (25% of the total). Further, the bivariate analysis was done on the training dataset using the Chi-square test of independence in the IBM SPSS software. After analysing the chi-square test results, three input categorical variables which were SEX, RACE, and FRAILTY_GROUP, were dropped.

Further, the categorical input variables in both the training as well as testing datasets were dummy encoded to make each category a separate variable in the datasets. This encoding was employed to interpret each category's impact on the predictions done by the model. After this, there were in total 97 attributes (96 input variables and 1 target variable 'Mortality') in both the training as well as the testing datasets. The target variable categories were also encoded and denoted in numeric form. 'Class 1' is now 0, 'Class 2' is 1, and 'Class 3' is 2.

Next, under-sampling was conducted for class balancing as there were uneven number of samples in the target classes of the training dataset. Thus, after under-sampling, the training dataset had 144,426 records. In the model development phase, first, a benchmark model which is a regression model was built using Multinomial Logistic Regression with LASSO in the RStudio desktop application. The rest of the models were built using Python programming in Jupyter Notebooks with the Anaconda Navigator desktop graphical user interface (GUI). As part of the proposed methodology, filter-based feature selection techniques such as Chisquared statistics and Information Gain (or Mutual Information) were used to select the features most relevant for classification.

Further, various classifiers such as the Multinomial Logistic Regression, Random Forest, XGBoost, and the One-Vs-Rest Classifier, were explained with their configurations for multiclass classification. Hyperparameter tuning was employed using the grid search crossvalidation. 10-fold cross-validation was also used and its details were elaborated. Then, various metrics considered to evaluate the performance of the models constructed were discussed. Lastly, an additional analysis was conducted using the Chi-Square Statistics function in Python to check the association for each of the 96 independent input features with the target variable to gain more insights.

## 5. RESULTS AND DISCUSSIONS

This section includes the output results and the findings gathered from them. The testing dataset has a total of 67,635 records. The different models that were designed and trained on the training data have predicted the target class for every sample of the testing dataset. First, the result of the benchmark model using Multinomial Logistic Regression with LASSO is analysed. The confusion matrix computed is studied to see how multiclass classification has been done by this regression model. Then, the models built by considering the features selected by the Chi-square Statistics and Information Gain feature selection techniques have been analysed. The models are built with Multinomial Logistic Regression, Random Forest, XGBoost, and One-Vs-Rest Classifier.

Evaluation of all these models has been conducted based on their individual performance metrics such as accuracy, recall, precision, and f1-score. ROC has also been computed with the One-Vs-Rest Classifier as it performs binary classification for each class by using the onevs-rest strategy. The results of the confusion matrix plotted for every multiclass classification performed by various models have been elaborated. Further, insights have been gathered by analysing the association of the input features to the target classes.

## 5.1 Evaluation of the Benchmark Model

The benchmark model constructed using Multinomial Logistic Regression with LASSO has given an accuracy of 0.529 after predicting the class outcomes for every sample present in the hold-out test dataset which has 67,635 records. Table 5.1 shows the confusion matrix produced for this benchmark model. In the benchmark model, wherein the features have been penalized by LASSO regularization, there are 8,934 instances of 'Class 1' that have been correctly predicted out of 16,048 instances. For 'Class 2', there are 6,347 instances correctly predicted as 'Class 2' out of 22,738 instances.

Table 5.1 Confusion matrix for Multinomial Logistic Regression with LASSO Regularization

|  |  | Actual/True | | | |
|---|---|---|---|---|---|
|  |  | Class 1 | Class 2 | Class 3 | Total |
| **Predicted** | Class 1 | 8934 | 7805 | 3362 | 20101 |
|  | Class 2 | 3461 | 6347 | 4992 | 14800 |
|  | Class 3 | 3653 | 8586 | 20495 | 32734 |
|  | Total | 16048 | 22738 | 28849 | 67635 |

For 'Class 3', there are 20,495 instances that are correctly predicted out of 28,849 instances. 3,461 instances of Class 1 are incorrectly predicted as Class 3 and 3,461 instances of Class 1 are incorrectly predicted as Class 2. As can be seen from Table, instances actually belonging to 'Class 2' are mostly incorrectly predicted as Class 3 (8,586 instances) and then as Class 1 (7,805 instances). If we see Class 3, the 4,992 instances that actually belong to Class 3 are incorrectly predicted as Class 2. The remaining 3,362 instances that actually belong to Class 3 are incorrectly predicted as Class 1.

## 5.2 Evaluation of Other Classifiers

The classifiers employed to perform multiclass classification are – Multinomial Logistic Regression, Random Forest, XGBoost, and One-Vs-Rest Classifier. Each of these has been trained and tested with the features selected by two feature selection techniques – Chi-Square Statistics and Information Gain. Thus, there are a total of 8 models built. Out of these 8, the two models that have used the One-Vs-Rest Classifier, are compared separately as these perform binary classifications for each target class using the one-vs-rest strategy. The other classifiers perform multiclass classification considering the target variable with all classes at once. Hence, the remaining six models have been compared together.

As can be seen in Table 5.2, the best accuracy score obtained while hyperparameter tuning is consistently low across all the models that were tried with certain parameters. The low accuracy is also seen while 10-fold cross-validation was conducted on the training dataset. The samples were even stratified in each fold during cross-validation to preserve the distribution of the target classes in each fold. The consistently low performance of all the models during training indicates that the models are struggling to identify patterns and learn from the training data. Training dataset even though it has a huge number of records, the presence of the large number of input features might be posing a challenge.

Table 5.2 Accuracy during the training of Multinomial Logistic Regression, Random Forest, and XGBoost models

| Classifier | Feature Selection | Accuracy | |
|---|---|---|---|
| | | **Hyperparameter Tuning Best Score** | **10-Fold Cross-validation (Mean)** |
| Multinomial Logistic Regression | Chi-Squared Statistics | 51.35% | 51.39% |
| | IG or Mutual Information | 51.24% | 51.23% |
| Random Forest | Chi-Squared Statistics | 50.29% | 50.26% |
| | IG or Mutual Information | 50.17% | 50.24% |
| XGBoost | Chi-Squared Statistics | 51.23% | 51.29% |
| | IG or Mutual Information | 51.10% | 51.22% |

Now, for the predictions done on the testing data, table 5.3 shows the comparison of performance metrics for the Multinomial Logistic Regression, Random Forest, and XGBoost models. The performance metrics for each target class along with the accuracy given by the model are shown in this table. As can be seen from Table 5.3, irrespective of the feature selection technique, the accuracy of random forest models is 0.52 (or 52%) and the accuracy of Multinomial Logistic Regression and XGBoost models are approximately 0.53 (or 53%). The accuracy gives the percentage of total samples for which the target classes were correctly predicted. In this study, the models have given very low prediction performance results. The multiclass classification results for both models are available in Appendix C: Section 5 – Additional Results.

Table 5.3 Performance evaluation metrics of Multinomial Logistic Regression, Random Forest, and XGBoost models on the testing dataset

| Classifier | Feature Selection | Target Class | Performance Evaluation Metrics | | | |
|---|---|---|---|---|---|---|
| | | | Accuracy | Precision | Recall | F1-Score |
| Multinomial Logistic Regression | Chi-Squared Statistics | Class 1 | **0.529** | 0.45 | **0.56** | 0.49 |
| | | Class 2 | | 0.43 | 0.29 | 0.35 |
| | | Class 3 | | 0.63 | 0.7 | 0.66 |
| | IG or Mutual Information | Class 1 | 0.5269 | 0.44 | 0.55 | 0.49 |
| | | Class 2 | | 0.43 | 0.29 | 0.34 |
| | | Class 3 | | 0.63 | 0.7 | 0.66 |
| Random Forest | Chi-Squared Statistics | Class 1 | 0.52 | 0.43 | 0.55 | 0.48 |
| | | Class 2 | | 0.43 | 0.23 | 0.3 |
| | | Class 3 | | 0.6 | 0.73 | 0.66 |
| | IG or Mutual Information | Class 1 | 0.5202 | 0.43 | 0.55 | 0.48 |
| | | Class 2 | | 0.43 | 0.23 | 0.3 |
| | | Class 3 | | 0.61 | 0.73 | 0.66 |
| XGBoost | Chi-Squared Statistics | Class 1 | **0.5303** | 0.45 | 0.54 | 0.49 |
| | | Class 2 | | 0.43 | **0.31** | 0.36 |
| | | Class 3 | | 0.63 | 0.7 | 0.66 |
| | IG or Mutual Information | Class 1 | 0.529 | 0.45 | 0.54 | 0.49 |
| | | Class 2 | | 0.43 | 0.31 | 0.36 |
| | | Class 3 | | 0.63 | 0.7 | 0.66 |

The recall of Class 3 is always moderate (0.7 or 0.73). The recall score of 0.7 indicates that the model can correctly identify 70% of the instances belonging to that class. Still higher recall is required as this problem is related to medical diagnosis. The recall is consistently poor for Class 2 across all the models irrespective of the feature selection technique used. The recall of Class 1 across all models is better than Class 2 but obviously, it is not satisfactory at all. Though the recall of Class 3 is highest for the random forest models, the recall of Class 2 is lowest for these random forest models.

Overall, the values of various performance metrics are consistently low for all the machine learning models mentioned. The low values are suggesting that none of the models is performing well and are struggling to predict the correct target classes for the instances in the testing dataset. The confusion matrix of the Multinomial Logistic Regression and XGBoost models with the Chi-Squared feature selection technique will be seen in detail to see how the target classes are getting predicted. The selected Multinomial Logistic Regression model has given the highest recall of 0.56 for Class 1 compared to other models. The selected XGBoost model has given the highest recall of 0.31 of Class 2 compared to other models. The recall of Class 3 is 0.7 for both models. Also, during execution, it has been noticed that the Chi-square feature selection is faster than the Information Gain (or Mutual Information) technique.

### 5.2.1   Multinomial Logistic Regression

The Multinomial Logistic Regression model constructed with features selected by the Chisquared statistics feature selection technique has given an accuracy of 0.529 (as seen in Table 5.3) after predicting the class outcomes for every sample present in the hold-out test dataset which has 67,635 records. Table 5.4 shows the confusion matrix produced for this model.

Table 5.4 Confusion matrix for Multinomial Logistic Regression model with Chi-Squared Statistics feature selection

|  |  | Predicted | | | |
|---|---|---|---|---|---|
|  |  | **Class 1** | **Class 2** | **Class 3** | **Total** |
| **Actual** | **Class 1** | 8909 | 3622 | 3517 | 16048 |
|  | **Class 2** | 7732 | 6644 | 8362 | 22738 |
|  | **Class 3** | 3323 | 5300 | 20226 | 28849 |
|  | **Total** | 19964 | 15566 | 32105 | **67635** |

As seen in Table 5.4, there are 8,909 instances of 'Class 1' that have been correctly predicted out of 16,048 instances actually belonging to Class 1. For 'Class 2', there are 6,644 instances correctly predicted as 'Class 2' out of 22,738 instances actually belonging to Class 2. For 'Class 3', there are 20,226 instances that are correctly predicted out of 28,849 instances actually belonging to Class 3. The 3,622 instances of Class 1 are incorrectly predicted as Class 2 and 3,517 instances of Class 1 are incorrectly predicted as Class 3. As can be seen from Table, instances actually belonging to 'Class 2' are mostly incorrectly predicted as Class 3 (8,362 instances) and then as Class 1 (7,732 instances). If we see Class 3, the 3,323 instances that actually belong to Class 3 are incorrectly predicted as Class 1. The remaining 5,300 instances that actually belong to Class 3 are incorrectly predicted as Class 2.

### 5.2.2   XGBoost

The XGBoost model constructed with features selected by the Chi-squared statistics feature selection technique, has given an accuracy of 0.5303 (as seen in Table 5.3) after predicting the class outcomes for every sample present in the hold-out test dataset which has 67,635 records.
Table 5.5 shows the confusion matrix produced for this model.

Table 5.5 Confusion matrix for XGBoost with Chi-Squared Statistics Feature Selection

|  |  | Predicted | | | |
|---|---|---|---|---|---|
|  |  | Class 1 | Class 2 | Class 3 | Total |
| Actual | Class 1 | 8688 | 3870 | 3490 | 16048 |
|  | Class 2 | 7309 | 7083 | 8346 | 22738 |
|  | Class 3 | 3151 | 5603 | 20095 | 28849 |
|  | Total | 19148 | 16556 | 31931 | **67635** |

As seen in Table 5.5, there are 8,688 instances of 'Class 1' that have been correctly predicted out of 16,048 instances actually belonging to Class 1. For 'Class 2', there are 7,083 instances correctly predicted as 'Class 2' out of 22,738 instances actually belonging to Class 2. For 'Class 3', there are 20,095 instances that are correctly predicted out of 28,849 instances actually belonging to Class 3. The 3,870 instances of Class 1 are incorrectly predicted as Class 2 and 3,490 instances of Class 1 are incorrectly predicted as Class 3.

As can be seen from Table 5.5, instances actually belonging to 'Class 2' are mostly incorrectly predicted as Class 3 (8,346 instances) and then as Class 1 (7,309 instances). If we see Class 3, the 3151 instances that actually belong to Class 3 are incorrectly predicted as Class 1. The remaining 5,603 instances that actually belong to Class 3 are incorrectly predicted as Class 2. The XGBoost model correctly predicted a greater number of 'Class 2' instances when compared to the Multinomial Logistic Regression model discussed in section 5.2.1. But the number of correct predictions of 'Class 1' and 'Class 3' by this XGBoost model were less when compared to the Multinomial Logistic Regression model. Thus, there is no significant improvement in the performance of the XGBoost model.

### 5.2.3 One-Vs-Rest Classifier

The two One-Vs-Rest Classifier models constructed, one with features selected by the Chisquared statistics and the other one by the information gain feature selection technique, have given the same AUC values for each class as shown in Table 5.6. The AUC shows the model's ability to discriminate between the positive and negative classes. Thus, the models have good discrimination ability for Class 1 and Class 3. They are better than chance as the AUC of chance level is 0.5. It is seen that the discrimination ability for Class 2 is low even in binary classification performed by the One-Vs-Rest Classifier. The ROC curves plotted for both models are available in Appendix C: Section 5 – Additional Results.

Table 5.6 AUC scores of One-Vs-Rest Classifier Model

| Classifier | Feature Selection | AUC | | | |
|---|---|---|---|---|---|
| | | Class 1 | Class 2 | Class 3 | microaveraged |
| One-Vs-Rest Classifier | Chi-Squared Statistics | 0.74 | 0.59 | 0.77 | 0.72 |
| | IG or Mutual Information | 0.74 | 0.59 | 0.77 | 0.72 |

The final predictions were done by finding the maximum probability among the three probabilities given by the binary classifications performed per class, for each sample of the testing dataset. For some samples, more than one binary classifier predicted the value as positive. Therefore, the classifier that had the highest probability was selected as the class for the samples of the testing dataset. Table 5.7 shows the performance evaluation metrics calculated for the final multiclass classification predictions made by each mentioned model. Thus, it can be seen in this table that the accuracy is again very low for both models.

As can be seen in Table 5.7, the recall of Class 1 is 0.59 for the first model and 0.58 for the second model. Also, the Chi-square feature selection is faster than Information Gain (or Mutual Information). Thus, the One-Vs-Rest Classifier model with the Chi-Square statistics feature selection has been considered further to analyse the outcomes of each binary classification and then the final multiclass classification. The binary as well as the multiclass classification results for both models are available in Appendix C: Section 5 – Additional Results.

Table 5.7 Performance evaluation metrics of One-Vs-Rest Classifier Model

| Classifier | Feature Selection | Target Class | Performance Evaluation Metrics | | | |
|---|---|---|---|---|---|---|
| | | | Accuracy | Precision | Recall | F1-Score |
| One-Vs-Rest Classifier | Chi-Squared Statistics | Class 1 | **0.5245** | 0.43 | **0.59** | 0.5 |
| | | Class 2 | | 0.43 | 0.23 | 0.3 |
| | | Class 3 | | 0.62 | 0.72 | 0.67 |
| | IG or Mutual Information | Class 1 | **0.5238** | 0.43 | **0.58** | 0.5 |
| | | Class 2 | | 0.43 | 0.24 | 0.3 |
| | | Class 3 | | 0.62 | 0.72 | 0.67 |

The one-vs-rest strategy transforms the problem of multiclass classification into multiple binary classifications. Thus, each binary classifier focuses on distinguishing one class from the rest which helps in increasing the prediction performance of that class. Table 5.8 shows the performance metrics for every binary classification performed by the One-Vs-Rest Classifier. As seen in Table, the accuracy of binary classification for Class 1 Vs rest is 0.7 followed by Class 3 vs rest which is 0.69. The accuracy for Class 2 vs rest is 0.56 which is very low. Despite using the one-vs-rest strategy, there is no significant improvement in the prediction of Class 2. The recall which is the True Positive Rate is only 0.58 for Class 2. This means that 58% of the actual Class 2 instances were correctly predicted as 'Class 2'.

Table 5.8 Performance metrics for every binary classification performed by the One-Vs-Rest Classifier (with Chi-Square feature selection)

| One-Vs-Rest Classifier | Class Label | Accuracy | Precision | Recall | F1-Score |
|---|---|---|---|---|---|
| Class 1 Vs. Rest | No | **0.7** | 0.86 | 0.72 | 0.79 |
|  | Yes |  | 0.41 | **0.64** | 0.5 |
| Class 2 Vs. Rest | No | 0.56 | 0.72 | 0.55 | 0.62 |
|  | Yes |  | 0.4 | **0.58** | 0.47 |
| Class 3 Vs. Rest | No | **0.69** | 0.77 | 0.65 | 0.71 |
|  | Yes |  | 0.61 | **0.74** | 0.67 |

Even though the accuracy of Class 1 Vs the rest is 0.7, the recall which is the True Positive Rate is only 0.64 for Class 1. This means that 64% of the actual Class 1 instances were correctly predicted as 'Class 1'. Though the accuracy of Class 3 Vs rest is 0.69, the recall which is the True Positive Rate is 0.74 for Class 3. This means that the 74% instances that belong to Class 3, were correctly predicted as Class 3. So, if the binary classifications are compared, Class 3 vs rest binary classification performance is better than the other two.

The probability score indicates how likely a sample belongs to a class. For every testing data sample, this score was computed by the One-Vs-Rest Classifier for each target class compared with the rest of the classes. Thus, each testing data sample had 3 probabilities. The first probability indicates how likely it belongs to Class 1, the second probability indicates how likely the sample belongs to Class 2, and the third probability indicates how likely the sample belongs to Class 3. If the first probability score is 0.9 then it means the model is 90% confident about the sample belonging to Class 1. If the first probability score is 0.3 then it means the model is only 30% confident that it belongs to Class 1. Thus, the more the score nears the value of 1, the more the confidence of the model about the target class.

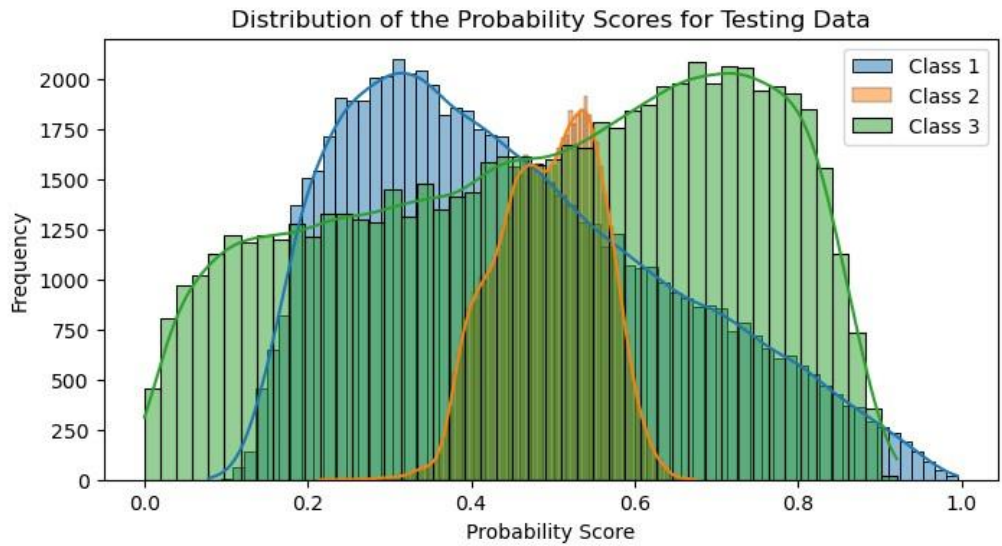

Figure 5.1 Distribution of the probability scores computed by the One-vs-rest Classifier for the testing data samples

Figure 5.1 shows the distribution of the probability scores for every sample of the testing data for each target class. This means the blue distribution is for Class 1 when compared with all other classes (Class 2 and Class 3) together. This means Class 1 is the positive class and the rest of the classes are considered as the negative class. The highest number of samples fall between the probability score of 0.2 and 0.4. Thus, most of the samples have low probability scores for Class 1. The distribution makes sense as the testing data has only 16,048 positive class (Class 1) instances and the remaining 51,587 instances belong to the negative class (rest of the classes).

The orange distribution is for Class 2 when compared with all other classes (Class 1 and Class 3) together. This means Class 2 is the positive class and the rest of the classes are considered as negative class. The highest number of samples fall between the probability score of 0.5 and 0.6. But the number of samples are also high between the probability score of 0.4 and 0.5. Thus, most of the samples have a probability score near 0.5 for Class 2. The model is uncertain and thus struggling to discriminate between Class 2 and the rest of the classes. The spread of the distribution is very narrow compared to that of Class 1 and Class 3.

The green distribution is for Class 3 when compared with all other classes (Class 1 and Class 2) together. This means Class 3 is the positive class and the rest of the classes are considered as negative class. The highest number of samples fall between the probability score of 0.6 and 0.8. Thus, a high frequency of samples have a high probability score for Class 3. The spread of the distribution is wide. The number of samples between probability score 0.0 to 0.5 for Class 3 is also high. This means a high number of samples have been estimated to not belong to Class 3, which means they belong to other classes. The distribution makes sense as the testing data has 28,849 positive class (Class 3) instances and the remaining 38,786 instances belong to the negative class (rest of the classes).

Table 5.9 Confusion matrix for One-Vs-Rest Classifier (final prediction) with Chi-Squared Statistics Feature Selection

|        |         | Predicted |         |         |       |
|--------|---------|-----------|---------|---------|-------|
|        |         | Class 1   | Class 2 | Class 3 | Total |
| Actual | Class 1 | 9394      | 2881    | 3773    | 16048 |
|        | Class 2 | 8511      | 5343    | 8884    | 22738 |
|        | Class 3 | 3826      | 4284    | 20739   | 28849 |
|        | Total   | 21731     | 12508   | 33396   | **67635** |

As seen in Table 5.9, there are 9,394 instances of 'Class 1' that have been correctly predicted out of 16,048 instances actually belonging to Class 1. For 'Class 2', there are 5,343 instances correctly predicted as 'Class 2' out of 22,738 instances actually belonging to Class 2. For 'Class 3', there are 20,739 instances that are correctly predicted out of 28,849 instances actually belonging to Class 3. The 2,881 instances of Class 1 are incorrectly predicted as Class 2 and 3,773 instances of Class 1 are incorrectly predicted as Class 3. As can be seen from Table, instances actually belonging to 'Class 2' are mostly incorrectly predicted as Class 3 (8,884 instances) and then as Class 1 (8,511 instances). If we see Class 3, the 3,826 instances that actually belong to Class 3 are incorrectly predicted as Class 1. The remaining 4,284 instances that actually belong to Class 3 are incorrectly predicted as Class 2.

The model seems to be struggling in assigning a higher probability to several samples to clearly distinguish the target class to which the sample may belong. This is evident from Figure 5.1 as the concentration of samples is high near 0.5 probability score, for all the three distributions plotted. This may be because the input data of such samples is complex and the model is unable to learn the pattern for clearly classifying these samples.

## 5.3 Cohen's Kappa Results

First, the models built using the features selected by the Chi-Squared feature selection have been additionally evaluated using Cohen's Kappa value. Table 5.10 shows the kappa values computed for the respective models. The values fall under the 0.21 - 0.40 range which indicates that there is a fair level of agreement between the actual class labels of the testing data and the predicted labels by the models. These values indicate that the predictions are slightly better than random chance. The highest kappa value of 0.28 is given by the Multinomial Logistics Regression model and the XGBoost model.

Table 5.10 Cohen's Kappa value for Models constructed with Chi-Squared Feature Selection

| Model | Kappa Score | Level of Agreement |
|---|---|---|
| Multinomial Logistic Regression | **0.28** | Fair Agreement |
| Random Forest | 0.26 | Fair Agreement |
| XGBoost | **0.28** | Fair Agreement |
| One-Vs-Rest Classifier | 0.27 | Fair Agreement |

Now, the kappa scores are computed by comparing predictions from two models. As the kappa value for the random forest model is the lowest (See Table 5.10), it is not considered for further analysis. Table 5.11 shows the level of agreement between the models. The kappa values computed between two different models fall in the range of 0.81 - 1.00 which indicates that the level of agreement is almost perfect (as mentioned in section 3.3.4 of section 3).

Table 5.11 Level of agreement between the selected models constructed with Chi-Squared Feature Selection

| Model 1 | Model 2 | Kappa Score | Level of Agreement |
|---|---|---|---|
| Multinomial Logistic Regression | XGBoost | 0.84 | Almost Perfect |
| Multinomial Logistic Regression | OneVsRest Classifier | **0.92** | Almost Perfect |
| XGBoost | OneVsRest Classifier | 0.82 | Almost Perfect |

The highest kappa value of 0.92 is given by the agreement between the predictions done by the Multinomial Logistic Regression and the One-Vs-Rest Classifier. The second highest kappa value of 0.84 is given by the agreement between the predictions done by the Multinomial Logistic Regression and the XGBoost model. The kappa value takes chance agreements into account. Though the models have given low accuracy and overall low performances as seen in the earlier section, these kappa values are indicating that the predictions of these models are consistent and not by chance. As there is a high level of agreement for the predictions among the models, there is a high possibility that the models are not predicting randomly. The models are giving inaccurate predictions for several instances but it may be due to complex patterns in the input data. There is a consistent agreement between the models even if some of the predictions are inaccurate.

## 5.4    Evaluation of Input Features

The input features play an important role in determining the sample may belong to which of the target classes. As seen in earlier sections, the performance of all the models developed is low for multiclass classification. Thus, additional analysis has been conducted to check the association of individual features with the target variable. These analyses have been performed to find the potential reasons why the models have struggled to identify patterns in the data that can help differentiate between the three target classes – 'Class 1', 'Class 2', and

'Class 3'. First, the 66 input features selected by LASSO and Chi-Square Statistics feature selection technique were seen and analysed. Then in general, all the 96 input features were analysed by the Chi-squared Statistics function using Python to get insights. Only the training dataset has been analysed as the models had learned to predict using training data only.

### 5.4.1 Feature Selection Techniques - LASSO vs. Chi-Squared Statistics

The Multinomial Logistic Regression with LASSO regularization is a benchmark model that performed multiclass classification using regression. The LASSO regularization shrinks the coefficients of irrelevant features to 0 and the rest of the features are used by the model for predicting the outcome. The chi-square statistics feature selection in Python, also selects the most relevant features that have an association with the target variable. In this study, the 66 independent input features selected by these feature selection techniques with the common features highlighted in green, have been mentioned in APPENDIX C: Section 5 – Additional Results. Table 5.12 shows the list of independent features that were different in LASSO and Chi-Square Statistics out of the 66 features.

Table 5.12 List of different independent features selected by the LASSO technique and ChiSquared Statistics

| Sr. No. | LASSO | Chi-Square Statistics |
|---|---|---|
| 1 | HTN | HYPOTHYROID |
| 2 | CABG | LIVER |
| 3 | BP_RX | PUD |
| 4 | PRIORITY_GROUP 2 | RHEUMATIC |
| 5 | MARRIED_SINGLE | BLOODLOSS |
| 6 | SYSTOLIC_RANGE_120-129 | SEVERE_DEP |
| 7 | SYSTOLIC_RANGE_140-179 | AMI |
| 8 | SYSTOLIC_RANGE_>=180 | HYPERG |
| 9 | BMI_RANGE_40-49.9 | ABI |
| 10 | BMI_RANGE_>=50 | RETSCREEN |
| 11 | HDL_RANGE_Missing | PRIORITY_GROUP 1 |
| 12 | TRI_RANGE_>=200 | SERUMALB_RANGE_Missing |

So, out of the 12 features mentioned in Table 5.12 for the chi-square test, 10 features are comorbidities (Sr. No. 1 to 10). It can be seen that the chi-square test has selected the features related to comorbidities more than the features related to clinical biomarkers or demographics, as seen in the LASSO technique. In LASSO, only HTN and CABG are comorbidities. BP_RX feature is related to prescription. There are two features (PRIORITY_GROUP 2 and MARRIED_SINGLE) related to demographics. There are seven features related to clinical biomarkers (Sr. No. 6 to 12) selected by LASSO.

Despite there being some features that differed between the LASSO and Chi-Squared Statistical feature selection techniques, the performance of the various models employing these features was still seen to be low. Feature selection techniques identify the relationship between the features and the target variable. However, in this study regarding multiclass classification, it seems that the features do not have strong discriminatory power to correctly predict the target class. Therefore, even when different sets of features are selected and used by various machine learning algorithms of varying complexity, the prediction results are still low. A simple algorithm like logistic regression to a complex algorithm like XGBoost is facing difficulty in accurately discriminating between target classes and hence giving similar results.

### 5.4.2 Features Analysis using Chi-Squared Statistics

Features analysis using the chi2() feature selection Python function was conducted to see the association between the 96 independent input features and the dependent target variable 'Mortality', present in the training data. This was an additional analysis done to see the importance of every feature and hence all the 96 input variables of the training dataset were considered. Figure 5.2 shows the target variable is dependent on which features. Here, the target variable means all the classes are considered.

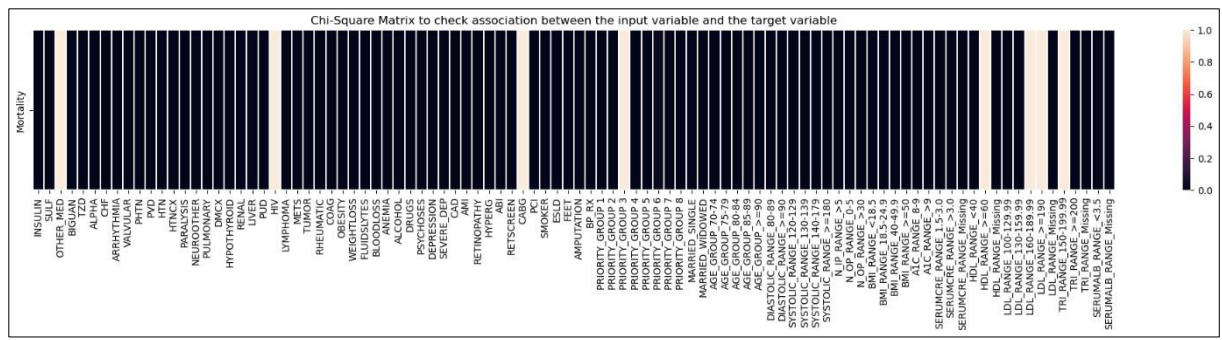

Figure 5.2 Chi-Square Test results matrix depicting the association between the input variables and the target variable 'Mortality'

As seen in Figure 5.2, the independent features for which the p-value is not less than 0.05 are shown to have a value of 1 to easily differentiate them on the heatmap displayed. Thus, the features which are in cream colour or denoted as 'non-black' are the features that are not associated with the target variable 'Mortality'. So, out of the 96 independent input variables or features, only 8 features have no association with the target variable. The rest 88 features denoted as black on the plotted heatmap, are associated with the target variable as their pvalue is less than 0.05. Out of these 88, the top 66 that were selected during the Chi-squared feature selection performed before building models, are listed in APPENDIX C: Section 5 – Additional Results.

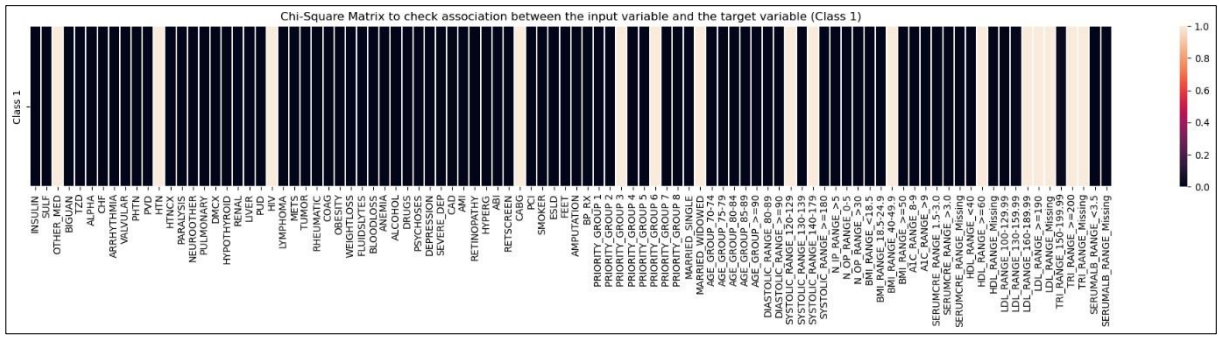

Figure 5.3 Chi-Square Test results matrix depicting the association between the input variables and the target 'Class 1'

Now, every independent feature may not be relevant in predicting all the target classes. To check if this is possible, the association between each of the independent input features and each class of the target variable 'Mortality', was also checked individually. The process to compute the p-values has been discussed in section 4.9 of section 4. Figure 5.3, shows the heatmap that depicts the association between the independent features and the target 'Class 1'. As can be seen in Figure 5.3, there are 16 features denoted as non-black, that are not associated with the target variable 'Class 1'. So, out of the 96 independent input variables or features, 16 features have no association with the target 'Class 1'. The rest 80 features denoted as black on the plotted heatmap, are associated with the target 'Class 1'.

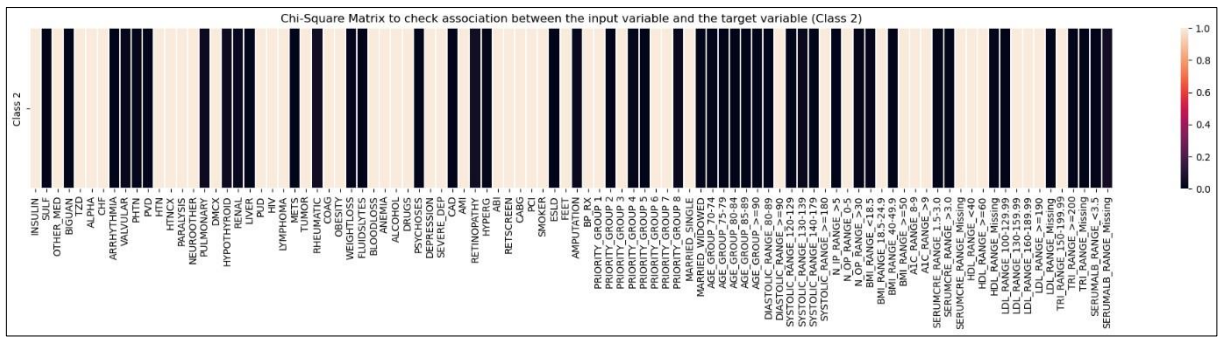

Figure 5.4 Chi-Square Test results matrix depicting the association between the input variables and the target 'Class 2'

Similarly, figure 5.4 shows the heatmap that depicts the association between the independent features and the target 'Class 2'. There are 49 features denoted as non-black, that are not associated with the target variable 'Class 2'. So, out of the 96 independent input variables or features, 49 features have no association with the target 'Class 2'. The rest 47 features denoted as black on the plotted heatmap, are associated with the target 'Class 2'. Now, figure 5.5 shows the heatmap that depicts the association between the independent features and the target 'Class 3'. As can be seen in this figure, there are 8 features denoted as non-black, that are not associated with the target variable 'Class 3'. So, out of the 96 independent input variables or features, only 8 features have no association with the target 'Class 3'. The rest 88 features denoted as black on the plotted heatmap, are associated with the target 'Class 3'.

Figure 5.5 Chi-Square Test results matrix depicting the association between the input variables and the target 'Class 3'

All the features (except the 'PRIORITY_GROUP 2' feature) that do not have any association with the target 'Class 3', do not have association with the overall target variable 'Mortality' as well. The difference is that the 'PRIORITY_GROUP 2' is not associated with target 'Class 3' but it is associated with the 'Mortality' target variable. This feature is considered as associated with the target 'Mortality' because of its association with 'Class 1' and 'Class 2'. The 'PRIORITY_GROUP 3' feature even though it is associated with 'Class 3', is considered as not associated with target 'Mortality' because it's not associated with 'Class 1' and 'Class 2'. Similarly, the 'TRI_RANGE_150-199.99' feature even though it is associated with 'Class 1', is considered as not associated with target 'Mortality' because it's not associated with 'Class 2' and 'Class 3'.

As seen from these figures, almost all the features are relevant to more than one class of the target variable. Thus, this observation further proves the lack of discriminatory power of the features to differentiate between the three classes present in the 'Mortality' variable in this study. Due to the presence of such relevant features, the models might find it difficult to distinctly differentiate between the classes during training. It then becomes challenging for the model to correctly predict the class which leads to low performances due to incorrect predictions. Most of the features seem to have an association with the target 'Class 3' then followed by 'Class 1'. The least number of features have an association with target 'Class 2'. This seems to be the reason why the recall of 'Class 2' is always the lowest.

Also, the chi-squared score can give insights into how high the association is between the independent input feature and the target variable. Thus, figure 5.6 shows the chi-squared values to depict the chi-squared values between each of the 96 input variables and the target variable and also with the individual classes. The higher the chi-squared score, the higher is the association between the variables that are compared. The lower the chi-squared score, the weaker is the association between the variables that are compared. For 'Class 2' it can be seen in Figure 5.6 that the chi-squared values are low even if there is an association between the input variable and the target 'Class 2' (See Figure 5.4). For example, the feature 'SULF' is associated with all the classes (See Figures 5.3, 5.4 and 5.5). But if we see the chi-squared score of this 'SULF' feature in Figure 5.6, it is very low for 'Class 2' when compared to 'Class 1' and 'Class 3'. The highest association of the feature 'SULF' is with 'Class 3'.

We had earlier seen that out of 96 features, 80 features are associated with the target 'Class 1'. Now, if we check individually, out of these 80 features, the chi-squared value is higher for

'Class 1' compared to that of 'Class 3' for only 37 features. Thus, the remaining 43 features have a higher association with 'Class 3' compared to 'Class 1'. Please note that the comparison of scores has been done using Microsoft Excel and the output is not available in this thesis. There also seems to be a closeness between the chi-squared scores for 'Class 1' and 'Class 3' for many features. This means that the association level between the input variable and these two classes is similar. Thus, in such cases, the input variable may not strongly be able to differentiate between these classes.

After conducting this feature analysis, it seems that dummy encoding the variables is ineffective for this multiclass classification problem. The dummy encoding of categorical features was even performed in the previous study conducted by Griffith et al., 2020 on this mortality dataset but it was not a multiclass classification problem. But in this study, it seems that a large number of dummy features have confused the model more than helped it correctly predict the target class. Thus, the dummy encoding does not seem to be an effective approach for a multiclass classification problem using this mortality dataset.

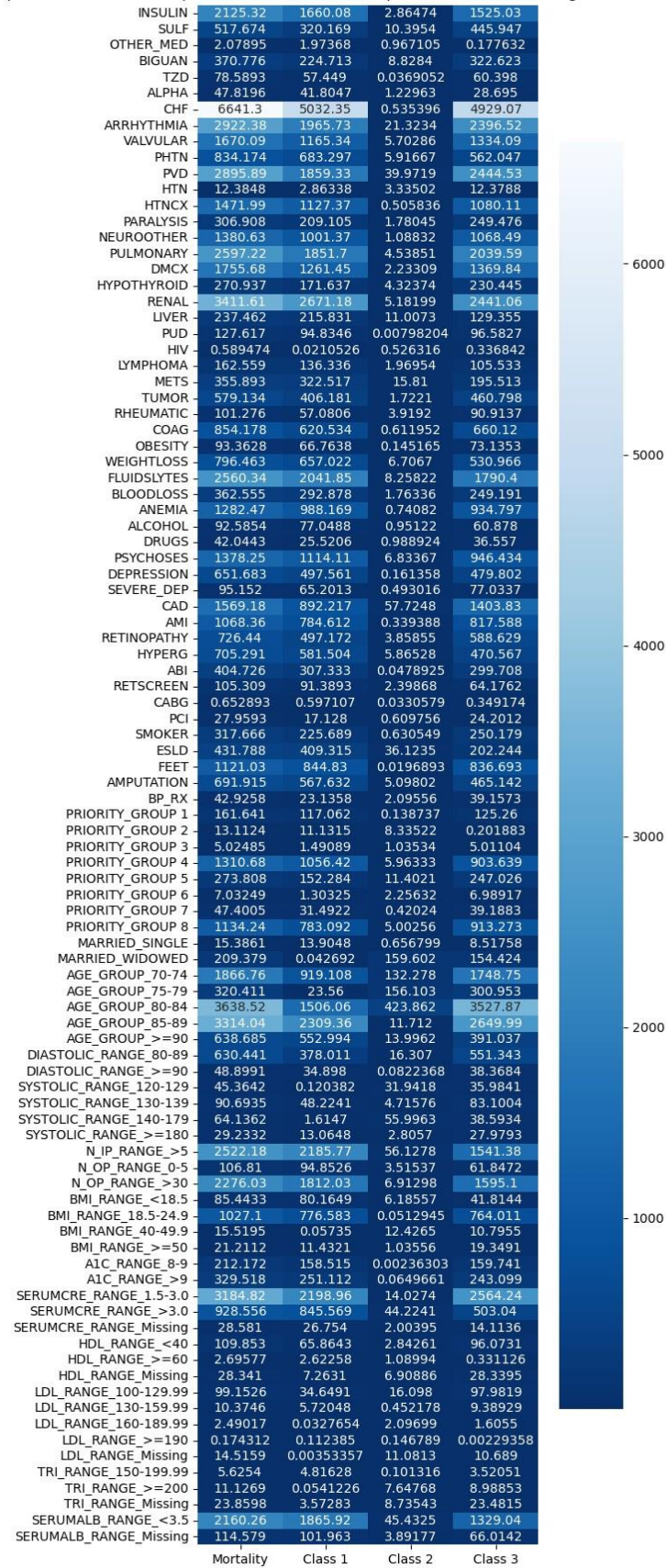

Figure 5.6 Chi-Squared values matrix depicting the level of association between the input variables and the target

This section includes the results and analysis of the entire study that has been conducted to perform multiclass classification on the mortality dataset. The benchmark model constructed using Multinomial Logistic Regression with LASSO has given an accuracy of 0.529. Then the performance of the four classifiers – Multinomial Logistic Regression, Random Forest, XGBoost, and One-Vs-Rest Classifier, have been summarized. Each classifier has been constructed using the filter-based selection techniques – Chi-squared and Information Gain. Thus, other than the benchmark model, there are a total of 8 models. It has been observed that all these multiclass classification models have given very low accuracy (52% - 53%) on the testing dataset. For 'Class 3', the recall or true positive rate is the highest (70% - 72%) and for the target Class 2, it is the lowest, by all the models.

The probabilities have also been estimated by the One-Vs-Rest classifier for each target class by considering one target class as positive and the rest of the classes as negative class. These probabilities when plotted using a histogram along with a density curve, show that the concentration of samples is high near 0.5 probability score, for all three distributions. The model seems to be struggling with some samples in assigning a higher probability to clearly distinguish the target class to which the sample may belong. Also, the kappa values for pairs of models indicate that the predictions of these models even if inaccurate are consistent and not by chance.

Even when some different features have been selected by the LASSO and the Chi-squared statistical feature selection technique, the overall performance of these models is not good. It seems that the features do not have strong discriminatory power to correctly predict the target class. Further analysis has been done by using Chi-Squared Statistics with the training dataset to see which input features have an association with each of the target classes. It has been observed that almost all the features are relevant to more than one class of the target variable.

Most of the features seem to have an association with the target 'Class 3' then followed by 'Class 1'. The least number of features have an association with target 'Class 2'. Even the chisquared values are low for 'Class 2' even if there is an association between the input variable and the target 'Class 2'. In general, 'Class 3' dominates 'Class 1' when the level of association is compared. The closeness between the chi-squared values for 'Class 1' and 'Class 3' for many input features indicates that such input variables may face difficulty in strongly differentiating between these classes.

## 6. CONCLUSION

In this section, the overall study has been discussed briefly. Based on the results achieved, conclusions have been drawn and elaborated in this section.

### 6.1 Summary

Individuals having Type 2 Diabetes Mellitus (T2DM) can reduce the mortality risk by following proper treatments. Health practitioners can form proper treatment plans when factors such as the individual's comorbidities and remaining life, are considered. But predicting the remaining life is a challenge! Machine Learning (ML) algorithms have been used in past studies to develop prediction models that can predict the occurrence of all-cause mortality in individuals with T2DM. Most of these studies have employed the Cox Proportional Hazards Regression models for mortality prediction. The literature review on allcause mortality prediction concluded that the impact of clinical trial data and real-world data, the inclusion of comorbidities and prescriptions, the transportability of models to different age groups, genetics and cultural backgrounds, and consideration of external validation; are still needed to be explored more and preferably with large-scale data.

As the human body ages, its immune system weakens. Thus, older adults who are aged 65 years and older are more susceptible to health-related problems. The risk of complications and comorbidities increases if they have diabetes. They are more susceptible to hypoglycaemia or premature death. Thus, they need a proper diabetes treatment plan so that the diabetes is manageable and their health doesn't get severely affected. This can prevent hospitalization or at least reduce hospital readmissions. A prediction model is required for the older population aged 65 years or older with Type 2 Diabetes to identify high-risk individuals and prevent premature mortality.

The previous research conducted on the dataset used in this study had built two separate mortality risk prediction models using the Logistic LASSO Regressions. These prediction models were built for older patients with T2DM. The 5-year model gave C-statistics of 0.74 and the 10-year model gave C-statistics of 0.76. The significant part of the study was the dataset. The same mortality dataset has been considered in this current study because it is large-scale data with 275,190 patient records and a total of 68 potential predictors of mortality. The data has features related to demographic, clinical biomarkers, medical prescriptions and a lot of comorbidities as well. In the current study, the proposal has been to create a multiclass classification model that would predict whether the patients' remaining life is "up to 5 years", "more than 5 but up to 10 years", or "more than 10 years".

The number of features is high in the dataset. Therefore, some recent studies related to feature selection techniques used for medical datasets were discussed which concluded that various filter-based feature selection techniques have been used by researchers. These are deemed better than wrapper-based or embedded ones. Thus, the Chi-Squared Statistics and the Information Gain filter-based techniques were employed. Then, some recent research works regarding the use of ML algorithms for multiclass classification in healthcare industry problems were also seen. ML classifiers such as the Multinomial Logistic Regression, Random Forest, XGBoost, and One-vs-Rest Classifier were selected for this current study. A benchmark model using Multinomial Logistic Regression with LASSO has also been built for this multiclass classification problem.

## 6.2 Summary of Findings

The benchmark model constructed using Multinomial Logistic Regression with LASSO has given an accuracy of 0.529. Then the four classifiers – Multinomial Logistic Regression, Random Forest, XGBoost, and One-Vs-Rest Classifier, have been constructed using the filterbased selection techniques – Chi-squared Statistics and Information Gain. Thus, other than the benchmark model, there are a total of 8 models. It has been observed that all these multiclass classification models have given very low accuracy (52% - 53%) on the testing dataset. For 'Class 3', the recall or true positive rate is the highest (70% - 72%) and for the target 'Class 2' it is the lowest, by all the models. Cohen's kappa values (0.82, 0.84, and 0.92) indicate that there is perfect agreement between the predictions of the models and not by chance. Even if the models are not predicting by chance, the low accuracy indicates that maybe they are unable to capture the complexities of the multiclass classification problem.

Though all the models have shown acceptable performance for predicting Class 3 (remaining life is more than 10 years), the prediction performance for Class 2 (remaining life is more than 5 but up to 10 years) is the worst. The prediction performance of Class 1 (remaining life is up to 5 years) is better than Class 2 but still significantly low. Results significantly lower than the acceptable value were also given by the benchmark model which is a regression model built using the Multinomial Logistic Regression with LASSO. Thus, the similar Logistic LASSO Regression employed in the previous research on this dataset also produced low results when used as multinomial regression for the current multiclass classification problem.

Contrary to expectations, this study has not produced any well-performing ML model. To understand why all the multiclass classification models have been constantly giving a low performance, additional analyses have been conducted at the feature level using the training dataset. Checking the association of input features with each target class provided some significant insights. Most of the input variables are associated with more than one target class. For most features, the association is highest with Class 3 followed by Class 1. The least number of features are associated with Class 2. These findings indicate each model's inability to effectively capture every underlying complex pattern of the data.

The dummy encoding of categorical variables was considered during the data pre-processing steps to see the impact of each unique category present in these variables on the target variable. This was even performed in the previous study conducted on this mortality dataset but it was not a multiclass classification problem. Thus, the high dimensionality of the input data after dummy encoding the categorical input features seems to have confused the models more than helped them correctly predict the target class in this multiclass classification problem. The process of dummy encoding input variables though not incorrect but has been ineffective for this multiclass classification problem.

### 6.1.2 Contribution of this Study

The extension of the original problem to multiclass has never been tried before and hence there are no previous research results of multiclass classification for this mortality dataset. The entire research approach used in this study to perform multiclass classification has been tried for the first time on this dataset. Here, the multiclass classification has been performed in three different ways – by regression, then using multiclass classifiers, and finally by an ensemble of multiple binary classifications (One-Vs-Rest strategy). The constantly low overall prediction results given by all the models indicate that at least the approach taken in this study is not helping the models to perform multiclass classification. Though the methodology proposed in this study has been ineffective in achieving the aim, the extension to multiclass has still opened up new frontiers. It has provided a foundation in the unexplored areas for future investigations.

### 6.2 Future Recommendations

This study has provided some valuable insights into mortality prediction using multiclass classification for older adults having Type 2 Diabetes Mellitus. There is a need to acknowledge the low results and the limitations. As challenges were encountered with the current data pre-processing steps, there is a need to modify and try other techniques. Missing values can be handled by imputation instead of keeping it as a separate category. Outliers can be treated using alternate methods instead of handling them by discretization. Alternative methods can be explored after thorough research of past studies. Maybe modifying the data pre-processing steps to improve the results of multiclass classification can be a good starting point rather than directly employing other advanced machine learning algorithms. Further, hyperparameter tuning can be conducted in a more detailed manner if computational resources are available.

## REFERENCES

Abdellatif, A., Abdellatef, H., Kanesan, J., Chow, C.O., Chuah, J.H. and Gheni, H.M., (2022) An Effective Heart Disease Detection and Severity Level Classification Model Using Machine Learning and Hyperparameter Optimization Methods. *IEEE Access* [online], V.10, pp.79974–79985. Available at: https://ieeexplore.ieee.org/document/9831786. [Accessed: 27th March 2023]

Agaal, A. and Essgaer, M., (2022) Influence of Feature Selection Methods on Breast Cancer Early Prediction Phase using Classification and Regression Tree. In: *Proceedings - 2022 International Conference on Engineering and MIS, ICEMIS 2022*. [online] IEEE. 04-06 July. Available at: https://ieeexplore.ieee.org/document/9914078. [Accessed: 26th March 2023]

Ahire, N., Awale, R.N., Patnaik, S. and Wagh, A., (2022) A comprehensive review of machine learning approaches for dyslexia diagnosis. *Multimedia Tools and Applications (ACM)*, V.82(9), pp.13557–13577.

Ali, K., Shaikh, Z.A., Khan, A.A. and Laghari, A.A., (2022) Multiclass skin cancer classification using EfficientNets – a first step towards preventing skin cancer. *Neuroscience Informatics* [online], V.2(4). Available at: https://doi.org/10.1016/j.neuri.2021.100034. [Accessed: 5th February 2023]

Alyas, T., Hamid, M., Alissa, K., Faiz, T., Tabassum, N. and Ahmad, A., (2022) Empirical Method for Thyroid Disease Classification Using a Machine Learning Approach. *BioMed Research*

*International* [online], V.2022. Available at: https://doi.org/10.1155/2022/9809932. [Accessed: 1st June 2023]

de Andrade, E.C., Pinheiro, L.I.C.C., Pinheiro, P.R., Nunes, L.C., Pinheiro, M.C.D., Pereira, M.L.D., de Abreu, W.C., Filho, R.H., Simão Filho, M., Pinheiro, P.G.C.D. and Nunes, R.E.C., (2023) Hybrid model for early identification post-Covid-19 sequelae. *Journal of Ambient Intelligence and Humanized Computing* [online], Available at: https://doi.org/10.1007/s12652-023-04555-3. [Accessed: 1st April 2023]

Bárcenas, R. and Fuentes-García, R., (2022) Risk assessment in COVID-19 patients: A multiclass classification approach. *Informatics in Medicine Unlocked* [online], V.32(2022). Available at: https://doi.org/10.1016/j.imu.2022.101023. [Accessed: 2nd February 2023]

Barsasella, D., Gupta, S., Malwade, S., Aminin, Susanti, Y., Tirmadi, B., Mutamakin, A., Jonnagaddala, J. and Syed-Abdul, S., (2021) Predicting length of stay and mortality among hospitalized patients with type 2 diabetes mellitus and hypertension. *International Journal of Medical Informatics* [online], V.154. Available at: https://doi.org/10.1016/j.ijmedinf.2021.104569. [Accessed: 2nd February 2023]

Bashir, S., Khattak, I.U., Khan, A., Khan, F.H., Gani, A. and Shiraz, M., (2022) A Novel Feature Selection Method for Classification of Medical Data Using Filters, Wrappers, and Embedded Approaches. *Complexity* [online], V.2022. Available at: https://doi.org/10.1155/2022/8190814. [Accessed: 26th March 2023]

Basu, S., Sussman, J.B., Berkowitz, S.A., Hayward, R.A., Bertoni, A.G., Correa, A., Mwasongwe, S. and Yudkin, J.S., (2018) Validation of risk equations for complications of type 2 diabetes (RECODe) using individual participant data from diverse longitudinal cohorts in the U.S. *Diabetes Care*, V.41(3), pp.586–595.

Basu, S., Sussman, J.B., Berkowitz, S.A., Hayward, R.A. and Yudkin, J.S., (2017) Development and validation of Risk Equations for Complications Of type 2 Diabetes (RECODe) using individual participant data from randomised trials. *The Lancet Diabetes and Endocrinology*, V.5(10), pp.788–798.

Biswas, S., Kumar, V. and Das, S., (2021) Multiclass classification models for Personalized Medicine prediction based on patients Genetic Variants. In: *2021 IEEE International Conference on Technology, Research, and Innovation for Betterment of Society, TRIBES 2021*. [online] IEEE. 17-19 December, Available at: https://ieeexplore.ieee.org/document/9751631. [Accessed: 26th March 2023]

Cascaro, R.J., Gerardo, B.D. and Medina, R.P., (2019) Filter selection methods for multiclass classification. In: *ACM International Conference Proceeding Series*. [online] Association for Computing Machinery, 18 October pp.27–31. Available at: https://doi.org/10.1145/3366650.3366655.

Centers for Disease Control and Prevention, (2022) *Learn how people with type 2 diabetes can live longer | Diabetes | CDC*. [online] Available at: https://www.cdc.gov/diabetes/resourcespublications/research-summaries/reaching-treatment-goals.html [Accessed 10 Feb. 2023].

Chang, Y.K., Huang, L.F., Shin, S.J., Lin, K.D., Chong, K., Yen, F.S., Chang, H.Y., Chuang, S.Y., Hsieh, T.J., Hsiung, C.A. and Hsu, C.C., (2017) A Point-based Mortality Prediction System for Older Adults with Diabetes. *Scientific Reports*, V.7(1), pp.1–10.

Chen, T., Madanian, S., Airehrour, D. and Cherrington, M., (2022a) Machine learning methods for hospital readmission prediction: systematic analysis of literature. *Journal of Reliable Intelligent Environments*, V.8(1), pp.49–66.

Chen, Y., Su, J., Qin, Y., Luo, P., Shen, C., Pan, E., Lu, Y., Miao, D., Zhang, N., Zhou, J., Yu, X. and Wu, M., (2022b) Fresh fruit consumption, physical activity, and five-year risk of mortality among patients with type 2 diabetes: A prospective follow-up study. *Nutrition, Metabolism and Cardiovascular Diseases*, V.32(4), pp.878–888.

Chiu, S.Y.H., Chen, Y.I., Lu, J.F.R., Ng, S.C. and Chen, C.H., (2021) Developing a prediction model for 7-year and 10-year all-cause mortality risk in type 2 diabetes using a hospital-based prospective cohort study. *Journal of Clinical Medicine*, V.10(20), pp.1–14.

Chong, J., Tjurin, P., Niemelä, M., Jämsä, T. and Farrahi, V., (2021) Machine-learning models for activity class prediction: A comparative study of feature selection and classification algorithms. *Gait and Posture*, V.89(6), pp.45–53.

Clarke, P.M., Gray, A.M., Briggs, A., Farmer, A.J., Fenn, P., Stevens, R.J., Matthews, D.R., Stratton, I.M. and Holman, R.R., (2004) A model to estimate the lifetime health outcomes of patients with Type 2 diabetes: The United Kingdom Prospective Diabetes Study (UKPDS) Outcomes Model (UKPDS no. 68). *Diabetologia*, V.47(10), pp.1747–1759.

Copetti, M., Biancalana, E., Fontana, A., Parolini, F., Garofolo, M., Lamacchia, O., De Cosmo, S., Trischitta, V. and Solini, A., (2021) All-cause mortality prediction models in type 2 diabetes: applicability in the early stage of disease. *Acta Diabetologica*, V.58(10), pp.1425–1428.

Copetti, M., Shah, H., Fontana, A., Scarale, M.G., Menzaghi, C., De Cosmo, S., Garofolo, M., Sorrentino, M.R., Lamacchia, O., Penno, G., Doria, A. and Trischitta, V., (2019) Estimation of Mortality Risk in Type 2 Diabetic Patients (ENFORCE): An Inexpensive and Parsimonious Prediction Model. *Journal of Clinical Endocrinology and Metabolism*, V.104(10), pp.4900–4908.

De Cosmo, S., Copetti, M., Lamacchia, O., Fontana, A., Massa, M., Morini, E., Pacilli, A., Fariello, S., Palena, A., Rauseo, A., Viti, R., Di Paola, R., Menzaghi, C., Cignarelli, M., Pellegrini, F. and Trischitta, V., (2013) Development and validation of a predicting model of all-cause mortality in patients with type 2 diabetes. *Diabetes Care*, V.36(9), pp.2830–2835.

Demirkıran, F., Çayır, A., Ünal, U. and Dağ, H., (2022) An ensemble of pre-trained transformer models for imbalanced multiclass malware classification. *Computers and Security*, V.121(October 2022), pp.1–38.

Dewangan, S., Rao, R.S. and Mishra, A., (2022) Code Smell Detection Using Ensemble Machine Learning Algorithms. *Applied Sciences*, [online] V.12(20). Available at: https://doi.org/10.3390/app122010321. [Accessed: 26th March 2023]

Ebiaredoh-Mienye, S.A., Swart, T.G., Esenogho, E. and Mienye, I.D., (2022) A Machine Learning Method with Filter-Based Feature Selection for Improved Prediction of Chronic Kidney Disease. *Bioengineering* [online], V.9(8). Available at: https://doi.org/10.3390/bioengineering9080350. [Accessed: 26th March 2023]

El-Sappagh, S., Ali, F., Abuhmed, T., Singh, J. and Alonso, J.M., (2022a) Automatic detection of Alzheimer's disease progression: An efficient information fusion approach with heterogeneous ensemble classifiers. *Neurocomputing*, V.512(November 2022), pp.203–224.

El-Sappagh, S., Saleh, H., Ali, F., Amer, E. and Abuhmed, T., (2022b) Two-stage deep learning model for Alzheimer's disease detection and prediction of the mild cognitive impairment time. *Neural Computing and Applications*, V.34(17), pp.14487–14509.

Emam, W., Tashkandy, Y., Hamedani, G.G., Shehab, M.A., Ibrahim, M. and Yousof, H.M., (2023) A Novel Discrete Generator with Modeling Engineering, Agricultural and Medical Count and ZeroInflated Real Data with Bayesian, and Non-Bayesian Inference. *Mathematics* [online], V.11(5). Available at: https://doi.org/10.3390/math11051125. [Accessed: 1st June 2023]

Galar, M., Fernández, A., Barrenechea, E., Bustince, H. and Herrera, F., (2011) An overview of ensemble methods for binary classifiers in multi-class problems: Experimental study on one-vs-one and one-vs-all schemes. *Pattern Recognition*, V.44(8), pp.1761–1776.

González, S., García, S., Del Ser, J., Rokach, L. and Herrera, F., (2020) A practical tutorial on bagging and boosting based ensembles for machine learning: Algorithms, software tools, performance study, practical perspectives and opportunities. *Information Fusion*, V.64(December 2020), pp.205–237.

Griffith, K., (2021) *A Novel Dataset of Predictors of Mortality for Older Veterans Living with Type II Diabetes - Mendeley Data*. [online] Available at: https://data.mendeley.com/datasets/kn8v3678n9/3 [Accessed 10 Feb. 2023].

Griffith, K.N., Prentice, J.C., Mohr, D.C. and Conlin, P.R., (2020) Predicting 5-and 10-year mortality risk in older adults with diabetes. *Diabetes Care*, V.43(8), pp.1724–1731.

Gujral, U.P., Pradeepa, R., Weber, M.B., Narayan, K.M.V. and Mohan, V., (2013) Type 2 diabetes in South Asians: Similarities and differences with white Caucasian and other populations. *Annals of the New York Academy of Sciences*, V.1281(1), pp.51–63.

Guleria, K., Sharma, S., Kumar, S. and Tiwari, S., (2022) Early prediction of hypothyroidism and multiclass classification using predictive machine learning and deep learning. *Measurement: Sensors* [online], V.24. Available at: https://doi.org/10.1016/j.measen.2022.100482. [Accessed: 26th March 2023]

Hayes, A.J., Leal, J., Gray, A.M., Holman, R.R. and Clarke, P.M., (2013) UKPDS Outcomes Model 2: A new version of a model to simulate lifetime health outcomes of patients with type 2 diabetes mellitus using data from the 30 year united kingdom prospective diabetes Study: UKPDS 82. *Diabetologia*, V.56(9), pp.1925–1933.

Hess, A.S. and Hess, J.R., (2017) Understanding tests of the association of categorical variables: the Pearson chi-square test and Fisher's exact test. *Transfusion*, V.57(4), pp.877–879.

Jayasri, N.P. and Aruna, R., (2022) Big data analytics in health care by data mining and classification techniques. *ICT Express*, V.8(2), pp.250–257.

Joo, B., Ahn, S.S., An, C., Han, K., Choi, D., Kim, H., Park, J.E., Kim, H.S. and Lee, S.K., (2023) Fully automated radiomics-based machine learning models for multiclass classification of single brain tumors: Glioblastoma, lymphoma, and metastasis. *Journal of Neuroradiology*, V.50(4), pp.388–395.

Kim, H.S., Lee, S. and Kim, J.H., (2018) Real-world evidence versus randomized controlled trial: Clinical research based on electronic medical records. *Journal of Korean Medical Science*, V.33(34), pp.1–7.

Lee, S., Zhou, J., Leung, K.S.K., Wu, W.K.K., Wong, W.T., Liu, T., Wong, I.C.K., Jeevaratnam, K., Zhang, Q. and Tse, G., (2021) Development of a predictive risk model for all-cause mortality in patients with diabetes in Hong Kong. *BMJ Open Diabetes Research and Care*, V.9(1), pp.1–12.

Liu, C.S., Li, C.I., Wang, M.C., Yang, S.Y., Li, T.C. and Lin, C.C., (2021) Building clinical risk score systems for predicting the all-cause and expanded cardiovascular-specific mortality of patients with type 2 diabetes. *Diabetes, Obesity and Metabolism*, V.23(2), pp.467–479.

Ma, R.C.W. and Chan, J.C.N., (2013) Type 2 diabetes in East Asians: Similarities and differences with populations in Europe and the United States. *Annals of the New York Academy of Sciences*, V.1281(1), pp.64–91.

Mahboob Alam, T., Iqbal, M.A., Ali, Y., Wahab, A., Ijaz, S., Imtiaz Baig, T., Hussain, A., Malik, M.A., Raza, M.M., Ibrar, S. and Abbas, Z., (2019) A model for early prediction of diabetes. *Informatics in Medicine Unlocked* [online], V.16. Available at: https://doi.org/10.1016/j.imu.2019.100204. [Accessed: 5th February 2023]

Malik, M.B., Ganie, S.M. and Arif, T., (2022) *Machine learning techniques in healthcare informatics: Showcasing prediction of type 2 diabetes mellitus disease using lifestyle data*. [online] *Predictive Modeling in Biomedical Data Mining and Analysis*. Elsevier Inc. Available at: https://doi.org/10.1016/B978-0-323-99864-2.00001-9. [Accessed: 5th February 2023]

Martins, M. V., Baptista, L., Machado, J. and Realinho, V., (2023) Multi-Class Phased Prediction of Academic Performance and Dropout in Higher Education. *Applied Sciences (Switzerland)* [online], V.13(8). Available at: https://doi.org/10.3390/app13084702. [Accessed: 3rd June 2023]

Mchugh, M.L., (2013) The Chi-square test of independence Lessons in biostatistics. *Biochemia Medica*, V.23(2), pp.143–149.

McHugh, M.L., (2012) Lessons in biostatistics interrater reliability : the kappa statistic. *Biochemica Medica*, V.22(3), pp.276–282.

Meng, J. and Xing, R., (2022) Inside the "Black-Box": Embedding Clinical Knowledge in Data-driven Machine Learning for Heart Disease Diagnosis. *Cardiovascular Digital Health Journal*, V.3(6), pp.276–288.

Morillo-Velepucha, D., Réategui, R., Valdiviezo-Díaz, P. and Barba-Guamán, L., (2022) Congestive heart failure prediction based on feature selection and machine learning algorithms. In: *Iberian Conference on Information Systems and Technologies, CISTI*. [online] IEEE, 22-25 June pp.22–25. Available at: https://ieeexplore.ieee.org/document/9820312. [Accessed: 1st June 2023]

Mukherjee, I., Sahu, N.K. and Sahana, S.K., (2022) Simulation and Modeling for Anomaly Detection in IoT Network Using Machine Learning. *International Journal of Wireless Information Networks*, V.30(2), pp.173–189.

Nibbering, D. and Hastie, T.J., (2022) Multiclass-penalized logistic regression. *Computational Statistics and Data Analysis* [online], V.169. Available at: https://doi.org/10.1016/j.csda.2021.107414. [Accessed: 6th February 2023]

Novitski, P., Cohen, C.M., Karasik, A., Hodik, G. and Moskovitch, R., (2022) Temporal patterns selection for All-Cause Mortality prediction in T2D with ANNs. *Journal of Biomedical Informatics* [online], V.134. Available at: https://doi.org/10.1016/j.jbi.2022.104198. [Accessed: 2nd February 2023]

Pagano, E., Konings, S.R.A., Di Cuonzo, D., Rosato, R., Bruno, G., van der Heijden, A.A., Beulens, J., Slieker, R., Leal, J. and Feenstra, T.L., (2021) Prediction of mortality and major cardiovascular complications in type 2 diabetes: External validation of UK Prospective Diabetes Study outcomes model version 2 in two European observational cohorts. *Diabetes, Obesity and Metabolism*, V.23(5), pp.1084–1091.

Painuli, D., Bhardwaj, S. and Köse, U., (2022) Recent advancement in cancer diagnosis using machine learning and deep learning techniques: A comprehensive review. *Computers in Biology and Medicine* [online], V.146. Available at: https://doi.org/10.1016/j.compbiomed.2022.105580. [Accessed: 4th April 2023]

Pankajavalli, P.B. and Karthick, G.S., (2022) An Independent Constructive Multi-class Classification Algorithm for Predicting the Risk Level of Stress Using Multi-modal Data. *Arabian Journal for Science and Engineering*, V.47(8), pp.10547–10562.

Parida, U., Nayak, M. and Nayak, A.K., (2021) News text categorization using random forest and naïve bayes. In: *1st Odisha International Conference on Electrical Power Engineering, Communication and Computing Technology, ODICON 2021*. [online] IEEE, 08-09 January pp.2021–2024. Available at: https://ieeexplore.ieee.org/document/9428925. [Accessed: 1st June 2023]

Puneet and Chauhan, A., (2020) Detection of Lung Cancer using Machine Learning Techniques Based on Routine Blood Indices. In: *2020 IEEE International Conference for Innovation in Technology, INOCON 2020*. [online] IEEE, 06-08 November pp.1–6. Available at: https://ieeexplore.ieee.org/document/9298407. [Accessed: 1st June 2023]

Qi, J., He, P., Yao, H., Xue, Y., Sun, W., Lu, P., Qi, X., Zhang, Z., Jing, R., Cui, B. and Ning, G., (2022) Developing a prediction model for all-cause mortality risk among patients with type 2 diabetes mellitus in Shanghai, China. *Journal of Diabetes*, V.2022, pp.1–9.

Ramos-Pérez, I., Arnaiz-González, Á., Rodríguez, J.J. and García-Osorio, C., (2022) When is resampling beneficial for feature selection with imbalanced wide data? *Expert Systems with Applications* [online], V.188. Available at: https://doi.org/10.1016/j.eswa.2021.116015. [Accessed: 9th February 2023]

Reddy, K.V.V., Elamvazuthi, I., Aziz, A.A., Paramasivam, S., Chua, H.N. and Pranavanand, S., (2021) Heart disease risk prediction using machine learning classifiers with attribute evaluators. *Applied Sciences (Switzerland)*, V.11(18).

Robinson, T.E., Elley, C.R., Kenealy, T. and Drury, P.L., (2015) Development and validation of a predictive risk model for all-cause mortality in type 2 diabetes. *Diabetes Research and Clinical Practice*, V.108(3), pp.482–488.

Shao, H., Fonseca, V., Stoecker, C., Liu, S. and Shi, L., (2018) Novel Risk Engine for Diabetes Progression and Mortality in USA: Building, Relating, Assessing, and Validating Outcomes (BRAVO). *PharmacoEconomics*, V.36(9), pp.1125–1134.

Sharma, S. and Mandal, P.K., (2023) A Comprehensive Report on Machine Learning-based Early Detection of Alzheimer's Disease using Multi-modal Neuroimaging Data. *ACM Computing Surveys*, V.55(2), pp.1–44.

Shobana, V. and Nandhini, K., (2022) Ensemble Techniques to improve the performance of the High Dimensional MultiClass Algorithms. In: *2022 1st International Conference on Electrical, Electronics, Information and Communication Technologies, ICEEICT 2022*. [online] IEEE. 16-18 February. Available at: https://ieeexplore.ieee.org/document/9768646. [Accessed: 6th February 2023]

Shorewala, V., (2021) Early detection of coronary heart disease using ensemble techniques. *Informatics in Medicine Unlocked*, V.26.

Shukla, A., Katal, A., Raghuvanshi, S. and Sharma, S., (2021) Criminal Combat: Crime Analysis and Prediction Using Machine Learning. In: *2021 International Conference on Intelligent Technologies, CONIT 2021*. [online] IEEE, 25-27 June pp.1–5. Available at: https://ieeexplore.ieee.org/document/9498397. [Accessed: 1st June 2023]

Sipper, M., (2022) Binary and multinomial classification through evolutionary symbolic regression. In: *Proceedings of the Genetic and Evolutionary Computation Conference Companion*. 19 July pp.300–303. Available at: https://dl.acm.org/doi/10.1145/3520304.3528922. [Accessed: 6th February 2023]

van Smeden, M., Reitsma, J.B., Riley, R.D., Collins, G.S. and Moons, K.G., (2021) Clinical prediction models: diagnosis versus prognosis. *Journal of Clinical Epidemiology*, V.132, pp.142–145.

Swain, S., Bhushan, B., Dhiman, G. and Viriyasitavat, W., (2022) Appositeness of Optimized and Reliable Machine Learning for Healthcare: A Survey. *Archives of Computational Methods in Engineering*, V.29(6), pp.3981–4003.

The World Health Organization, (2022) *Diabetes*. [online] Available at: https://www.who.int/newsroom/fact-sheets/detail/diabetes [Accessed 10 Feb. 2023].

Theis, J., Galanter, W.L., Boyd, A.D. and Darabi, H., (2022) Improving the In-Hospital Mortality Prediction of Diabetes ICU Patients Using a Process Mining/Deep Learning Architecture. *IEEE Journal of Biomedical and Health Informatics*, V.26(1), pp.388–399.

Tislenko, M.D., Gaidel, A. V. and Kupriyanov, A. V., (2022) Comparison of feature selection algorithms for Data classification problems. In: *2022 8th International Conference on Information Technology and Nanotechnology, ITNT 2022*. [online] IEEE, pp.1–5. Available at: https://ieeexplore.ieee.org/document/9848765. [Accessed: 1st June 2023]

Vaidya, A.U., Benavidez, G.A., Prentice, J.C., Mohr, D.C., Conlin, P.R. and Griffith, K.N., (2022) A novel dataset of predictors of mortality for older Veterans living with type II diabetes. *Data in Brief*, V.41(April 2022).

Vanacore, A., Pellegrino, M.S. and Ciardiello, A., (2023) Evaluating classifier predictive performance in multi-class problems with balanced and imbalanced data sets. *Quality and Reliability Engineering International*, V.39(2), pp.651–669.

Wan, E.Y.F., Fong, D.Y.T., Fung, C.S.C., Yu, E.Y.T., Chin, W.Y., Chan, A.K.C. and Lam, C.L.K., (2017) Prediction of five-year all-cause mortality in Chinese patients with type 2 diabetes mellitus – A population-based retrospective cohort study. *Journal of Diabetes and its Complications*, V.31(6), pp.939–944.

Wang, Z., Sun, X., Wang, B., Shi, S. and Chen, X., (2023) Lasso-Logistic regression model for the identification of serum biomarkers of neurotoxicity induced by strychnos alkaloids. *Toxicology Mechanisms and Methods*, V.33(1), pp.65–72.

Wells, B.J., Jain, A., Arrigain, S., Yu, C., Rosenkrans, W.A. and Rattan, M.W., (2008) Predicting 6year mortality risk in patients with type 2 diabetes. *Diabetes Care*, V.31(12), pp.2301–2306.

Yang, X., So, W.Y., Tong, P.C.Y., Ma, R.C.W., Kong, A.P.S., Lam, C.W.K., Ho, C.S., Cockram, C.S., Ko, G.T.C., Chow, C.C., Wong, V.C.W. and Chan, J.C.N., (2008) Development and validation of an all-cause mortality risk score in type 2 diabetes: The Hong Kong diabetes registry. *Archives of Internal Medicine*, V.168(5), pp.451–457.

Yashodhar, A. and Kini, S., (2021) Performance Measurements of different Classification techniques for the Alzheimer's Disease Neuroimaging Initiative. In: *2021 IEEE International Conference on Distributed Computing, VLSI, Electrical Circuits and Robotics, DISCOVER 2021 - Proceedings*. [online] IEEE, pp.322–326. Available at: https://ieeexplore.ieee.org/document/9663705. [Accessed: 1st June 2023]

Zhang, B., Li, Y. and Chai, Z., (2022) A novel random multi-subspace based ReliefF for feature selection. *Knowledge-Based Systems*, V.252.

Zhuo, X., Melzer Cohen, C., Chen, J., Chodick, G., Alsumali, A. and Cook, J., (2022) Validating the UK prospective diabetes study outcome model 2 using data of 94,946 Israeli patients with type 2 diabetes. *Journal of Diabetes and its Complications*, V.36(1).

# APPENDIX A: Section 2 - Summary of related works

| Citation | Main work | Dataset | Feature Selection Technique | Data Mining Technique | Performance Measure | Advantage | Weakness |
|---|---|---|---|---|---|---|---|
| (Yang et al., 2008) | 5-year mortality risk score prediction model developed using data of Chinese patients | Hong Kong Diabetes Registry cohort (n=7583; median age = 57 years; IQR = 47-67 years) | Cox proportional regression analysis with a backward algorithm | Cox proportional hazards regression to form a model; Hosmer and Lemeshow test to check calibration | Below time-specific measures are calculated using Chambless method, AUROC = 0.85 Sensitivity = 75.4% Specificity = 75.7% | 1. Observed and predicted death rates in testing dataset were similar with probability greater than 0.70  2. Good calibration and discrimination of model indicates mortality can be predicted using common clinical and chemical variables. | 1. To test the validity of the predicted score, a second validation with an external independent sample is needed.  2. Validation required on other populations as well. |
| (Wells et al., 2008) | 6-Year mortality risk prediction tool developed | EHR at Cleveland Clinic (n=33067; age >=18 years) | Deliberately not used | Cox proportional hazards regression | C-statistic = 0.752 | 1. Large sample size cohort used for the study.  2. Strong internal validation of model due to the use of crossvalidation.  3. Online risk calculator available for clinical use.  4. Interaction of medication classes with age, CHF and GFR. | 1. External validation with other population not done. 2. Misclassification bias potentially present for the oral medication in the Cleveland clinic data. 3. Less number of comorbid conditions considered as predictors. |

| Citation | Main work | Dataset | Feature Selection Technique | Data Mining Technique | Performance Measure | Advantage | Weakness |
|---|---|---|---|---|---|---|---|
| (De Cosmo et al., 2013) | Development of a 2-year parsimonious model | For training, real-life data GMS (n=679; mean age 62.1 years old) For validation, real-life data FMS (n=936; mean age 63.6 years old) | cNRI driven forward variable selection | Multivariate Cox proportional hazards regression | For GMS, C-statistic = 0.88. For FMS, C-statistic = 0.82. For pooled sample, Cstatistic = 0.84. | 1. Development of web-based risk engine which can be used for testing other samples. 2. Use of a different predictor variables selection method. | 1. It's a short-term mortality prediction model. 2. Study used small samples and of Italian ancestry only. 3. Comorbidities have not been considered. |
| (Hayes et al., 2013) | Development and validation of a new model UKPDS-OM2 to be the substitute for UKPDS-OM1 | From 20 years trial, UKPDS (n=5102; aged 25-65 years) From survivors entering 10-year post-trial, UKPDS (n=4031; aged 25-65 years) | Stepwise regression | Logistic Regression (in the year of a complication) Gompertz proportional hazards survival models (for years with no complications) | Remaining Life Expectancy for, patient group 50 - 54 years = 25.1 years patient group 60 - 64 years = 17.7 years patient group 70 - 74 years = 11.7 years | 1. Greater precision and number of significant covariates are more than UKPDS-OM1. 2. Longer follow-up data of 30 years used. | 1. External validation on realworld data not done. 2. Hyper- and Hypoglycaemia complications not considered. 3. The dataset is a bit inconsistent as during the extra 10-year follow-up the patients were not part of clinical trial. It was observational real-world data. |

| Citation | Main work | Dataset | Feature Selection Technique | Data Mining Technique | Performance Measure | Advantage | Weakness |
|---|---|---|---|---|---|---|---|
| (Robinson et al., 2015) | 5-year mortality risk score prediction model developed using data of patients in New Zealand | For development, NZDCS (n=26864; mean age 62 years old)  For validation, DCSS (n=7610; mean age 59 years old) | No technique - variables included as required | Cox proportional hazards regression | Best model which included markers of renal disease,  For development cohort, had a C-statistic of 0.80  For validation cohort, had a C-statistic of 0.79 | 1. Cohort consisted of multiethnic patients' data. 2. Effect of different marker types studied as three progressive models were developed with best one being the third with renal disease markers. | Apart from CVD, no other comorbidities are considered. |
| (Basu et al., 2017) | Development and Validation of RECODe | For development, ACCORD (n=9635; aged 40–79 years)  For validation, DPPOS (n=1018; 6.0 years median followup) Look AHEAD (n=4760; 10.6 years median follow-up) | Elastic net regularisation | Cox proportional hazards regression | Internal validation on C-statistic = 0.70 Calibration slope = 1.03 Calibration intercept = −0.002  External validation on DPPOS, C-statistic = 0.71 Calibration slope = 1.10 Calibration intercept = −0.012 | 1. Risk calculated for a longer time period of 10 years. 2. Online risk calculator produced. | 1. RECODe equations may not have good generalizability as the development and validation of equations are performed on clinical trials data. 2. C-statistic value for allcause mortality equation is lower than other all-cause mortality equations tested on other populations. |

| Citation | Main work | Dataset | Feature Selection Technique | Data Mining Technique | Performance Measure | Advantage | Weakness |
|---|---|---|---|---|---|---|---|
| (Chang et al., 2017) | Development of a point-based, elderly-specific 5-year mortality prediction model using Taiwanese patients' data | Taiwan's AHSP data, For development (n= 220832; aged >=65 years) For internal validation (n= 24538; aged >=65 years) For external validation, MJ Health Screening data (n=2093; aged >=65 years) | All variables included in the model | Multivariate Cox proportional hazards 10-fold crossvalidation is also used. Also, bootstrapping technique used to evaluate model overfitting. | Harrell's C in the development, internal-, and external-validation datasets were 0.737, 0.746, and 0.685 respectively. | 1. The average optimism of the Harrell's C statistics was 0.0032 for all 200 bootstrap models indicating no overfitting. 2. Easy-to-use model specific for older adults has been established which predicts 5-year mortality risk. | 1. Model's C-statistic has a significant decrease when validated on an external dataset. 2. The comorbidities and medication details of patients were not considered as predictors in this study. |
| (Wan et al., 2017) | Gender-specific models for prediction of 5year all-cause mortality in Chinese patients | n=132462; Chinese aged between 18 and 79 years randomly split by 2:1 into development and validation cohorts | Forward stepwise Cox regression using F-test to calculate p-values | Cox proportional hazards regression | For males, Harrell's Cstatistic = 0.768  D statistic = 1.586 $R^2$ = 37.5 For females, Harrell's C-statistic = 0.782 D statistic = 1.737 $R^2$ = 41.9 | 1. Separate gender-specific models for male and female population. 2. More accurate than the existing New Zealand and JADE models. 3. Study done on significantly large number of patient records. | 1. Models are developed and validated on only Chinese patients. 2. Models are only applicable to T2DM patients having no comorbidities. |

| Citation | Main work | Dataset | Feature Selection Technique | Data Mining Technique | Performance Measure | Advantage | Weakness |
|---|---|---|---|---|---|---|---|
| (Basu et al., 2018) | External validation of RECODe (Basu et al., 2017) among diverse population | For validation, MESA (n=1555; averaged 63.0 years old) JHS (n=1746; averaged 57.5 years old) | Refer (Basu et al., 2017) | Refer (Basu et al., 2017) | For MESA data - C-statistic = 0.81 Calibration Slope = 1.03 For JHS data - C-statistic = 0.78 Calibration Slope = 1.01 | 1. Use of US modern real-world individual patient data instead of UK clinical trials data 2. C-statistics is improved compared to that in (Basu et al., 2017) | External validation done on small dataset. |
| (Shao et al., 2018) | Development of a novel risk engine BRAVO for diabetes progression and mortality in USA | For development and internal validation, ACCORD clinical trial (n=10251; For external validation, ASPEN, ADVANCE, and CARDS clinical trials | Backward selection method | Gompertz proportional hazard 10-fold crossvalidation is also used. | c-statistic = 0.617 Brier score = 0.010 In external validation, slope = 1.071 $R^2$ = 0.86 | 1. The risk engine can be used to simulate the progression of diabetes and mortality for over 40 years. 2. This risk engine found that slightly above 7% of glycosylated hemoglobin level is associated with lowest risk for all-cause mortality. | External validation of BRAVO is also done on clinical trial data and thus needs to be done on realworld data of USA as well as other populations. |

| Citation | Main work | Dataset | Feature Selection Technique | Data Mining Technique | Performance Measure | Advantage | Weakness |
|---|---|---|---|---|---|---|---|
| (Copetti et al., 2019) | Development and validation of ENFORCE - a 6-year mortality risk prediction Model | For development, RWD - GMS (n=1019; Mean age = 61.1 years) External validation on, RWD - FMS (n=1045; Mean age = 63.6 years) RWD - PMS (n=972; Mean age = 59.6 years) CTD - ACCORD (n=3150; Mean age = 63.3 years) | Continuous net reclassification improvement | Multivariate Cox proportional hazards regression | C-statistic: 0.79 (GMS), 0.78 (FMS) and 0.75 (PMS). Pooling of three cohorts with real-life samples gave C-statistics: 0.80. For ACCORD, Cstatistic: 0.68. | 1. Trained on real-life sample data. 2. Confirmed that model performance lowers on clinical trial data which was initially seen in RECODe as well. | 1. Training and the other two validation samples used are restricted only to the Italy population. 2. Datasets have a significantly low number of records. |
| (Griffith et al., 2020) | Development of 5- and 10-Year mortality risk prediction models for older adults with T2DM | VHA CDW (n = 275,190; aged ≥ 65 years) | LASSO Regularization | Logistic LASSO Regression 10-fold crossvalidation is also used. | For the 5-year model, C-statistics of 0.74 Calibration slopes 1.01 For the 10-year model, C-statistics of 0.76 Calibration slopes 1.01 | 1. Use of a robust dataset that contains information routinely collected in the EHR. 2. Both the elderly-specific models showed good discrimination. | 1. Logistic LASSO regression methodology allows for bias in the resulting odds ratios for better predictive accuracy. 2. Dataset is specific to US veterans and mostly male. |

| Citation | Main work | Dataset | Feature Selection Technique | Data Mining Technique | Performance Measure | Advantage | Weakness |
|---|---|---|---|---|---|---|---|
| (Copetti et al., 2021) | Applicability of all-cause mortality prediction models in the early stage of T2DM | Training set - Pisa Early stage (n=302; mean age 63.5 years) Testing set - Pisa Mortality Study (n=392; mean age 57.5 years) | Refer (Basu et al., 2017; Copetti et al., 2019) | Refer (Basu et al., 2017; Copetti et al., 2019) | On training set, ENFORCE - C-statistic = 0.81 RECODe - C-statistic = 0.80 On testing set, C-statistic for ENFORCES: 0.78 and for RECODe: 0.78 | 1. ENFORCE and RECODe prediction models successfully applied in the early stage of T2DM. 2. ENFORCE and RECODe both performed similarly good in the two different samples. 3. These models can help address the reduction of life expectancy effectively. | The ENFORCE and RECODe models have been trained and validated on small sample sets having data of only Italian patients. |
| (Chiu et al., 2021) | Development of 7-year and 10year all-cause mortality risk prediction model using data of Taiwanese subjects | Hospital-based data from CGMH-K (n=18202; age >= 18 years) 50% considered as training set and remaining 50% as validation set. | Multivariable cox proportional hazards model using stepwise approach | Multivariate Cox Regression Akaike information criterion (AIC) and LASSO for model selection | For 7-year model, Harrell's C-statistic: 0.7955 Integrated AUC = 0.8136 For 10-year model, Harrell's C-statistic: 0.7775 Integrated AUC = 0.8045 | 1. Patients with missing values were also considered by adopting missing-indicator method. 2. Mortality data is taken from the nationwide death registry to identify all patients that died outside the hospital. | 1. Hospital patients' data is specific to a city in Taiwan thus the model prediction may differ among cities and countries. 2. Behavioural factors such as alcohol consumption, exercise, and cigarette smoking, are not present in the data. |

| Citation | Main work | Dataset | Feature Selection Technique | Data Mining Technique | Performance Measure | Advantage | Weakness |
|---|---|---|---|---|---|---|---|
| (Liu et al., 2021) | Risk scoring system created to estimate the short and longterm risks of allcause mortalities | DCMP (n=9692; mean age = 58 years) spit into 2:1 ratio as below, Training dataset (n=6461) Validation dataset (n=3231) | Not mentioned | Multivariate Cox proportional hazards regression Hosmer–Lemeshow $x^2$ test for goodness of fit | AUROC = 0.79 (3-year), 0.78 (5-year), 0.80 (10-year) and 0.80 (15-year) Harrell's c-statistics = 0.77 Calibration Slope = 1.00 | 1. Risk prediction for 3-, 5-, 10-, and 15-year all-cause mortality. 2. Internal validation performed by 1000-time bootstrap resampling to avoid overfitting. | 1. External validation with other population patients not done. 2. Co-morbidities and medicines use information taken through questionnaire thus may have incorrect data. |
| (Pagano et al., 2021) | External validation of UKPDS-OM2 using two European population cohorts | Italian CMS (n=1931; mean age = 67.8 years) Dutch DCS (n=5188; mean age = 64.8 years) | Refer (Hayes et al., 2013) | Refer (Hayes et al., 2013) | CMS - For 5-, 10-, 15-year allcause mortality, C-statistic = above 0.70 Bias (5-year) = 0.04 Bias (10-year) = 0.05 Bias (15-year) = 0.06 DCS - For 5- and 10-year allcause mortality, C-statistic = above 0.70 Bias (5-year) = 0.09 Bias (10-year) = 0.14 | 1. Use of real-world cohort data having follow-up of 10 and 15 years. 2. Results indicate that new or updated risk equations might be needed for other populations as transferability of existing model is unsatisfactory. | 1. Some microvascular endpoints (like renal failure, amputation) and cardiovascular death were excluded from analysis. 2. At follow-up, risk factor data were completely missing in both the cohorts. |

| Citation | Main work | Dataset | Feature Selection Technique | Data Mining Technique | Performance Measure | Advantage | Weakness |
|---|---|---|---|---|---|---|---|
| (Lee et al., 2021) | Development of a predictive risk model using machine/deep learning approach for Hong Kong diabetic patients | CDARS (n=273678; mean age = 65.4 years) | Multivariate Cox Regression | Multivariate Cox proportional hazards regression; Cox Regression (Cox Score); Random Survival Forests (RSF) and Deep neural survival learning (DeepSurv); Use of 5-fold cross-validation. | RSF was the best model with Precision = 0.88, Recall = 0.87, AUC = 0.86, and C-statistic = 0.87; Cox Score Model had a C-statistic of 0.73 that improved to 0.87 using RSF and 0.86 using DeepSurv. | 1. Long follow-up period of over 10 years. 2. Use of additional predictors such as laboratory test results, drug details, and various comorbidities apart from demographic details. 3. Latest machine learning techniques applied. | 1. Presence of potential information bias as a retrospective study data is used. 2. Interpretability of the deep neural network-based model is weak. |

| Citation | Main work | Dataset | Feature Selection Technique | Data Mining Technique | Performance Measure | Advantage | Weakness |
|---|---|---|---|---|---|---|---|
| (Chen et al., 2022b) | Associations among fresh fruit consumption, physical activity, and their dose-response relationship with 5-year all-cause mortality risk | CRPCD (n=20340; aged 21-94 years) | Not mentioned | Cox regression | Fruit consumption was inversely associated with all-cause mortality (HR 0.76; 95% CI 0.64-0.88) The HRs comparing the top vs bottom physical activity quartiles were 0.44 (0.37-0.53) for all-cause mortality. | 1. Study concluded that fruit consumption and physical activity may reduce all-cause mortality. 2. Use of restricted cubic spline regression for the dose-response analysis to examine the nonlinear relationships. | 1. Information on diets and physical activity collected through questionnaires. 2. Longer follow-up periods should be considered to have risk models for 10-, 15-, and more years as the age spectrum is large. |
| (Zhuo et al., 2022) | External validation of UKPDS-OM2 using data of Israeli patients | MHS (n=94946; aged >= 26 years) | Refer (Hayes et al., 2013) | Refer (Hayes et al., 2013) | C-statistic = 0.86 P/O = 1.32 | This study highlights that recalibration is important when applying OM2 to patients with different risk of developing diabetes complications. | 1. The model overpredicted all-cause mortality risk. 2. Results are specific to Israeli population and may not be generalizable. |
| (Qi et al., 2022) | Development of a predictive 5year all-cause mortality risk model for diabetic patients in Shanghai, | SLHD (n=399784; median age= 63 years; IQR 56-72 years) spit into 2:1 ratio as below, Training dataset (n=266523) Validation dataset (n=133261) | Univariate and multivariate Cox proportional hazards | Cox proportional hazards | Discrimination (Harrell C-index) = 0.8113 (mean) | 1. This is the first 5-year mortality risk prediction model with good predictive performance for south China patients. | 1. Needs external validation with other populations. 2. Longer follow-up periods should be |

| Citation | Main work | Dataset | Feature Selection Technique | Data Mining Technique | Performance Measure | Advantage | Weakness |
|---|---|---|---|---|---|---|---|
| | China | | | | | 2. Risk model was internally validated using 100 bootstrap samples. | considered to have risk models for 10-, 15-, and more years as the age spectrum is large. |

# APPENDIX B: Section 4 - Additional Results

1.  Box plots

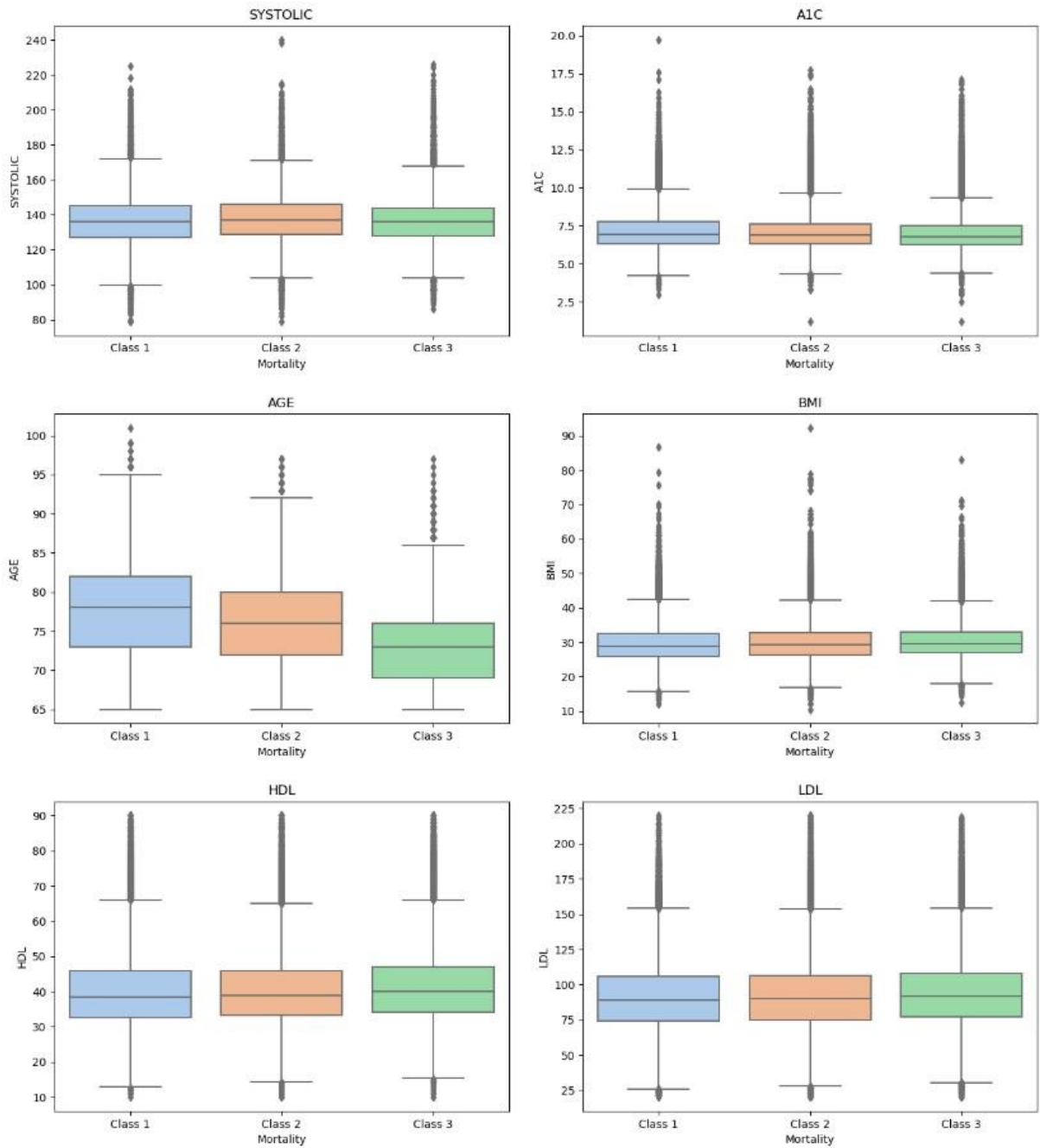

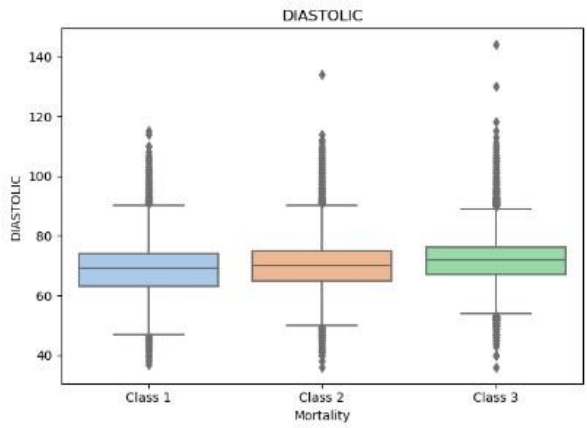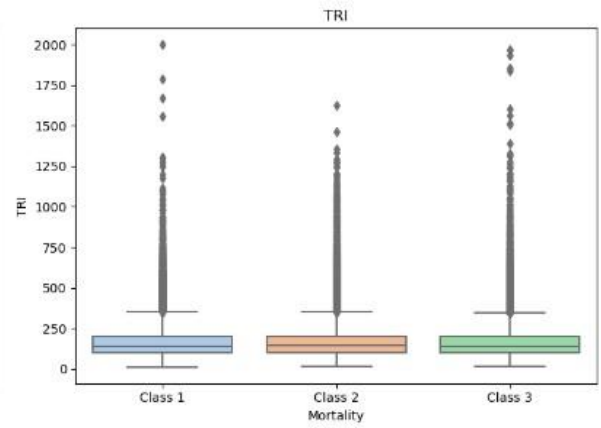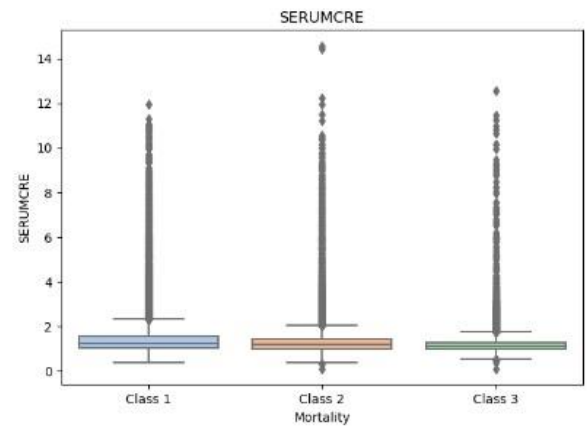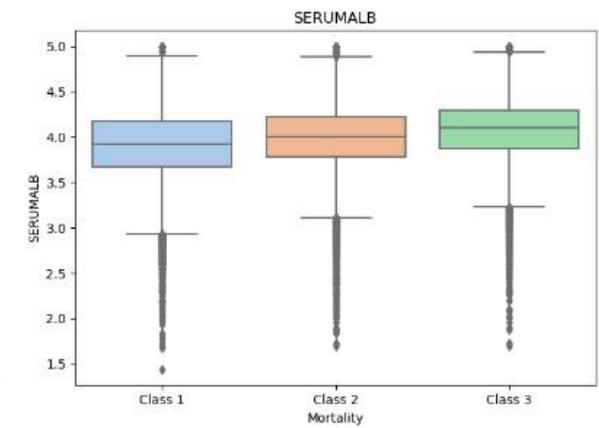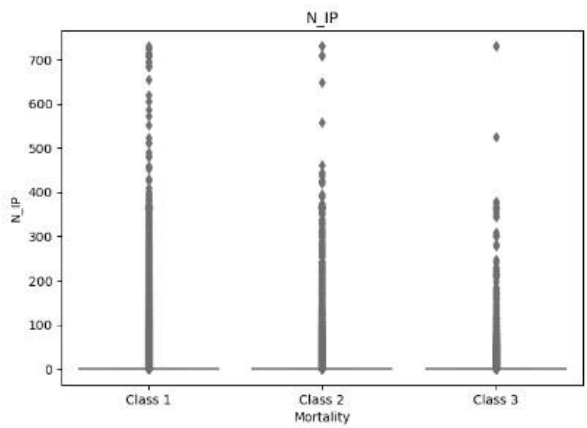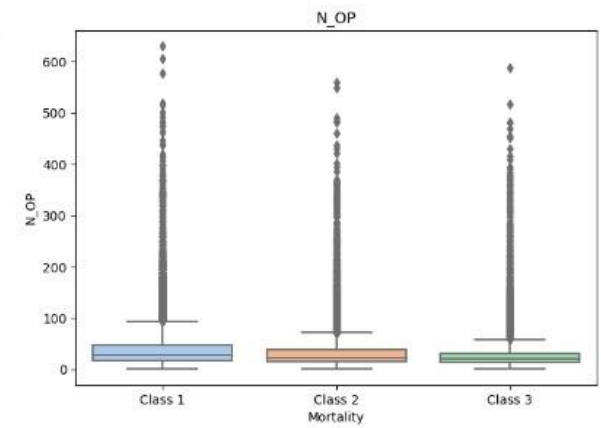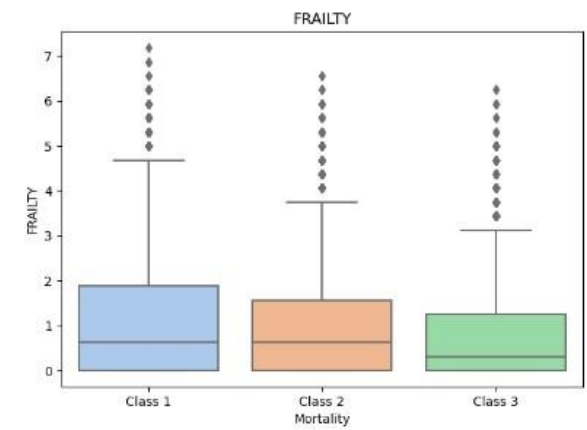

## 2. Count output

Count of each distinct value in all the columns present in the dataset:

```
GROUP 5    103303
GROUP 8    93692
GROUP 1    25850
GROUP 3    19951
GROUP 2    10293
GROUP 7    10233
GROUP 4    6057
GROUP 6    1075
Unknown    86
Name: PRIORITY, dtype: int64
```

------------------------------------------

```
MARRIED    175836
WIDOWED    52336
SINGLE     42368
Name: MARRIED, dtype: int64
```

------------------------------------------

```
0    228368
1    42172
Name: INSULIN, dtype: int64
```

------------------------------------------

```
0    148563
1    121977
Name: SULF, dtype: int64
```

------------------------------------------

```
0    270126
1    414
Name: OTHER_MED, dtype: int64
```

------------------------------------------

```
0    157875
1    112665
Name: BIGUAN, dtype: int64
```

------------------------------------------

```
0    243352
1    27188
Name: TZD, dtype: int64
```

------------------------------------------

```
0    267436
1      3104
```
Name: ALPHA, dtype: int64

------------------------------------------

```
1    267454
0      3086
```
Name: SEX, dtype: int64

------------------------------------------

```
White    234449
Black     27711
Other      8380
```
Name: RACE, dtype: int64

------------------------------------------

```
0    219679
1     50861
```
Name: CHF, dtype: int64

------------------------------------------

```
0    199485
1     71055
```
Name: ARRHYTHMIA, dtype: int64

------------------------------------------

```
0    236911
1     33629
```
Name: VALVULAR, dtype: int64

------------------------------------------

```
0    264088
1      6452
```
Name: PHTN, dtype: int64

------------------------------------------

```
0    209276
1     61264
```
Name: PVD, dtype: int64

------------------------------------------

```
1    247545
0     22995
Name: HTN, dtype: int64
```

------------------------------------------

```
0    241892
1     28648
Name: HTNCX, dtype: int64
```

------------------------------------------

```
0    267045
1      3495
Name: PARALYSIS, dtype: int64
```

------------------------------------------

```
0    256418
1     14122
Name: NEUROOTHER, dtype: int64
```

------------------------------------------

```
0    204444
1     66096
Name: PULMONARY, dtype: int64
```

------------------------------------------

```
0    163203
1    107337
Name: DMCX, dtype: int64
```

------------------------------------------

```
0    239194
1     31346
Name: HYPOTHYROID, dtype: int64
```

------------------------------------------

```
0    237857
1     32683
Name: RENAL, dtype: int64
```

------------------------------------------

```
0    263510
1      7030
Name: LIVER, dtype: int64
```

------------------------------------------

```
0    263282
1      7258
Name: PUD, dtype: int64
```

------------------------------------------

```
0    270364
1       176
Name: HIV, dtype: int64
```

------------------------------------------

```
0    268079
1      2461
Name: LYMPHOMA, dtype: int64
```
------------------------------------------

```
0    268148
1      2392
Name: METS, dtype: int64
```

------------------------------------------

```
0    223635
1     46905
Name: TUMOR, dtype: int64
```

------------------------------------------

```
0    261606
1      8934
Name: RHEUMATIC, dtype: int64
```

------------------------------------------

```
0    257497
1     13043
Name: COAG, dtype: int64
```

------------------------------------------

```
0    215015
1     55525
Name: OBESITY, dtype: int64
```

------------------------------------------

```
0    261688
1      8852
Name: WEIGHTLOSS, dtype: int64
```

```
------------------------------------------
0    237870
1     32670
Name: FLUIDSLYTES, dtype: int64

------------------------------------------
0    266594
1      3946
Name: BLOODLOSS, dtype: int64

------------------------------------------
0    249463
1     21077
Name: ANEMIA, dtype: int64

------------------------------------------
0    264749
1      5791
Name: ALCOHOL, dtype: int64

------------------------------------------
0    268305
1      2235
Name: DRUGS, dtype: int64

------------------------------------------
0    260760
1      9780
Name: PSYCHOSES, dtype: int64

------------------------------------------
0    232375
1     38165
Name: DEPRESSION, dtype: int64

------------------------------------------
0    263981
1      6559
Name: SEVERE_DEP, dtype: int64

------------------------------------------
0    158912
```

1     111628
Name: CAD, dtype: int64

------------------------------------------

0     245768
1     24772
Name: AMI, dtype: int64

------------------------------------------

0     211171
1     59369
Name: RETINOPATHY, dtype: int64

------------------------------------------

0     262812
1     7728
Name: HYPERG, dtype: int64

------------------------------------------

0     256064
1     14476
Name: ABI, dtype: int64

------------------------------------------ 0    217762
1     52778
Name: RETSCREEN, dtype: int64

------------------------------------------

0     270105
1     435
Name: CABG, dtype: int64

------------------------------------------

0     269155
1     1385
Name: PCI, dtype: int64

------------------------------------------

0     238923
1     31617
Name: SMOKER, dtype: int64

------------------------------------------

```
0    268159
1      2381
Name: ESLD, dtype: int64
```

------------------------------------------

```
0    251192
1     19348
Name: FEET, dtype: int64
```

------------------------------------------

```
0    266684
1      3856
Name: AMPUTATION, dtype: int64
```

------------------------------------------

```
1    252322
0     18218
Name: BP_RX, dtype: int64
```

------------------------------------------

```
Class 3    115398
Class 2     90952
Class 1     64190
Name: Mortality, dtype: int64
```

------------------------------------------

```
>=3.5      175441
Missing     82512
<3.5        12587
Name: SERUMALB_RANGE, dtype: int64
```

------------------------------------------

```
Severe       142381
Non-frail     93529
Moderate      34630
Name: FRAILTY_GROUP, dtype: int64
```

------------------------------------------

```
70-74    84165
75-79    72235
80-84    50161
65-69    48557
85-89    14016
>=90      1406
Name: AGE_GROUP, dtype: int64
```

------------------------------------------

<80     238746  
80-89   29219  
>=90    2575  
Name: DIASTOLIC_RANGE, dtype: int64

------------------------------------------

140-179   107999  
130-139   84295  
120-129   53453  
<120      23781  
>=180     1012  
Name: SYSTOLIC_RANGE, dtype: int64

------------------------------------------

0-5   253699  
>5    16841  
Name: N_IP_RANGE, dtype: int64

------------------------------------------

6-30   173873  
>30    93391  
0-5    3276  
Name: N_OP_RANGE, dtype: int64

------------------------------------------

25-39.9    221999  
18.5-24.9   36967   40-49.9    10285  
>=50        822  
<18.5       467  
Name: BMI_RANGE, dtype: int64 ------------------------------------------

<8    222304  
8-9   31281  
>9    16955  
Name: A1C_RANGE, dtype: int64

------------------------------------------

<1.5    211419  
1.5-3.0   45022   Missing    11676  
>3.0     2423  
Name: SERUMCRE_RANGE, dtype: int64

------------------------------------------

```
<40        137315
40-59.99   111450
>=60        14143
Missing      7632
Name: HDL_RANGE, dtype: int64
```

------------------------------------------

```
<100         165538
100-129.99    68667
Missing       17018
130-159.99    16015
160-189.99     2906
>=190           396
Name: LDL_RANGE, dtype: int64
```

------------------------------------------

```
<150        143263
>=200        66227
150-199.99   53839
Missing       7211
Name: TRI_RANGE, dtype: int64
```

------------------------------------------

# APPENDIX C: Section 5 – Additional Results

Table - List of 66 independent features selected by the LASSO technique and Chi-Square Statistics

| Sr. No. | LASSO | Chi-Square Statistics |
|---|---|---|
| 1 | INSULIN | INSULIN |
| 2 | SULF | SULF |
| 3 | BIGUAN | BIGUAN |
| 4 | CHF | CHF |
| 5 | ARRHYTHMIA | ARRHYTHMIA |
| 6 | VALVULAR | VALVULAR |
| 7 | PHTN | PHTN |
| 8 | PVD | PVD |
| 9 | HTN | HTNCX |
| 10 | HTNCX | PARALYSIS |
| 11 | PARALYSIS | NEUROOTHER |
| 12 | NEUROOTHER | PULMONARY |
| 13 | PULMONARY | DMCX |
| 14 | DMCX | HYPOTHYROID |
| 15 | RENAL | RENAL |
| 16 | LYMPHOMA | LIVER |
| 17 | METS | PUD |
| 18 | TUMOR | LYMPHOMA |
| 19 | COAG | METS |
| 20 | OBESITY | TUMOR |
| 21 | WEIGHTLOSS | RHEUMATIC |
| 22 | FLUIDSLYTES | COAG |
| 23 | ANEMIA | OBESITY |
| 24 | ALCOHOL | WEIGHTLOSS |
| 25 | PSYCHOSES | FLUIDSLYTES |
| 26 | DEPRESSION | BLOODLOSS |
| 27 | CAD | ANEMIA |
| 28 | RETINOPATHY | ALCOHOL |
| 29 | CABG | PSYCHOSES |
| 30 | SMOKER | DEPRESSION |
| 31 | ESLD | SEVERE_DEP |
| 32 | FEET | CAD |
| 33 | AMPUTATION | AMI |
| 34 | BP_RX | RETINOPATHY |
| 35 | PRIORITY_GROUP 2 | HYPERG |

| Sr. No. | LASSO | Chi-Square Statistics |
|---|---|---|
| 36 | PRIORITY_GROUP 4 | ABI |
| 37 | PRIORITY_GROUP 5 | RETSCREEN |
| 38 | PRIORITY_GROUP 8 | SMOKER |
| 39 | MARRIED_SINGLE | ESLD |
| 40 | MARRIED_WIDOWED | FEET |
| Sr. No. | LASSO | Chi-Square Statistics |
| 41 | AGE_GROUP_70-74 | AMPUTATION |
| 42 | AGE_GROUP_75-79 | PRIORITY_GROUP 1 |
| 43 | AGE_GROUP_80-84 | PRIORITY_GROUP 4 |
| 44 | AGE_GROUP_85-89 | PRIORITY_GROUP 5 |
| 45 | AGE_GROUP_>=90 | PRIORITY_GROUP 8 |
| 46 | DIASTOLIC_RANGE_80-89 | MARRIED_WIDOWED |
| 47 | SYSTOLIC_RANGE_120-129 | AGE_GROUP_70-74 |
| 48 | SYSTOLIC_RANGE_130-139 | AGE_GROUP_75-79 |
| 49 | SYSTOLIC_RANGE_140-179 | AGE_GROUP_80-84 |
| 50 | SYSTOLIC_RANGE_>=180 | AGE_GROUP_85-89 |
| 51 | N_IP_RANGE_>5 | AGE_GROUP_>=90 |
| 52 | N_OP_RANGE_0-5 | DIASTOLIC_RANGE_80-89 |
| 53 | N_OP_RANGE_>30 | SYSTOLIC_RANGE_130-139 |
| 54 | BMI_RANGE_<18.5 | N_IP_RANGE_>5 |
| 55 | BMI_RANGE_18.5-24.9 | N_OP_RANGE_0-5 |
| 56 | BMI_RANGE_40-49.9 | N_OP_RANGE_>30 |
| 57 | BMI_RANGE_>=50 | BMI_RANGE_<18.5 |
| 58 | A1C_RANGE_8-9 | BMI_RANGE_18.5-24.9 |
| 59 | A1C_RANGE_>9 | A1C_RANGE_8-9 |
| 60 | SERUMCRE_RANGE_1.5-3.0 | A1C_RANGE_>9 |
| 61 | SERUMCRE_RANGE_>3.0 | SERUMCRE_RANGE_1.5-3.0 |
| 62 | HDL_RANGE_<40 | SERUMCRE_RANGE_>3.0 |
| 63 | HDL_RANGE_Missing | HDL_RANGE_<40 |
| 64 | LDL_RANGE_100-129.99 | LDL_RANGE_100-129.99 |
| 65 | TRI_RANGE_>=200 | SERUMALB_RANGE_<3.5 |
| 66 | SERUMALB_RANGE_<3.5 | SERUMALB_RANGE_Missing |

Multiclass Classification results of Multinomial Logistic Regression Classifier

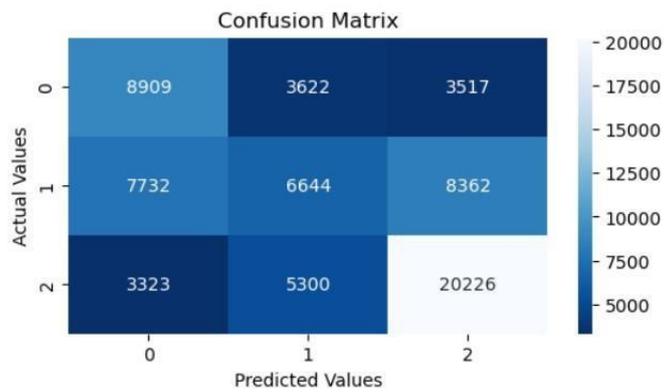

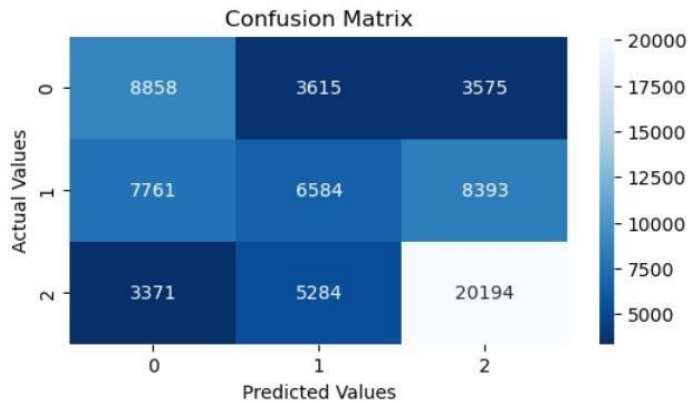

```
Multinomial Logistic Regression (with Information Gain Feature Selection) Model Performance on Hold-Out Test Set:
Accuracy: 52.69 %
```

Confusion Matrix (Multinomial Logistic Regression):
- Row 0: 8858, 3615, 3575
- Row 1: 7761, 6584, 8393
- Row 2: 3371, 5284, 20194

```
Classification Report:
              precision    recall  f1-score   support

           0       0.44      0.55      0.49     16048
           1       0.43      0.29      0.34     22738
           2       0.63      0.70      0.66     28849

    accuracy                           0.53     67635
   macro avg       0.50      0.51      0.50     67635
weighted avg       0.52      0.53      0.51     67635
```

Multiclass Classification results of Random Forest Classifier

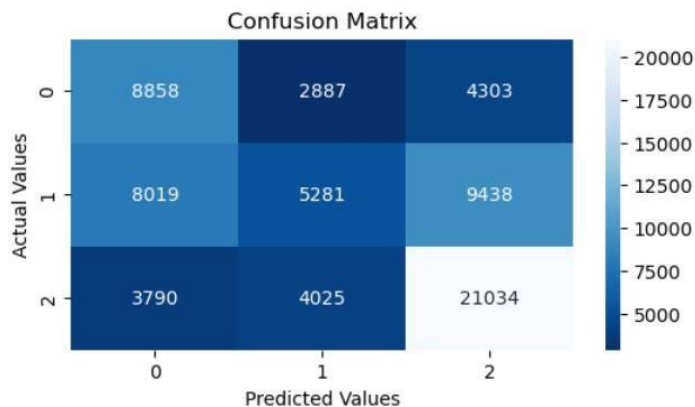

```
Random Forest Ensemble (with Chi-Squared Statistics Feature Selection) Model Performance on Hold-Out Test Set:
Accuracy: 52.0 %
```

Confusion Matrix (Random Forest):
- Row 0: 8858, 2887, 4303
- Row 1: 8019, 5281, 9438
- Row 2: 3790, 4025, 21034

```
Classification Report:
              precision    recall  f1-score   support

           0       0.43      0.55      0.48     16048
           1       0.43      0.23      0.30     22738
           2       0.60      0.73      0.66     28849

    accuracy                           0.52     67635
   macro avg       0.49      0.50      0.48     67635
weighted avg       0.51      0.52      0.50     67635
```

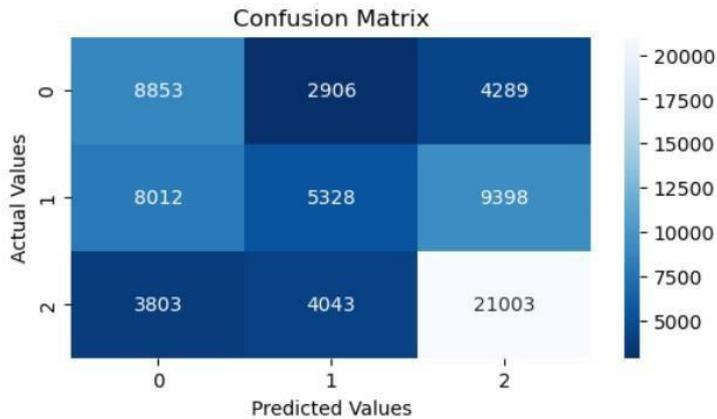

```
Random Forest Ensemble (with Information Gain Feature Selection) Model Performance on Hold-Out Test Set:
Accuracy: 52.02 %
```

Confusion Matrix (Random Forest)

```
Classification Report:
              precision    recall  f1-score   support

           0       0.43      0.55      0.48     16048
           1       0.43      0.23      0.30     22738
           2       0.61      0.73      0.66     28849

    accuracy                           0.52     67635
   macro avg       0.49      0.50      0.48     67635
weighted avg       0.51      0.52      0.50     67635
```

**Multiclass Classification results of XGBoost Classifier**

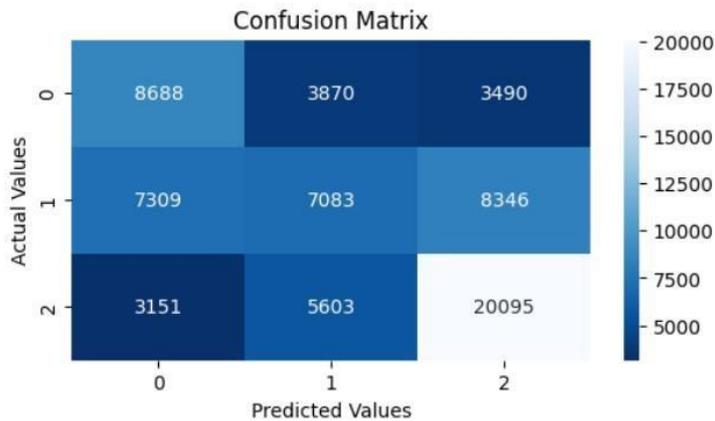

```
XGBoost Ensemble (with Chi-Squared Statistics Feature Selection) Model Performance on Hold-Out Test Set:
Accuracy: 53.03 %
```

Confusion Matrix (XGBoost)

```
Classification Report:
              precision    recall  f1-score   support

           0       0.45      0.54      0.49     16048
           1       0.43      0.31      0.36     22738
           2       0.63      0.70      0.66     28849

    accuracy                           0.53     67635
   macro avg       0.50      0.52      0.51     67635
weighted avg       0.52      0.53      0.52     67635
```

```
XGBoost Ensemble (with Information Gain Feature Selection) Model Performance on Hold-Out Test Set:

Accuracy: 52.94 %
```

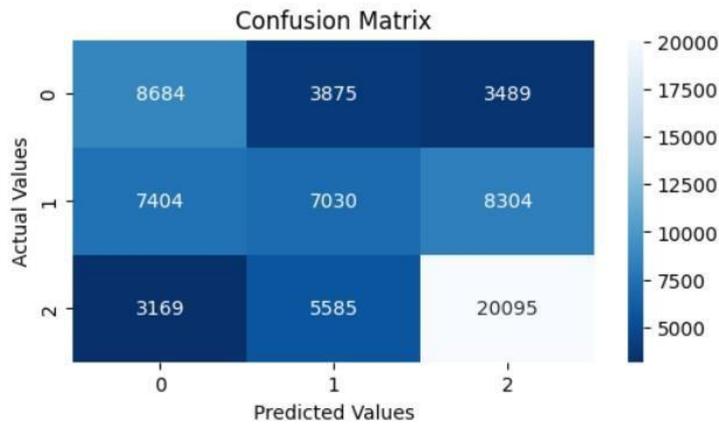

```
Classification Report:
              precision    recall  f1-score   support

           0       0.45      0.54      0.49     16048
           1       0.43      0.31      0.36     22738
           2       0.63      0.70      0.66     28849

    accuracy                           0.53     67635
   macro avg       0.50      0.52      0.50     67635
weighted avg       0.52      0.53      0.52     67635
```

**ROC for One-Vs-Rest Classifier**

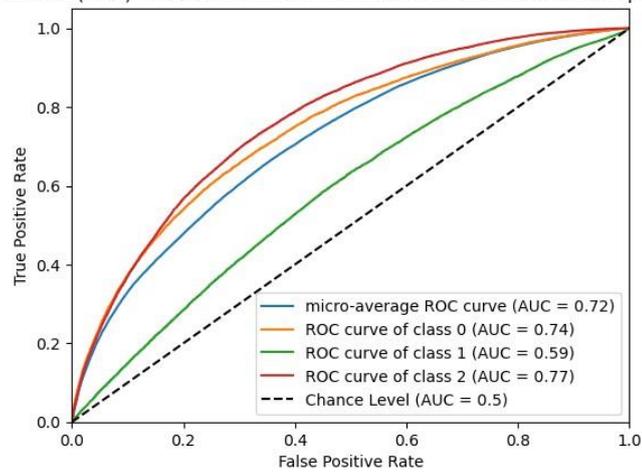

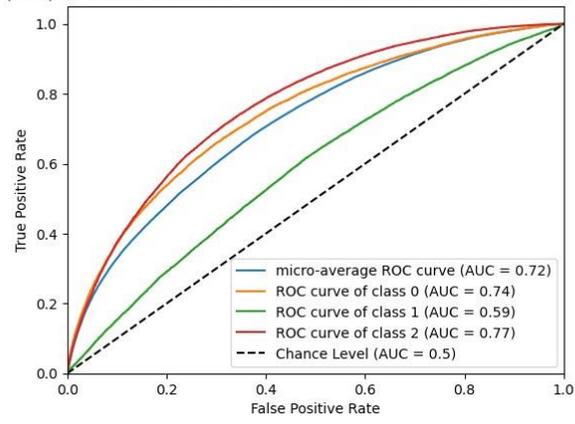

**One-Vs-Rest Classifier - Binary Classification using Chi-Squared Feature Selection**

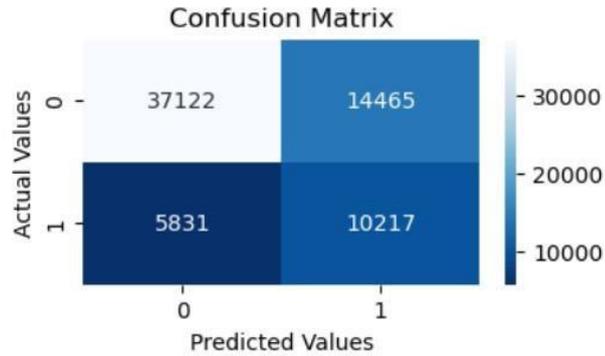

For Mortality Target Value  0 :

Accuracy: 69.99 %

Confusion Matrix:
|  | Predicted 0 | Predicted 1 |
|---|---|---|
| Actual 0 | 37122 | 14465 |
| Actual 1 | 5831 | 10217 |

Classification Report:

|  | precision | recall | f1-score | support |
|---|---|---|---|---|
| 0 | 0.86 | 0.72 | 0.79 | 51587 |
| 1 | 0.41 | 0.64 | 0.50 | 16048 |
| accuracy |  |  | 0.70 | 67635 |
| macro avg | 0.64 | 0.68 | 0.64 | 67635 |
| weighted avg | 0.76 | 0.70 | 0.72 | 67635 |

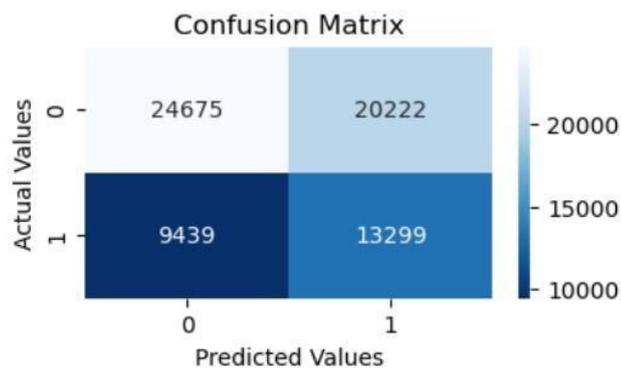

For Mortality Target Value  1 :

Accuracy: 56.15 %

Confusion Matrix:
|  | Predicted 0 | Predicted 1 |
|---|---|---|
| Actual 0 | 24675 | 20222 |
| Actual 1 | 9439 | 13299 |

Classification Report:

|  | precision | recall | f1-score | support |
|---|---|---|---|---|
| 0 | 0.72 | 0.55 | 0.62 | 44897 |
| 1 | 0.40 | 0.58 | 0.47 | 22738 |
| accuracy |  |  | 0.56 | 67635 |
| macro avg | 0.56 | 0.57 | 0.55 | 67635 |
| weighted avg | 0.61 | 0.56 | 0.57 | 67635 |

For Mortality Target Value 2 :

Accuracy: 69.02 %

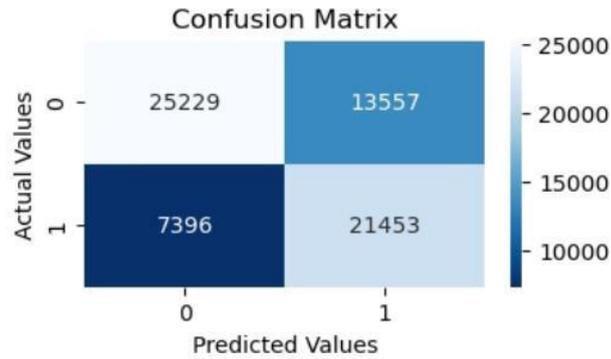

Classification Report:

```
              precision    recall  f1-score   support

           0       0.77      0.65      0.71     38786
           1       0.61      0.74      0.67     28849

    accuracy                           0.69     67635
   macro avg       0.69      0.70      0.69     67635
weighted avg       0.70      0.69      0.69     67635
```

**One-Vs-Rest Classifier - Final Multiclass Classification using Chi-Squared Feature Selection**

Accuracy: 52.45 %

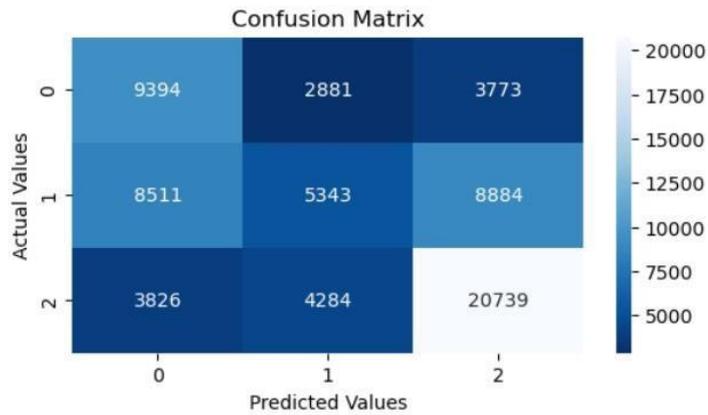

Classification Report:

```
              precision    recall  f1-score   support

           0       0.43      0.59      0.50     16048
           1       0.43      0.23      0.30     22738
           2       0.62      0.72      0.67     28849

    accuracy                           0.52     67635
   macro avg       0.49      0.51      0.49     67635
weighted avg       0.51      0.52      0.50     67635
```

**One-Vs-Rest Classifier - Binary Classification using Information Gain Feature Selection**

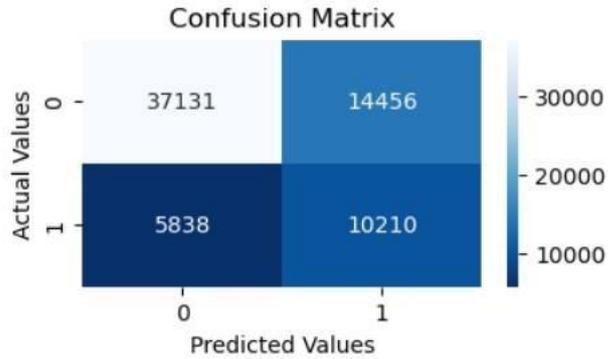

```
For Mortality Target Value  0 :

Accuracy: 69.99 %
```

Confusion Matrix

|  | 0 | 1 |
|---|---|---|
| 0 | 37131 | 14456 |
| 1 | 5838 | 10210 |

```
Classification Report:

              precision    recall  f1-score   support

           0       0.86      0.72      0.79     51587
           1       0.41      0.64      0.50     16048

    accuracy                           0.70     67635
   macro avg       0.64      0.68      0.64     67635
weighted avg       0.76      0.70      0.72     67635
```

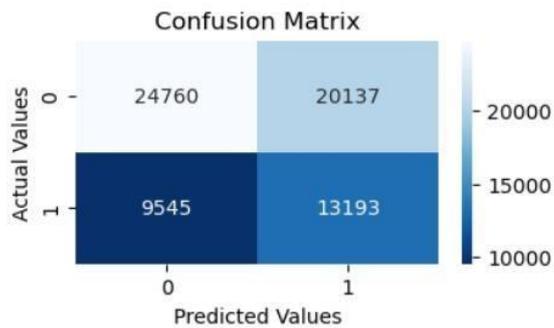

```
For Mortality Target Value  1 :

Accuracy: 56.11 %
```

Confusion Matrix

|  | 0 | 1 |
|---|---|---|
| 0 | 24760 | 20137 |
| 1 | 9545 | 13193 |

```
Classification Report:

              precision    recall  f1-score   support

           0       0.72      0.55      0.63     44897
           1       0.40      0.58      0.47     22738

    accuracy                           0.56     67635
   macro avg       0.56      0.57      0.55     67635
weighted avg       0.61      0.56      0.57     67635
```

For Mortality Target Value 2 :

Accuracy: 69.01 %

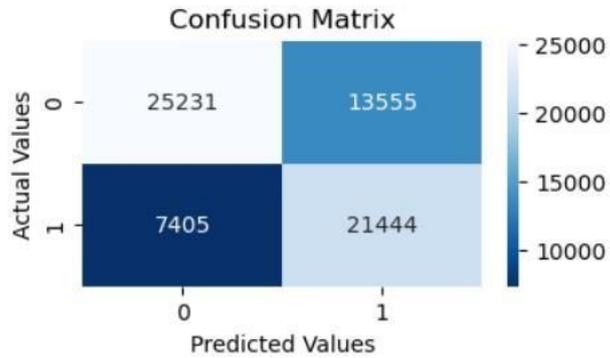

```
Classification Report:

              precision    recall  f1-score   support

           0       0.77      0.65      0.71     38786
           1       0.61      0.74      0.67     28849

    accuracy                           0.69     67635
   macro avg       0.69      0.70      0.69     67635
weighted avg       0.70      0.69      0.69     67635
```

**One-Vs-Rest Classifier - Final Multiclass Classification using Information Gain Feature Selection**

Accuracy: 52.38 %

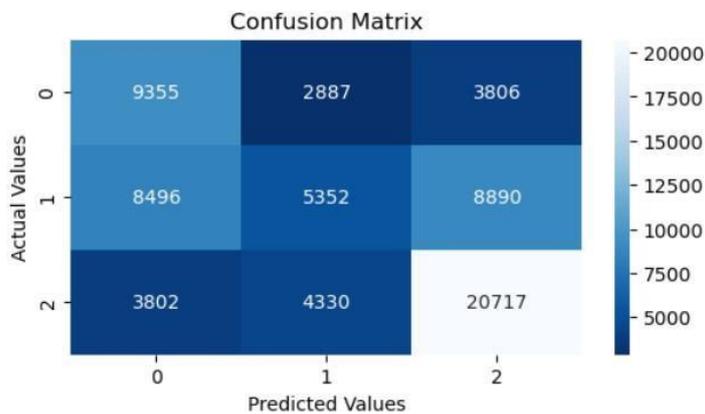

```
Classification Report:

              precision    recall  f1-score   support

           0       0.43      0.58      0.50     16048
           1       0.43      0.24      0.30     22738
           2       0.62      0.72      0.67     28849

    accuracy                           0.52     67635
   macro avg       0.49      0.51      0.49     67635
weighted avg       0.51      0.52      0.50     67635
```